\pdfoutput=1

\documentclass[twoside,11pt]{article}
\newif\ifdraft
\draftfalse

\usepackage[preprint]{jmlr2e}
\setcitestyle{square, comma, numbers,sort&compress, super}

\usepackage[T1]{fontenc}    %
\usepackage{hyperref}       %
\usepackage{url}            %
\usepackage{nicefrac}       %
\usepackage{microtype}      %

\usepackage{graphicx}%
\graphicspath{{figures/}}
\usepackage{multirow}%
\usepackage{amsmath}
\usepackage{amssymb}
\usepackage{amsfonts}
\usepackage{mathrsfs}%
\usepackage[title]{appendix}%
\usepackage{xcolor}%
\usepackage{textcomp}%
\usepackage{manyfoot}%
\usepackage{booktabs}%
\usepackage{algorithm}%
\usepackage{algorithmicx}%
\usepackage{algpseudocode}%
\usepackage{listings}%
\usepackage{algpseudocode}
\usepackage{enumitem}
\usepackage[noabbrev]{cleveref}
\usepackage{xspace}
\usepackage{tikz}
\usepackage{makecell}
\usepackage{longtable}
\usepackage{changepage}
\usepackage{color}
\usepackage{colortbl}
\usepackage{multirow}
\usepackage{subcaption}
\usepackage[font=small,labelfont=bf]{caption}
\usepackage{tablefootnote} %
\usepackage{threeparttable} %

\usepackage{setspace}
\usepackage{changepage} %
\usepackage[breakable]{tcolorbox}
\usepackage{float}
\newcommand{\mfourtcitation}{Seamless Communication et al, \textit{SeamlessM4T: Massively Multilingual \& Multimodal Machine Translation}, Arxiv, 2023}

\newcommand{\mineddata}{\textsc{SeamlessAlign}\xspace}
\newcommand{\mfourt}{\textsc{SeamlessM4T}\xspace}
\newcommand{\nllbv}{\textsc{SeamlessM4T-NLLB}\xspace}
\newcommand{\mfourtlg}{\textsc{SeamlessM4T-Large}\xspace}
\newcommand{\mfourtmd}{\textsc{SeamlessM4T-Medium}\xspace}
\newcommand{\sizelg}{2.3B\xspace}

\newcommand{\sizemd}{1.2B\xspace}

\newcommand{\unity}{\textsc{UnitY}\xspace}

\newcommand{\ntextlangs}{95\xspace}
\newcommand{\nasrlangs}{96\xspace}
\newcommand{\nsslangs}{100\xspace}
\newcommand{\ntslangs}{35\xspace}
\newcommand{\nvocoderlangs}{36\xspace}
\newcommand{\wvbertlangs}{143\xspace}

\newcommand{\mt}{{T2TT}\xspace}
\newcommand{\asr}{{ASR}\xspace}
\newcommand{\st}{{S2TT}\xspace}
\newcommand{\tu}{{T2U}\xspace}
\newcommand{\xt}{{X2T}\xspace}
\newcommand{\sst}{{S2ST}\xspace}
\newcommand{\tts}{{TTS}\xspace}
\newcommand{\ttst}{{T2ST}\xspace}

\newcommand{\xeng}{{X--eng}\xspace}
\newcommand{\engx}{{eng--X}\xspace}
\newcommand{\xy}{{X--X}\xspace}
\newcommand{\chrf}{chrF\nolinebreak\hspace{-.05em}\raisebox{.4ex}{\tiny\bf +}\nolinebreak\hspace{-.10em}\raisebox{.4ex}{\tiny\bf +}}
\newcommand{\spbleu}{\textsc{spBLEU}\xspace}
\newcommand{\bleu}{\textsc{BLEU}\xspace}
\newcommand{\wer}{\textsc{WER}\xspace}
\newcommand{\blaser}{\textsc{Blaser 2.0}\xspace}
\newcommand{\asrbleu}{\textsc{ASR-BLEU}\xspace}
\newcommand{\asrchrf}{\textsc{ASR-chrF}\xspace}
\newcommand{\asretox}{\textsc{ASR-ETOX}\xspace}
\newcommand{\holisticbias}{\textsc{HolisticBias}\xspace}
\newcommand{\multilingualholisticbias}{\textsc{Multilingual HolisticBias}\xspace}

\newcommand{\SpeechMatrix}{\textsc{SpeechMatrix}\xspace}
\newcommand{\NbLangsTotal}{100\xspace}

\newcommand{\NbLangsMined}{37\xspace} %
\newcommand{\DataRawAudioText}{4 million\xspace}
\newcommand{\xsim}{\texttt{xsim}\xspace}
\newcommand{\xsimpp}{\texttt{xsim++}\xspace}
\newcommand{\flores}{\textsc{Flores}\xspace}
\newcommand{\fleurs}{\textsc{Fleurs}\xspace}
\newcommand{\covost}{\textsc{CoVoST} 2\xspace}
\newcommand{\cvss}{\textsc{CVSS}\xspace}
\newcommand{\laser}{\textsc{Laser}\xspace}
\newcommand{\laserthree}{\textsc{Laser3}\xspace}
\newcommand{\nbLangsLaserLabse}{98\xspace} %
\newcommand{\sonar}{\textsc{Sonar}\xspace}
\newcommand{\stopes}{\textsc{Stopes}\xspace} %
\newcommand{\fairseq}{\textsc{Fairseq2}\xspace} %

\newcommand{\xlsr}{XLS-R\xspace}
\newcommand{\whisperlarge}{\textsc{Whisper-Large-v2}\xspace}
\newcommand{\whispermedium}{\textsc{Whisper-Medium}\xspace}
\newcommand{\nllbmedium}{\textsc{NLLB-3.3B}\xspace}
\newcommand{\nllbsmall}{\textsc{NLLB-1.3B}\xspace}

\newcommand{\nllbtinydistil}{\textsc{NLLB-600M-Distilled}\xspace}

\newcommand{\wvbert}{\textsc{w2v-BERT 2.0}\xspace}
\newcommand{\MC}{\multicolumn}
\newcommand{\ra}{\rightarrow}

\newcommand{\totalminedhours}{470,000\xspace}
\newcommand{\noiserobustnessgain}{38\%}
\newcommand{\speakerrobustnessgain}{49\%}
\newcommand{\audiopalmast}{\textsc{AudioPaLM-2-8B-AST}}
\newcommand{\xlsrstot}{\textsc{XLS-R-2B-S2T}}
\newcommand{\mmswithoutlm}{\textsc{MMS-L61-noLM-LSAH}}
\newcommand{\mmswithlm}{\textsc{MMS-L1107-CCLM-LSAH}}
\newcommand{\yourtts}{\textsc{YourTTS}}

\newcommand{\stgainoveraudiopalm}{20\%}
\newcommand{\sstgaincvss}{58\%}

\newcolumntype{H}{>{\setbox0=\hbox\bgroup}c<{\egroup}@{}}

\newenvironment{itemize*}%
  {\begin{itemize}%
    \setlength{\itemsep}{0pt}%
    \setlength{\parsep}{0pt}%
    \setlength{\parskip}{0pt}}%
  {\end{itemize}}

\usepackage{placeins}
\usepackage{verbatim}

\ifdraft
    \providecommand\marta[1]{[\textcolor{magenta}{Marta: {#1}}]}
    \providecommand\david[1]{[\textcolor{brown}{David: {#1}}]}
    \providecommand\maha[1]{[\textcolor{blue}{Maha: {#1}}]}
    \providecommand\christophe[1]{[\textcolor{teal}{Christophe: {#1}}]}
    \newcommand{\changhan}[1]{\textcolor{orange}{[Changhan: #1]}}
    \newcommand{\holger}[1]{\textcolor{brown}{[Holger: #1]}}
    \newcommand{\sravya}[1]{\textcolor{magenta}{[Sravya: #1]}}
    \newcommand{\loic}[1]{\textcolor{cyan}{[Loic: #1]}}
    \newcommand{\skyler}[1]{\textcolor{purple}{[Skyler: #1]}}
    \newcommand{\bokai}[1]{\textcolor{violet}{[Bokai: #1]}}
    \newcommand{\mariano}[1]{\textcolor{green}{[Mariano: #1]}}
    \newcommand{\daniel}[1]{\textcolor{lime}{[Daniel: #1]}}
    \newcommand{\juan}[1]{\textcolor{pink}{[Juan: #1]}}
    \newcommand{\alex}[1]{\textcolor{teal}{[Alex: #1]}}
    \newcommand{\annl}[1]{\textcolor{purple}{[Ann: #1]}}
    \newcommand{\ilia}[1]{\textcolor{red}{[Ilia: #1]}}
    \newcommand{\paden}[1]{\textcolor{orange}{[Paden: #1]}}
    \newcommand{\john}[1]{\textcolor{teal}{[John: #1]}}
    \newcommand{\paco}[1]{\textcolor{orange}{[Paco: #1]}}

\else
    \providecommand\marta[1]{}
    \providecommand\david[1]{}
    \providecommand\maha[1]{}
    \providecommand\christophe[1]{}
    \newcommand{\changhan}[1]{}
    \newcommand{\holger}[1]{}
    \newcommand{\sravya}[1]{}
    \newcommand{\loic}[1]{}
    \newcommand{\skyler}[1]{}
    \newcommand{\bokai}[1]{}
    \newcommand{\mariano}[1]{}
    \newcommand{\daniel}[1]{}
    \newcommand{\juan}[1]{}
    \newcommand{\alex}[1]{}
    \newcommand{\john}[1]{}
    \newcommand{\ilia}[1]{}
    \newcommand{\paden}[1]{}
    \newcommand{\annl}[1]{}
    \newcommand{\paco}[1]{}
\fi

\newlist{todolist}{itemize}{2}
\setlist[todolist]{label=$\square$}

\begin{document}

\title{SeamlessM4T: Massively Multilingual \& Multimodal Machine Translation}

\author{\centering
\textmd{Seamless Communication,
Lo\"{i}c Barrault\footnotemark[1],  
Yu-An Chung\footnotemark[1],  
Mariano Coria Meglioli\footnotemark[1],  
David Dale\footnotemark[1],  
Ning Dong\footnotemark[1],  
Paul-Ambroise Duquenne\footnotemark[1] \textsuperscript{,\textdaggerdbl},  
Hady Elsahar\footnotemark[1],  
Hongyu Gong\footnotemark[1],  
Kevin Heffernan\footnotemark[1],  
John Hoffman\footnotemark[1],  
Christopher Klaiber\footnotemark[1],  
Pengwei Li\footnotemark[1],  
Daniel Licht\footnotemark[1],  
Jean Maillard\footnotemark[1],  
Alice Rakotoarison\footnotemark[1],  
Kaushik Ram Sadagopan\footnotemark[1],  
Guillaume Wenzek\footnotemark[1],  
Ethan Ye\footnotemark[1],   
Bapi Akula,  
Peng-Jen Chen,  
Naji El Hachem,  
Brian Ellis,  
Gabriel Mejia Gonzalez,  
Justin Haaheim,  
Prangthip Hansanti,  
Russ Howes,  
Bernie Huang,  
Min-Jae Hwang,  
Hirofumi Inaguma,  
Somya Jain,  
Elahe Kalbassi,  
Amanda Kallet,  
Ilia Kulikov,  
Janice Lam,  
Daniel Li,  
Xutai Ma,  
Ruslan Mavlyutov,  
Benjamin Peloquin,  
Mohamed Ramadan,  
Abinesh Ramakrishnan,  
Anna Sun,  
Kevin Tran,  
Tuan Tran,  
Igor Tufanov,  
Vish Vogeti,  
Carleigh Wood,  
Yilin Yang,  
Bokai Yu,  
Pierre Andrews\footnotemark[2],  
Can Balioglu\footnotemark[2],  
Marta R. Costa-juss\`{a}\footnotemark[2] \footnotemark[3],  
Onur \,{C}elebi\footnotemark[2], 
Maha Elbayad\footnotemark[2], 
Cynthia Gao\footnotemark[2],  
Francisco Guzm\'an\footnotemark[2],  
Justine Kao\footnotemark[2],  
Ann Lee\footnotemark[2],  
Alexandre Mourachko\footnotemark[2],  
Juan Pino\footnotemark[2],  
Sravya Popuri\footnotemark[2],  
Christophe Ropers\footnotemark[2],  
Safiyyah Saleem\footnotemark[2],  
Holger Schwenk\footnotemark[2],  
Paden Tomasello\footnotemark[2],  
Changhan Wang\footnotemark[2],  
Jeff Wang\footnotemark[2],  
Skyler Wang\footnotemark[2] \textsuperscript{,\S}} \\
\begin{center}
\centering Meta AI, \textsuperscript{\textdaggerdbl}INRIA, \textsuperscript{\S}UC Berkeley
\end{center}
}

\maketitle

\renewcommand*{\thefootnote}{\fnsymbol{footnote}}

\footnotetext[1]{Equal contribution, alphabetical order}
\footnotetext[2]{Research and engineering leadership—equal contribution, alphabetical order}
\footnotetext[3]{Corresponding Author. Email: \textsc{costajussa@meta.com}.}

\renewcommand*{\thefootnote}{\arabic{footnote}}
\setcounter{footnote}{0}

\textbf{Abstract}

What does it take to create the Babel Fish, a tool that can help individuals translate speech between any two languages? While recent breakthroughs in text-based models have pushed machine translation coverage beyond 200 languages, unified speech-to-speech translation models have yet to achieve similar strides. 
More specifically, conventional speech-to-speech translation systems rely on cascaded systems composed of multiple subsystems performing translation progressively, putting scalable and high-performing unified speech translation systems out of reach. 
To address these gaps, we introduce \textbf{SeamlessM4T}—\textbf{M}assively \textbf{M}ultilingual \& \textbf{M}ultimodal \textbf{M}achine \textbf{T}ranslation—a single model that supports
speech-to-speech translation, 
speech-to-text translation, 
text-to-speech translation,
text-to-text translation, 
and automatic speech recognition for up to 100 languages. 
To build this, we used 1 million hours of open speech audio data to learn self-supervised speech representations with \wvbert.
Subsequently, we created a multimodal corpus of automatically aligned speech translations, dubbed \mineddata.
Filtered and combined with human-labeled and pseudo-labeled data (totaling 406,000 hours), we developed the first multilingual system capable of translating from and into English for both speech and text.
On \fleurs, \mfourt sets a new standard for translations into multiple target languages, achieving an improvement of \stgainoveraudiopalm~\bleu over the previous state-of-the-art in direct speech-to-text translation.
Compared to strong cascaded models, \mfourt improves the quality of into-English translation by 1.3 \bleu points in speech-to-text and by 2.6 \asrbleu points %
in speech-to-speech.
On \cvss and compared to a 2-stage cascaded model for speech-to-speech translation, \mfourtlg's performance is stronger by \sstgaincvss.
Preliminary human evaluations of speech-to-text translation outputs evinced similarly impressive results; for translations from English, XSTS scores for 24 evaluated languages are consistently above 4 (out of 5). For into English directions, we see significant improvement over \whisperlarge's baseline for 7 out of 24 languages.
To further evaluate our system, we developed \blaser, which enables evaluation across speech and text with similar accuracy compared to its predecessor when it comes to quality estimation. 
Tested for robustness, our system performs better against background noises and speaker variations in speech-to-text tasks
(average improvements of \noiserobustnessgain~and \speakerrobustnessgain, respectively)
compared to the current state-of-the-art model. 
Critically, we evaluated \mfourt on gender bias and added toxicity to assess translation safety. Compared to the state-of-the-art, we report up to 63\% reduction in added toxicity in our translation outputs. 
Finally, all contributions in this work—including models, inference code, 
finetuning recipes backed by our improved modeling toolkit \fairseq, and metadata to recreate the unfiltered \totalminedhours hours of \mineddata—are open-sourced and accessible at \url{https://github.com/facebookresearch/seamless_communication}.

\clearpage
\setcounter{tocdepth}{2}
\tableofcontents
\clearpage

\section{Introduction}
\label{sec:intro}

\textit{The Hitchhiker's Guide to the Galaxy’s} Babel Fish, \textit{Star Trek’s} Universal Translator, and \textit{Doctor Who’s} Tardis Translation Circuit are all variants of the same thing—computational devices that grant the ability to translate between any two languages. Casting aside their chimeric origins, the social need for realizing such visions has never been greater. For one, an increasingly interconnected world calls for the development of technologies that can facilitate and streamline multilingual contact both online and offline. Moreover, the proliferation of mobile devices and the platform economy worldwide provides the vehicle for on-demand speech-to-speech translation (\sst) to become a staple in most people's lives.

Despite the centrality of speech in everyday communication, machine translation (MT) systems today remain text-centric. Speech support, if and when present, is often seen as cursory to its text-based counterpart. While single, unimodal models such as No Language Left Behind (NLLB; \citep{nllb2022}) push text-to-text translation (\mt) coverage to more than 200 languages, unified \sst models are far from achieving similar scope or performance. This modality-based disparity could be attributed to many causes, but audio data scarcity and modeling constraints remain key obstacles. The very challenge around why speech is harder to tackle from an MT standpoint—that it encodes more information and expressive components—is also why it is superior at conveying intent and forging stronger social bonds between interlocutors. 

Bringing the Babel Fish into technical reality hinges on developing foundational speech-to-speech translation (\sst) systems.
Today, existing systems of such kind suffer from three main shortcomings. 
One, they tend to focus on high-resource languages such as English, Spanish, and French, leaving many low-resource languages behind. 
Two, they mostly service translations from a source language into English (\xeng) and not vice versa (\engx). 
Three, most \sst systems today rely heavily on cascaded systems composed of multiple subsystems that perform translation progressively—e.g., from automatic speech recognition (\asr) to \mt, and subsequently text-to-speech (TTS) synthesis in a 3-stage system.
Attempts to unify these multiple capabilities under one singular entity have led to early iterations of end-to-end speech translation systems \citep{Lavie1997JanusIIIST,jia:interspeech:2019,lee-etal-2022-direct}. 
However, these systems do not match the performance of their cascaded counterparts \citep{agrawal-etal-2023-findings}, which are more equipped to leverage large-scale multilingual components (e.g., NLLB for \mt or Whisper for \asr~\citep{whisper}) and unsupervised or weakly-supervised data.

To address these limitations, we introduce \textbf{\mfourt} (\textbf{M}assively \textbf{M}ultilingual \& \textbf{M}ultimodal \textbf{M}achine \textbf{T}ranslation), a unified system that supports 
\asr, 
\mt, 
speech-to-text translation (\st), 
text-to-speech translation (\ttst),
and \sst (see \Cref{tbl:tasks} for an overview). 
To build this, we used 1 million hours of open speech audio data to learn self-supervised speech representations with \wvbert.
Subsequently, we created a multimodal corpus of automatically aligned speech translations of more than \totalminedhours hours, dubbed \mineddata.
We then combined a filtered subset of this corpus with human-labeled and pseudo-labeled data, totaling 406,000 hours.
Drawing on this assembled dataset, we developed the first multitasking system that performs 
\sst from \nsslangs languages to English (\nsslangs-eng) and from English to \ntslangs languages (eng-\ntslangs), 
\st for \nsslangs-eng and eng-\ntextlangs languages, 
\asr for \nasrlangs,
zero-shot \ttst for \ntextlangs-eng and eng-\ntslangs languages,
as well as T2TT for \ntextlangs-eng and eng-\ntextlangs (see \Cref{tbl:coverage} for an overview). 

\begin{table}[!th]
\centering
\small
\begin{tabular}{ll}
\toprule
{\bf Task} & {\bf Description} \\
\midrule
\asr & Automatic Speech Recognition \\
\sst & Speech-to-Speech Translation \\
\st & Speech-to-Text Translation \\
\ttst & Text-to-Speech Translation \\
\mt & Text-to-Text Translation \\
\xt & \{Speech,Text\}-to-Text Translation (multitasking models translating into text) \\
\midrule
Task \engx & A translation task from English\\
Task \xeng & A translation task into English\\
Task \xy & A translation task on non-English-centric direction\\
\bottomrule
\end{tabular}
\caption{Notations of tasks in this work.}\label{tbl:tasks}
\end{table}

\begin{table}[!th]
\label{tab:models}
\centering
\small
\begin{tabular}{lcccccc}
\toprule
\multirow{2}{*}{\bf Model} & \multirow{2}{*}{\bf size} & \multicolumn{5}{c}{\bf Task Language Coverage$^\dagger$} \\
\cmidrule(lr){3-7}
&  & \st{} & \sst{} & \asr{} & \mt{} & \ttst \\
\midrule
\multicolumn{6}{l}{\textit{Proprietary models}} \\
USM~\citep{google_usm} & 2B+ & 21-eng & - & 102 & - & - \\
\cite{Rubenstein2023AudioPaLMAL} \\
\phantom{abc}{AudioPaLM-2-8B-AST} & 8.0B & 98-eng & - & 98 & - & - \\
\phantom{abc}{AudioPaLM-8B-S2ST} & 8.0B & 113-Eng & 113-eng & 98 & - &  -\\
\midrule
\multicolumn{6}{l}{\textit{Open models}} \\
\cite{nllb2022} \\
\phantom{abc}{\nllbtinydistil} & 0.6B & - & - & - & 202-202 & - \\
\phantom{abc}{\nllbsmall} & 1.3B & - & - & - &  202-202 & - \\
\phantom{abc}{\nllbmedium} & 3.3B & - & - & - & 202-202 & - \\
\cite{babu2021xls} \\
\phantom{abc}{\xlsrstot} & 2.6B & \makecell[c]{21-eng\\ eng-15} & - & - & - \\
\cite{whisper} \\
\phantom{abc}{\whispermedium} & 0.8B & 96-eng & - & 97 & - & - \\
\phantom{abc}{\whisperlarge} & 1.6B & 96-eng & - & 97 & - & -\\
{MMS}~\citep{mms}  \\
\phantom{abc}{\mmswithoutlm} & 1.0B & - & - & 61 & - & - \\
\phantom{abc}{\mmswithlm} & 1.0B & - & - & 1107 & - & - \\
\midrule
This work (\mfourt) \\
\phantom{abc}\mfourtlg  & \sizelg 
& \makecell[c]{\nsslangs{}-eng\\ eng-\ntextlangs{}} 
& \makecell[c]{\nsslangs{}-eng\\ eng-\ntslangs{}}
& \nasrlangs{}
& \makecell[c]{\ntextlangs{}-eng\\ eng-\ntextlangs{}} 
& \makecell[c]{\ntextlangs{}-eng\\eng-\ntslangs{}} \\
\cmidrule{2-7}
\phantom{abc}\mfourtmd  & \sizemd 
& \makecell[c]{\nsslangs{}-eng\\eng-\ntextlangs{}} 
& \makecell[c]{\nsslangs{}-eng\\eng-\ntslangs{}} 
& \nasrlangs{}
& \makecell[c]{\ntextlangs{}-eng\\eng-\ntextlangs{}} 
& \makecell[c]{\ntextlangs{}-eng\\eng-\ntslangs{}} \\
\cmidrule{2-7}
\phantom{abc}{\nllbv-1.3B} & 1.3B & - & - & - 
& \makecell[c]{\ntextlangs{}-eng\\eng-\ntextlangs{}} 
& - \\
\bottomrule
\end{tabular}
\caption{A list of state-of-the-art baseline models and \mfourt{} models.
$^\dagger$Language coverage is estimated based on use of supervised labeled data or evaluated zero-shot languages and directions. 
}\label{tbl:coverage}
\end{table}

We find that \mfourtlg, the larger model of the two we release, outperforms the previous state-of-the-art (SOTA) end-to-end \st model (\audiopalmast~\citep{Rubenstein2023AudioPaLMAL}) by 4.2 \bleu points on \fleurs~\citep{fleurs2022arxiv} when translating into English (i.e., an improvement of 20\%). Compared to cascaded models, \mfourtlg improves translation accuracy by over 2 \bleu points.
When translating from English, \mfourtlg improves on the previous SOTA (\xlsrstot~\citep{babu2021xls}) by 2.8 \bleu points on \covost~\citep{wang2021covost}, and its performance is on par with cascaded systems on \fleurs.
On the \sst task, \mfourtlg outperforms strong 3-stage cascaded models (\asr, \mt and TTS) by 2.6 \asrbleu points on \fleurs. On \cvss, \mfourtlg outperforms a 2-stage cascaded model (\whisperlarge + \yourtts~\citep{casanova2022yourtts}) by a large margin of 8.5 \asrbleu points (a 50\% improvement).
Preliminary human evaluations of \st outputs evinced similarly impressive results. For translations from English, XSTS scores for 24 evaluated languages are consistently above 4 (out of 5); for into English directions, we see significant improvement over \whisperlarge's baseline for 7 out of 24 languages. 

In addition, \mfourtlg further outperforms \whisperlarge~\citep{whisper} on \fleurs~\asr with an average word error rate (\wer) reduction of 45\% over 77 overlapping languages. When evaluating \mt on \flores~\citep{flores101:2021}, our model matches the performance of NLLB-3.3B~\citep{nllb2022} when translating into English and improves by 1 \chrf~point on average when translating from English.
To further evaluate \mfourt's performance in \st and \sst, we developed \blaser, a language and modality-agnostic evaluation metric for text or speech translation. \blaser 
enables evaluation across speech and text modalities with similar accuracy to its predecessor —BLASER~\citep{chen-etal-2023-blaser}—when it comes to quality estimation.
We also evaluated model robustness against background noises and speaker variations by creating open robustness benchmarks based on \fleurs. 
Result-wise, \mfourtlg is more robust than \whisperlarge against background noises and speaker variations with an average improvement of \noiserobustnessgain~and \speakerrobustnessgain, respectively.

Regarding Responsible AI, we focused on added toxicity and gender bias evaluation. On average, we find a low prevalence of added toxicity, varying between 0.11\% and 0.21\% across modalities, datasets, and translation directions. We significantly reduce added toxicity in all conditions when compared to state-of-the-art models (ranging from 26\% to 63\%). The greatest added toxicity reduction is achieved for \st{} when compared to \whisperlarge. Beyond this, we also evaluated for gender bias on the \multilingualholisticbias datasets and found that \mfourt overgeneralizes to masculine forms when translating from neutral terms (with an average preference of $\sim$10\%) while showing a lack of robustness when varying gender by an amount of $\sim$3\%. For these conditions, \mfourt achieved comparable results to state-of-the-art models. 
We document these effects to motivate further mitigation efforts.

To spur further research in speech translation and to make our work available to the community, we open-source the following at \url{https://github.com/facebookresearch/seamless_communication}:
\begin{itemize}
\item \mfourt models, including model weights for \mfourtlg (2.3B parameters) and \mfourtmd (1.2B parameters), as well as their inference code and fine-tuning recipes powered by our new modeling toolkit \fairseq.\footnote{\url{https://github.com/facebookresearch/fairseq2}}

\item Tools for creating aligned speech data, including metadata to recreate the unfiltered \totalminedhours hours of \mineddata,  \stopes-based pipelines\footnote{\url{https://github.com/facebookresearch/stopes}} to create alignments similar to \mineddata, and \sonar for speech encoders in 37 languages and text encoders in 200 languages.\footnote{\url{https://github.com/facebookresearch/SONAR}}

\item A text-free S2ST automatic evaluation model, \blaser, inclusive of model weights and inference scripts.
\end{itemize}

The rest of the article is structured as follows: \Cref{sec:appendix:problem} describes the sociotechnical dimensions of multimodal translation and motivates why speech is an important modality to tackle in the context of MT research. 
It also includes the list of languages and evaluation metrics that our work covers. 
\Cref{sec:appendix:data} discusses how we created a corpus of automatically aligned speech translations of more than \totalminedhours hours by developing an extended speech-language identification system and a new multimodal text embedding space imperative to our data mining process. 
\Cref{sec:appendix:modeling} details the various modeling techniques we devised to train a multimodal and multitasking translation model that supports multiple languages for source and target sides in both text and speech.
\Cref{sec:appendix:evaluation} documents the automatic and human evaluation of our translation outputs, and the robustness of our models in various settings. 
\Cref{sec:appendix:ra} focuses on our Responsible AI effort, where we evaluated our model outputs for bias and toxicity. Finally, we conclude in \Cref{sec:conclusion}, where we discuss the social impact of our work while reflecting on existing challenges and future possibilities.

\FloatBarrier
\section{The Sociotechnical Dimensions of Multimodal Translation}
\label{sec:appendix:problem}

\subsection{Why Prioritize Speech in Machine Translation?}
As is the case with most technologies within natural language processing (NLP) and other language-based research enterprises, MT reached greater maturity in the modality that affords easier record-keeping, data storage, and dispersion: text. By extension, the abundance of digital text makes it a prime candidate for NLP research. In contrast, the relative paucity of speech data relegates research in this area to secondary importance. More specifically, speech is not just spoken text—the two modalities can differ in grammar, registers, and morphology \citep{plag1999morphological}. In most situations, speech may also appear to be a richer modality, possessing prosodic and expressive parameters unmatchable by text \citep{kraut1992task}. Distinctive in their level of interactivity and sociality, speech directs focus at the speaker or audience, while text spotlights the content of a message \citep{kraut1992task}.

\paragraph{Speech \& social bonding} Research suggests that compared to text-based exchange, communication through speech creates stronger social bonds between interlocutors. For example, in one study, researchers found that interactions including speech (phone, video call, and voice chat) spurred deeper connections between conversation partners compared to those who communicated via text-based media \citep[595]{kumar2021s}. Juxtaposed against speech, which comes with paralinguistic cues such as volume, intonation, and pace, text-based communication is perceived as more impersonal. Interestingly, seeing another person did not make individuals feel more connected than if they had just spoken with their partners. In another study, hearing an outgroup member explain their views out loud made study participants consider them more thoughtful and emotionally warm than reading an explanation of their views \citep{schroeder2017humanizing}. Across a variety of settings, research demonstrates that speech appears to be unique in its ability to convey one's human traits and, consequentially, strengthen the connection between those sharing an exchange.

\paragraph{Inclusion \& accessibility} Speech is not only key to communication from a relational standpoint but is also often the most practical and accessible option. For one, UNESCO estimates that 773 million adults (12.5 percent of all adults) worldwide have not received the education necessary to read or write, thus precluding them from using text to communicate or acquire information \citep{Markelova_2021}. Another group more reliant on speech than text in their everyday lives is those who are blind or with visual impairments. Globally, approximately 43 million people belong to this former category, and 295 million others have moderate to severe visual impairment \citep{bourne2021trends}. Even though voice assistants, text-to-speech systems, and voice-activated technologies today play an important role in supporting these individuals to accomplish everyday tasks, their access to multilingual speech-based translation or communicative tools remains limited. In a world where the volume of auditory content (i.e., podcasts, audiobooks, short-form videos, etc.) is on the rise, the prohibitive nature of this sociotechnical gap may deprive them of experiences or exchanges that could be meaningful and enriching.

\paragraph{Script variance} Beyond these factors, text-based communication or translation is further complicated by script variance. For instance, some languages are written in different scripts on either side of a geopolitical border. %
Urdu, for example, could be written either in the Arabic or Devanagari script depending on where one lives (i.e., Pakistan or India). In such a context, \mt outputs into Urdu may be illegible to those shown in a script they are unfamiliar with. \sst, which produces speech outputs, circumvents this multiscript conundrum. In a few other cases, political instabilities around a language's writing system may also motivate the need for speech-based translation. For example, in the last 1,000 years, Uzbek has changed its writing system five times. Despite the fact that—as of February 2021—Uzbekistan announced Uzbek's official transition from the Cyrillic script to a Latin-based alphabet, the former continues to be widely deployed in the country \citep{jung2023coexistence}. For languages where writing systems are actively negotiated, speech-based technologies and translation systems may provide stabilized access to information as transitions unfold.

\begin{table}[t]
    \small
    \centering
    \begin{tabular}{lcH@{}}
        \toprule
        \multicolumn{3}{l}{\textit{Cascaded models for \st}} \\
        \midrule
        \whispermedium + \nllbtinydistil & 2-stage cascaded & 1.4B  \\
        \whisperlarge + \nllbsmall & 2-stage cascaded & 2.8B \\
        \midrule
        \multicolumn{3}{l}{\textit{Cascaded models for \sst}}\\
        \midrule
        \whisperlarge + \nllbsmall + \yourtts & 3-stage cascaded & 2.8B  \\
        \whisperlarge (\st) + \yourtts & 2-stage cascaded & 1.5B \\
        \midrule
        \mfourt (this work) & unified & 2B \\
        \bottomrule
    \end{tabular}
    \caption{Options for 2-stage and 3-stage cascaded systems for \st and \sst. These 
 cascades pair Whisper \asr models~\citep{whisper} with NLLB's \mt models~\citep{nllb2022}. 
    }    \label{tbl:cascadingmodelsfors2tt}
\end{table}

\subsection{Speech Translation Today}
\paragraph{Cascaded systems}
Before the emergence of unified speech translation models in recent years, much attention in speech-based research has been directed at cascaded approaches by chaining subsystems that perform disparate tasks such as \asr, \mt, and TTS~\citep{Lavie1997JanusIIIST,Wahlster2000VerbmobilFO,nakamura2006}. For example, in a 3-stage \sst cascaded scenario, speech input is first transcribed into text through an \asr system, followed by \mt, and finally synthesized into speech using TTS (see \Cref{tbl:cascadingmodelsfors2tt}). The main benefit of cascaded systems is that they can take advantage of advancements made in areas associated with each subsystem, such as recently released large-scale multilingual \mt models~\citep{nllb2022, siddhant2022towards, m2m100} and weakly-supervised ASR models~\citep{whisper, google_usm, mms}.

That said, cascaded systems have their limitations. For one, the output of a 2-stage cascaded \st{} system involving \asr and \mt does not match the quality achievable by a single large-scale \mt model.
This drop in performance underscores the challenge of transferring and translating meaning across modalities and can be attributed to many factors, including:
(1) poor transcriptions by \asr models for non-English languages, particularly for low-resourced ones,
(2) an increased likelihood of error propagation from the \asr model to the \mt model and other subsequent models in the cascade (the accumulation of errors exacerbates performance), and
(3) domain mismatches between these separately trained subsystems (for example, if an ASR model trained on Wikipedia is used in conjunction with a \mt model optimized for conversational data, this formation may lead to a distribution mismatch at the \mt stage). Beyond these reasons, the overemphasis on text in cascaded systems omits paralinguistic features and may not adequately handle elements such as proper names and nouns \citep{Rubenstein2023AudioPaLMAL}. %

\paragraph{Direct \st models}
Early research into end-to-end speech translation started with producing text as output~\citep{7472621,Berard2016ListenAT,berard2018end}.
Since the emergence of multilingual end-to-end \st models in 2019 \citep{Gangi2019OnetoManyME,inaguma2019multilingual}, \st has become an increasingly popular research area, and many existing models today are powered by the emergence of open multilingual speech corpora like \mbox{MuST-C}~\citep{di2019must}, EuroParl-ST~\citep{iranzo2020europarl}, \covost~\citep{wang2021covost} and VoxPopuli~\citep{wang-etal-2021-voxpopuli}.  End-to-end models today have made significant progress and achieved parity with cascaded models on academic benchmarks in several contexts (e.g., constrained data, in-domain settings, specific language pairs, etc.)~\citep{ansari-etal-2020-findings,potapczyk-przybysz-2020-srpols}

While recent state-of-the-art pre-trained models have seen rapid improvements in language coverage, going from 128 in~\cite{babu2021xls} to more than 1,400 in~\cite{mms}, they only translate into English and not the other way around. Another prominent model, Google's Universal Speech Model~\citep{google_usm}, is pre-trained in more than 300 languages and can perform \asr on more than 100 languages. Technically, USM can also be adapted to perform \asr and \st tasks in any of the 300+ covered languages
once given supervised data (but the model was fine-tuned and evaluated on \covost, which only covers translations from 21 languages into English).

OpenAI's Whisper~\citep{whisper} is another large-scale model that serves translations into English, not vice versa. As a multitasking model, Whisper demonstrates that scaling weakly supervised pre-training is sufficient for achieving SOTA \asr and \st results sans self-supervision and self-training techniques. Trained on 680,000 hours of data, Whisper has achieved SOTA translation quality in 82 \fleurs languages into English.

Combining a text-based \citep{anil2023palm} and speech-based language model \citep{10158503}, the most recently released AudioPaLM~\citep{Rubenstein2023AudioPaLMAL} is a large language model designed for joint text and speech processing and generation. Akin to USM, AudioPaLM only evaluates text translation outputs from 101 \fleurs languages into English. Upon the publication of this paper, AudioPaLM is the current SOTA model, outperforming Whisper~\citep{whisper} in both \asr and \st tasks. 

\begin{table}[t]
\label{tab:metrics}
\centering
\footnotesize
\begin{tabular}{lllll}
\toprule
{\bf Task} & {\bf Metric} &  {\bf Type} &{ \bf \makecell[l]{Area}} & {\bf Details} \\ 
\midrule
{\bf ASR} & \wer & & \makecell[l]{Quality\\ Robustness} & {Text normalization follows Whisper$^\star$}\\
\midrule
{\bf \mt} & 
\makecell[l]{\chrf$^\dagger$} &  Automatic &
Quality &
\makecell[l]{SacreBLEU signature:\\ {\tiny nrefs:1|case:mixed|eff:yes|nc:6|nw:2|space:no|version:2.3.1}}  \\
\cmidrule{2-5}
& \makecell[l]{BLEU$^\ddagger$}  & Automatic & 
Quality &
\makecell[l]{ SacreBLEU signature:\\
{\tiny nrefs:1|case:mixed|eff:no|tok:13a|smooth:exp|version:2.3.1}\\
 Except for cmn, jpn, tha, lao and mya with\\
 character-level tokenization:\\
{\tiny nrefs:1|case:mixed|eff:no|tok:char|smooth:exp|version:2.3.1}
}
\\
\cmidrule{2-5}
& \blaser & \makecell[l]{Automatic\\Model-based}  & Quality\\
\midrule
{\bf \st }& \bleu  & Automatic & \makecell[l]{Quality\\ Robustness\\Bias} &  Similar to T2TT \\
\cmidrule{2-5}
& \blaser & \makecell[l]{Automatic\\Model-based} & Quality & \cite{chen-etal-2023-blaser} \\ \cmidrule{2-5}
&  XSTS &  Human &  Quality &\cite{licht2022xsts}\\
\cmidrule{2-5}
&  chrF$_{MS}$ & Automatic & \makecell[l]{Robustness\\Bias} & \makecell[l]{following \cite{wang-etal-2020-covost}, replaced \bleu \\with chrF for the quality metric\\SacreBLEU signature:\\\tiny nrefs:1|case:mixed|eff:yes|nc:6|nw:2|space:no|version:2.3.1} \\
\cmidrule{2-5}
&  CoefVar$_{MS}$ & Automatic & Robustness & \makecell[l]{ following \cite{wang-etal-2020-covost}, replaced \bleu\\with chrF for the quality metric\\SacreBLEU signature:\\\tiny nrefs:1|case:mixed|eff:yes|nc:6|nw:2|space:no|version:2.3.1} \\
\cmidrule{2-5}
& ETOX & Automatic & Toxicity\\
\midrule
{\bf \sst } & \asrbleu & Automatic & \makecell[l]{Quality} & \makecell[l]{%
 Transcribing English with \whispermedium\\  
 and non-English with \whisperlarge\\
\bleu on normalized transcriptions\\ following \cite{whisper}} \\
\cmidrule{2-5}
& \asrchrf & Automatic & Bias & \makecell[l]{%
 Transcribing English with \whispermedium\\  
 and non-English with \whisperlarge\\
 chrF on normalized transcriptions\\ following \cite{whisper}} \\%
\cmidrule{2-5}
& \blaser & \makecell[l]{Automatic\\Model-based} & \makecell[l]{Quality\\ Bias}\\ 
\cmidrule{2-5}
&  XSTS &  Human & Quality  &\\
\cmidrule{2-5}
&  MOS &  Human &  Naturalness &  \\
\cmidrule{2-5}
& \asretox & Automatic & Toxicity & \makecell[l]{%
 Transcribing English with \whispermedium\\  
 and non-English with \whisperlarge\\
 ETOX on normalized transcriptions\\ following \cite{whisper}}  \\
 \midrule
{\bf \ttst } & \asrbleu & Automatic & \makecell[l]{Quality} & \makecell[l]{%
Similar to \sst} \\
\bottomrule
\end{tabular}
\caption{The list of automatic and human evaluation metrics used by this work. %
$^\star$~\url{https://github.com/openai/whisper/tree/main/whisper/normalizers} $^\dagger$ \cite{popovic2015chrf} $^\ddagger$ \cite{papineni2002bleu}}\label{tbl:eval:metrics}
\end{table}

\paragraph{Direct \sst models}
Beyond text outputs, recent speech translation research has focused on building models that directly produce target speech representations (i.e., spectrograms, discrete units, etc.). In this area, Translatotron~\citep{jia:interspeech:2019} emerged as the first direct \sst model. When it comes to quality, however, the model lagged behind 2-stage cascaded systems by a large margin.
Translatotron-2~\citep{pmlr-v162-jia22b} significantly improved its predecessor's performance and bridged the gap with cascaded systems by incorporating a two-pass decoding approach. Although Translatotron relied on \st as an auxiliary task during training, the target spectrograms were directly generated at inference time. Translatotron-2, on the other hand, relies on the intermediate decoding outputs of phonemes.

Concurrently with Translatotron, \cite{tjandra2019speech} proposed \sst models based on discrete speech representations that do not require text transcriptions in training.
These discrete representations or \textit{units} are learned through unsupervised term discovery and a sequence-to-sequence model trained to translate units from one language to another.
Relatedly, \cite{lee-etal-2022-direct} uses HuBERT \citep{hsu2021hubert}, a pre-trained speech representation model, to encode speech and learn target-side discrete units. \sst is, thus, decomposed into speech-to-unit (S2U) and subsequently unit-to-speech with a speech re-synthesizer~\citep{polyak21_interspeech}.

\paragraph{On coverage and evaluation of \sst systems}\label{sec:statement:evaluation}
To date, the aforementioned AudioPaLM \citep{Rubenstein2023AudioPaLMAL}, which supports both text and speech as input and output, is the current SOTA for \st and \sst. 
Although the model design suggests that it can support multilingual translation on both source and target sides, its performance is only reported for translating into English. 
Similarly, although Whisper can transcribe non-English languages, it only supports \st into English. 
To consolidate the current landscape of language coverage and related tasks in speech translation systems, we provide in \Cref{tbl:coverage} a list of SOTA models in text and speech translation. This language coverage
is estimated based on supervised labeled data or evaluated zero-shot languages and directions. We also provide the list of \asr, \mt, \st and \sst evaluation metrics used by this work in \Cref{tbl:eval:metrics}. For \sst, our evaluation focuses on the semantic content of the translation.
Throughout this paper, we primarily evaluated our models on the following datasets:
\begin{itemize}
    \item \flores-200~\citep{nllb2022}: a many-to-many multilingual translation benchmark dataset for 200 languages (we evaluated on devtest).
    \item \fleurs~\citep{fleurs2022arxiv}: an n-way parallel speech and text dataset in 102 languages built on the text translation \flores-101 benchmark~\citep{flores101:2021}. \fleurs is well suited for several downstream tasks involving speech and text.
    We evaluated on the test set, except in ablation experiments where we evaluated on the dev set.
    \item \covost~\citep{wang2021covost}: a large-scale multilingual \st corpus covering translations from 21 languages into English and from English into 15 languages. We evaluated on the test set.
    \item \cvss~\citep{cvss}: a multilingual-to-English speech-to-speech translation (\sst) corpus, covering sentence-level parallel \sst pairs from 21 languages into English. We evaluated text-based semantic accuracy on \cvss-C for the tasks of \sst and \ttst. We note that some samples from the evaluation data were missing (in 8 out of 21 languages: Catalan, German, Estonian, French, Italian, Mongolian, Persian, and Portuguese). 
\end{itemize}

\paragraph{The overarching goals of this effort}
In light of the gaps delineated above, our work seeks to advance speech translation in the following ways:
\begin{enumerate}
    \item Creating a unified large model that can handle the full suite of tasks involved in text and speech translation: \sst, \st, \ttst, \mt, and \asr. This lays the important groundwork for the next generation of on-device and on-demand multimodal translation, which can be derived from this model.
    \item Expanding language coverage both in terms of the number of supported languages and translation directions (i.e., going beyond translations into English by including translation from English). That roughly two dozen languages account for more than half of the world's speaking population means that a relatively small group of languages (out of more than 7,000) produce a disproportionately large linguistic footprint. Whether in the text or speech modality, these languages are deemed high-resource, giving them prioritization in today's AI development. That said, when language technologies are developed primarily with this group in mind, the needs of half the world's population are left behind. Our effort seeks to bridge the translation gap between those who speak high and low-resource languages.
    \item Maintaining systematic evaluations of our systems throughout our workflow to ensure safe and robust performance. This allows us to understand how to direct our efforts to make both the current and future iterations of our contribution more equitable and fair across user demographics. %
\end{enumerate}

\subsection{Languages}

Today, broadly accessible speech translation models cover anywhere between 21 \citep{google_usm} to 113 \citep{Rubenstein2023AudioPaLMAL} source languages depending on the wide range of tasks involved. However, none of these existing speech-based translation models can also service \mt. 
To build a unified, multimodal, and multitask model that can handle both speech and text as source inputs, we set our speech source language goal at 100. 

We summarize information about each of our supported languages in \Cref{tab:all_languages}. Further details on the table headers are provided below.

\paragraph{Code}
We represent each language with a three-letter ISO 639-3 code. 

\paragraph{Language}
There may be multiple ways to refer to the same language; due to formatting limitations, only one of the versions is displayed. The language names have been cross-referenced with major linguistic information platforms such as Ethnologue~\citep{ethnologue2009} and Glottolog ~\citep{glottolog2022}.

\paragraph{Family and subgrouping}
We provide Language family information for each language based on the Glottolog database~\citep{glottolog2022}.

\paragraph{Script}
We provide script information in ISO 15924 codes for writing systems. 

\paragraph{Resource level} We categorize the speech resource level as high, medium, or low depending on the volume of available primary data for \st into English (with $x$ the amount of primary data in hours, \textit{high} if $x>1000$, \textit{medium} if $x\in]500, 1000]$ and \textit{low} if $x\in[0, 500]$).

\textit{Primary data} is defined as open-source \st and pseudo-labeled ASR data. Absent such data, we report the language as zero-shot (when evaluating \st into English).

\paragraph{Source}
We indicate whether a source language is in the speech (Sp) or text (Tx) modality, or both.

\paragraph{Target}
We indicate whether a target language is in the speech (Sp) or text (Tx) modality, or both.

\begin{table*}[!p]\centering\centering\scriptsize\begin{tabular}{clllllll}
\toprule
\textbf{Code} & \textbf{Language name} & \textbf{Family} & \textbf{Subgrouping} & \textbf{Script} & \textbf{Resource} & \textbf{Source} & \textbf{Target} \\
\midrule
afr & Afrikaans & Indo-European & Germanic & Latn & low & Sp, Tx & Tx \\
amh & Amharic & Afro-Asiatic & Semitic & Ethi & low & Sp, Tx & Tx \\
arb & Modern Standard Arabic & Afro-Asiatic & Semitic & Arab & high & Sp, Tx & Sp, Tx \\
ary & Moroccan Arabic & Afro-Asiatic & Semitic & Arab & low & Sp, Tx & Tx \\
arz & Egyptian Arabic & Afro-Asiatic & Semitic & Arab & low & Sp, Tx & Tx \\
asm & Assamese & Indo-European & Indo-Aryan & Beng & low & Sp, Tx & Tx \\
ast & Asturian & Indo-European & Italic & Latn & zero-shot & Sp & -- \\
azj & North Azerbaijani & Turkic & Common Turkic & Latn & low & Sp, Tx & Tx \\
bel & Belarusian & Indo-European & Balto-Slavic & Cyrl & high & Sp, Tx & Tx \\
ben & Bengali & Indo-European & Indo-Aryan & Beng & high & Sp, Tx & Sp, Tx \\
bos & Bosnian & Indo-European & Balto-Slavic & Latn & low & Sp, Tx & Tx \\
bul & Bulgarian & Indo-European & Balto-Slavic & Cyrl & low & Sp, Tx & Tx \\
cat & Catalan & Indo-European & Italic & Latn & high & Sp, Tx & Sp, Tx \\
ceb & Cebuano & Austronesian & Malayo-Polynesian & Latn & zero-shot & Sp, Tx & Tx \\
ces & Czech & Indo-European & Balto-Slavic & Latn & high & Sp, Tx & Sp, Tx \\
ckb & Central Kurdish & Indo-European & Iranian & Arab & low & Sp, Tx & Tx \\
cmn & Mandarin Chinese & Sino-Tibetan & Sinitic & Hans, Hant & high & Sp, Tx & Sp, Tx \\
cym & Welsh & Indo-European & Celtic & Latn & medium & Sp, Tx & Sp, Tx \\
dan & Danish & Indo-European & Germanic & Latn & medium & Sp, Tx & Sp, Tx \\
deu & German & Indo-European & Germanic & Latn & high & Sp, Tx & Sp, Tx \\
ell & Greek & Indo-European & Graeco-Phrygian & Grek & medium & Sp, Tx & Tx \\
eng & English & Indo-European & Germanic & Latn & high & Sp, Tx & Sp, Tx \\
est & Estonian & Uralic & Finnic & Latn & medium & Sp, Tx & Sp, Tx \\
eus & Basque & Basque & Basque & Latn & medium & Sp, Tx & Tx \\
fin & Finnish & Uralic & Finnic & Latn & high & Sp, Tx & Sp, Tx \\
fra & French & Indo-European & Italic & Latn & high & Sp, Tx & Sp, Tx \\
gaz & West Central Oromo & Afro-Asiatic & Cushitic & Latn & zero-shot & Sp, Tx & Tx \\
gle & Irish & Indo-European & Celtic & Latn & low & Sp, Tx & Tx \\
glg & Galician & Indo-European & Italic & Latn & low & Sp, Tx & Tx \\
guj & Gujarati & Indo-European & Indo-Aryan & Gujr & low & Sp, Tx & Tx \\
heb & Hebrew & Afro-Asiatic & Semitic & Hebr & low & Sp, Tx & Tx \\
hin & Hindi & Indo-European & Indo-Aryan & Deva & medium & Sp, Tx & Sp, Tx \\
hrv & Croatian & Indo-European & Balto-Slavic & Latn & medium & Sp, Tx & Tx \\
hun & Hungarian & Uralic & Hungarian & Latn & medium & Sp, Tx & Tx \\
hye & Armenian & Indo-European & Armenic & Armn & low & Sp, Tx & Tx \\
ibo & Igbo & Atlantic-Congo & Benue-Congo & Latn & low & Sp, Tx & Tx \\
ind & Indonesian & Austronesian & Malayo-Polynesian & Latn & medium & Sp, Tx & Sp, Tx \\
isl & Icelandic & Indo-European & Germanic & Latn & low & Sp, Tx & Tx \\
ita & Italian & Indo-European & Italic & Latn & high & Sp, Tx & Sp, Tx \\
jav & Javanese & Austronesian & Malayo-Polynesian & Latn & medium & Sp, Tx & Tx \\
jpn & Japanese & Japonic & Japanesic & Jpan & high & Sp, Tx & Sp, Tx \\
kam & Kamba & Atlantic-Congo & Benue-Congo & Latn & zero-shot & Sp & -- \\
kan & Kannada & Dravidian & South Dravidian & Knda & low & Sp, Tx & Tx \\
kat & Georgian & Kartvelian & Georgian-Zan & Geor & low & Sp, Tx & Tx \\
kaz & Kazakh & Turkic & Common Turkic & Cyrl & medium & Sp, Tx & Tx \\
kea & Kabuverdianu & Indo-European & Italic & Latn & zero-shot & Sp & -- \\
khk & Halh Mongolian & Mongolic-Khitan & Mongolic & Cyrl & low & Sp, Tx & Tx \\
khm & Khmer & Austroasiatic & Khmeric & Khmr & low & Sp, Tx & Tx \\
kir & Kyrgyz & Turkic & Common Turkic & Cyrl & low & Sp, Tx & Tx \\
kor & Korean & Koreanic & Korean & Kore & medium & Sp, Tx & Sp, Tx \\
lao & Lao & Tai-Kadai & Kam-Tai & Laoo & low & Sp, Tx & Tx \\
lit & Lithuanian & Indo-European & Balto-Slavic & Latn & low & Sp, Tx & Tx \\
ltz & Luxembourgish & Indo-European & Germanic & Latn & zero-shot & Sp & -- \\
lug & Ganda & Atlantic-Congo & Benue-Congo & Latn & medium & Sp, Tx & Tx \\
luo & Luo & Nilotic & Western Nilotic & Latn & zero-shot & Sp, Tx & Tx \\
lvs & Standard Latvian & Indo-European & Balto-Slavic & Latn & low & Sp, Tx & Tx \\
mai & Maithili & Indo-European & Indo-Aryan & Deva & zero-shot & Sp, Tx & Tx \\
mal & Malayalam & Dravidian & South Dravidian & Mlym & low & Sp, Tx & Tx \\
mar & Marathi & Indo-European & Indo-Aryan & Deva & low & Sp, Tx & Tx \\
mkd & Macedonian & Indo-European & Balto-Slavic & Cyrl & low & Sp, Tx & Tx \\
mlt & Maltese & Afro-Asiatic & Semitic & Latn & low & Sp, Tx & Sp, Tx \\
mni & Meitei & Sino-Tibetan & Kuki-Chin-Naga & Beng & zero-shot & Sp, Tx & Tx \\
mya & Burmese & Sino-Tibetan & Burmo-Qiangic & Mymr & low & Sp, Tx & Tx \\
\bottomrule
\end{tabular}
\end{table*}

\begin{table*}[!p]
\centering
\scriptsize
\begin{tabular}{llllllll}
\toprule
\textbf{Code} & \textbf{Language name} & \textbf{Family} & \textbf{Subgrouping} & \textbf{Script} & \textbf{Resource} & \textbf{Source} & \textbf{Target} \\
\midrule
nld & Dutch & Indo-European & Germanic & Latn & high & Sp, Tx & Sp, Tx \\
nno & Norwegian Nynorsk & Indo-European & Germanic & Latn & low & Sp, Tx & Tx \\
nob & Norwegian Bokmål & Indo-European & Germanic & Latn & low & Sp, Tx & Tx \\
npi & Nepali & Indo-European & Indo-Aryan & Deva & low & Sp, Tx & Tx \\
nya & Nyanja & Atlantic-Congo & Benue-Congo & Latn & low & Sp, Tx & Tx \\
oci & Occitan & Indo-European & Italic & Latn & zero-shot & Sp & -- \\
ory & Odia & Indo-European & Indo-Aryan & Orya & low & Sp, Tx & Tx \\
pan & Punjabi & Indo-European & Indo-Aryan & Guru & low & Sp, Tx & Tx \\
pbt & Southern Pashto & Indo-European & Iranian & Arab & medium & Sp, Tx & Tx \\
pes & Western Persian & Indo-European & Iranian & Arab & low & Sp, Tx & Sp, Tx \\
pol & Polish & Indo-European & Balto-Slavic & Latn & high & Sp, Tx & Sp, Tx \\
por & Portuguese & Indo-European & Italic & Latn & medium & Sp, Tx & Sp, Tx \\
ron & Romanian & Indo-European & Italic & Latn & high & Sp, Tx & Sp, Tx \\
rus & Russian & Indo-European & Balto-Slavic & Cyrl & medium & Sp, Tx & Sp, Tx \\
slk & Slovak & Indo-European & Balto-Slavic & Latn & medium & Sp, Tx & Sp, Tx \\
slv & Slovenian & Indo-European & Balto-Slavic & Latn & low & Sp, Tx & Tx \\
sna & Shona & Atlantic-Congo & Benue-Congo & Latn & zero-shot & Sp, Tx & Tx \\
snd & Sindhi & Indo-European & Indo-Aryan & Arab & zero-shot & Sp, Tx & Tx \\
som & Somali & Afro-Asiatic & Cushitic & Latn & low & Sp, Tx & Tx \\
spa & Spanish & Indo-European & Italic & Latn & high & Sp, Tx & Sp, Tx \\
srp & Serbian & Indo-European & Balto-Slavic & Cyrl & low & Sp, Tx & Tx \\
swe & Swedish & Indo-European & Germanic & Latn & low & Sp, Tx & Sp, Tx \\
swh & Swahili & Atlantic-Congo & Benue-Congo & Latn & medium & Sp, Tx & Sp, Tx \\
tam & Tamil & Dravidian & South Dravidian & Taml & medium & Sp, Tx & Tx \\
tel & Telugu & Dravidian & South Dravidian & Telu & medium & Sp, Tx & Sp, Tx \\
tgk & Tajik & Indo-European & Iranian & Cyrl & low & Sp, Tx & Tx \\
tgl & Tagalog & Austronesian & Malayo-Polynesian & Latn & medium & Sp, Tx & Sp, Tx \\
tha & Thai & Tai-Kadai & Kam-Tai & Thai & medium & Sp, Tx & Sp, Tx \\
tur & Turkish & Turkic & Common Turkic & Latn & medium & Sp, Tx & Sp, Tx \\
ukr & Ukrainian & Indo-European & Balto-Slavic & Cyrl & medium & Sp, Tx & Sp, Tx \\
urd & Urdu & Indo-European & Indo-Aryan & Arab & medium & Sp, Tx & Sp, Tx \\
uzn & Northern Uzbek & Turkic & Common Turkic & Latn & medium & Sp, Tx & Sp, Tx \\
vie & Vietnamese & Austroasiatic & Vietic & Latn & medium & Sp, Tx & Sp, Tx \\
xho & Xhosa & Atlantic-Congo & Benue-Congo & Latn & zero-shot & Sp & -- \\
yor & Yoruba & Atlantic-Congo & Benue-Congo & Latn & low & Sp, Tx & Tx \\
yue & Cantonese & Sino-Tibetan & Sinitic & Hant & low & Sp, Tx & Tx \\
zlm & Colloquial Malay & Austronesian & Malayo-Polynesian & Latn & low & Sp & -- \\
zsm & Standard Malay & Austronesian & Malayo-Polynesian & Latn & low & Tx & Tx \\
zul & Zulu & Atlantic-Congo & Benue-Congo & Latn & low & Sp, Tx & Tx \\
\bottomrule
\end{tabular}
\caption{
\label{tab:all_languages}
\textbf{\mfourt languages.} We display the language code, name, family, subgroup, and script, as well as the speech resource level and whether the language is supported as a source or a target in the speech and/or text modalities. Zero-shot here refers to \st or \sst tasks with the language in question as source.
}
\end{table*}

\FloatBarrier

\section{\mineddata: Automatically Creating Aligned Data for Speech}
\label{sec:appendix:data}
Developing an effective multilingual and multimodal translation system like \mfourt requires sizable resources across many languages and modalities. Some human-labeled resources for translation are freely available, albeit often limited to a small set of languages or in very specific domains. Well-known examples are parallel text collections such as Europarl \citep{koehn:2005:europarl} and the United Nations Corpus \citep{ziemski-etal-2016-united}. A few human-created collections also involve the speech modality, like CoVoST \citep{wang-etal-2020-covost,wang2021covost} and mTEDx \citep{salesky2021mtedx}. Yet no open dataset currently matches the size of those used in initiatives like Whisper \citep{whisper} or USM \citep{google_usm}, which proved to unlock unprecedented performance. 

Parallel data mining emerges as an alternative to using closed data, both in terms of language coverage and corpus size. The dominant approach today is to encode sentences from various languages and modalities into a joint fixed-size embedding space and to find parallel instances based on a similarity metric. Mining is then performed by pairwise comparison over massive monolingual corpora, where sentences with similarity above a certain threshold are considered mutual translations \citep{Schwenk:2018:acl_mine,Artetxe:2019:mine_acl}. This approach was first introduced using the multilingual \laser space \citep{Artetxe:2019:tacl_massive_ml}. Teacher-student training was then used to scale this approach to 200 languages \citep{laser3:2022:emnlp,nllb2022} and subsequently, the speech modality \citep{Duquenne:2021:neurips,speechmatrix-acl23}.

\begin{figure}[t]
    \centering
    \includegraphics[width=1.05\textwidth]{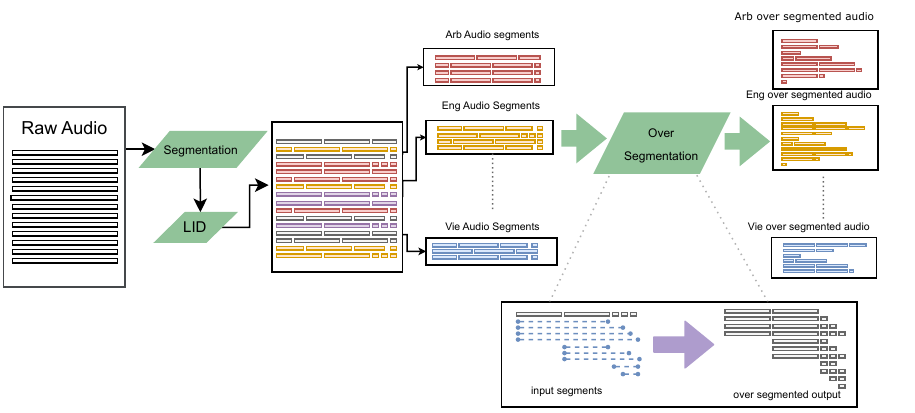}
    \caption{Workflow of speech processing.}
    \label{fig:appendix:data:workflow}
\end{figure}

In this section, we describe how we employed parallel data mining to create \mineddata: the largest open dataset for multimodal translation to date, totaling \totalminedhours hours. The overall workflow is summarized in Figure~\ref{fig:appendix:data:workflow}, and builds on the approach deployed in \SpeechMatrix \citep{speechmatrix-acl23}. Starting with a large collection of raw audio, we 
chunked files into overlapping segments
and applied speech Language Identification (LID).
On the text side, we used the same sentence-segmented dataset drawn from NLLB \citep{nllb2022}. Speech and text corpora were then projected into a common embedding space, in which mining was performed to identify translation pairs with optimal segmentation.
Several improvements over the original \SpeechMatrix pipeline are introduced:
\begin{itemize}
    \itemsep 0em 
    \item an improved and extended speech language identification (LID) model,
    \item a novel multimodal embedding space,
    \item increased coverage from 17 to \NbLangsMined languages,
    \item increased raw audio amount, totaling \DataRawAudioText hours.
\end{itemize}

 In the current version, mining was focused on \NbLangsMined target languages of the \mfourt system. Scaling to all \NbLangsTotal languages will be explored in future iterations of our work. 

\subsection{Speech-language identification}
\label{sec:appendix:data:lid}
Language identification (LID) of raw audio data is a critical component of our workflow. Incorrectly labeling speech at this stage can prevent high-quality audio segments from being aligned or, worse, add noise to the resulting paired sets. This can adversely affect the performance of the downstream translation system.

While numerous off-the-shelf LID models exist, none could cover our target list of \NbLangsTotal languages.\footnote{MMS \citep{mms} has recently been released and covers them all, but it was not available when this project started} Therefore, we trained our own model, following the ECAPA-TDNN architecture introduced in \citep{desplanques-ecapa-tdnn-2020}, for which an open-source model trained on VoxLingua107 \citep{valk2021slt} is available.
The new model adds support for several new languages, including Moroccan Arabic, Egyptian Arabic, Central Kurdish, West Central Oromo, Irish,
Igbo, Kyrgyz, Ganda,
Maithili, Meitei, Nyanja,
Odia,
Cantonese, and Zulu.

\subsubsection{Training}

\paragraph{Baseline} We first retrained a system from scratch on VoxLingua107 data to reproduce a baseline. This system, dubbed \textit{VL107 baseline}, achieved a classification error rate of 5.25\% on the development set of VoxLingua107 at epoch 30. Comparatively, the open-sourced model available on HuggingFace,\footnote{\url{https://huggingface.co/TalTechNLP/voxlingua107-epaca-tdnn}} referred to as \textit{VL107 HF}, yields an error rate of 7\%.

\paragraph{Experimental setup} With our training pipeline validated, we finally trained our own model for 40 epochs. This required about 172 hours on 8 GPUs. A total of 17k hours of speech were used, with an average of 171 hours per language, ranging from 1 to 600 hours. The test corpus covers our 100 languages of interest and is composed of the FLEURS test set, the VoxLingua107 development set, and extra test data extracted from VAANI,\footnote{\url{http://vaani.iisc.ac.in}} IIITH \citep{kumarvuddagiri18b_sltu}
and KENCORPUS\footnote{\url{https://dataverse.harvard.edu/dataset.xhtml?persistentId=doi:10.7910/DVN/6N5V1K}} \citep{wanjawa2022kencorpus}.

\begin{table}[htb]
    \centering
    \small
    \begin{tabular}{lrrrr}
        \toprule
        & \MC{2}{c}{{\bf Overall}} & \MC{2}{c}{{\bf Intersection} } \\\cmidrule(r){2-3}\cmidrule(l){4-5}
        & \makecell[c]{$\uparrow$F1-micro\\{\it (n=100)}}& \makecell [c]{$\uparrow$F1-macro\\{\it (n=100)}} & \makecell[c]{$\uparrow$F1-micro\\{\it (n=79)}}  & \makecell[c]{$\uparrow$F1-macro \\{\it (n=79)}}\\
        \midrule
        VL107 HF        & 82.3\% & - & 94.1\% & 92.6\% \\
        VL107 baseline  & 82.5\% & - & 94.4\% & 93.0\% \\
        LID100          & 86.0\% & 81.9\% & 92.9\% & 91.1\% \\
        \bottomrule
    \end{tabular}
    \caption{F1 micro and macro average for the considered LID systems over all \mfourt languages and the intersection of supported languages across models. Dashes are used for models that do not support the full 100 scope.}
    \label{tbl:appendix:data:lid:results}
\end{table}

\paragraph{Results} The F1 scores on the test data for all models are presented in Table~\ref{tbl:appendix:data:lid:results}.
The results are given for the 100 \mfourt languages, and the 79 languages in common with VoxLingua107. We can see that training on the additional languages slightly decreases the overall performance for the common set of languages, which is a direct consequence of the presence of a higher number of close languages. For example, Zulu (zul) is very often confused with Nyanja (nya), Igbo (ibo) with Yoruba (yor), and Modern Standard Arabic (arb) with Moroccan Arabic (ary) and Egyptian Arabic (arz).
Our model improves classification (F1 difference greater than 5\%) on 17 languages with an average gain of 14.6\%, not counting the newly covered languages, while decreasing classification for 12 (with an average loss of 9.8\%).

\subsubsection{Filtering}
\label{sec:appendix:data:lid:threshold}

While it is important to retrieve the maximum amount of data for mining, we must also ensure high quality in LID labeling. 
Depending on the quantity of data available for a particular language, it may be useful to filter it to retain higher-quality data. We thus estimated the Gaussian distribution of the LID scores per language for correct and incorrect classifications on the development corpus. We selected a threshold per language such that $p(correct|score) > p(incorrect|score)$. 
By rejecting 8\% of the data, we were able to further increase the F1 measure by almost 3\%.

\begin{table}[htb]
    \centering
    \small
    \begin{tabular}{lrr}
        \toprule
        &{\bf  $\uparrow$F1 micro} &  {\bf $\uparrow$Coverage} \\
        \midrule
        LID100 & 86.0\%  & 100\%\\
        +filtering & 89.5\% & 92.1\% \\
        \bottomrule
    \end{tabular}
    \caption{F1 micro average and coverage across 100 languages for the LID100 system with and without filtering.}
    \label{tbl:appendix:data:lid:results_filtering}
\end{table}

\subsection{Gathering raw audio and text data at scale}
\label{sec:appendix:data:m2c2}

\paragraph{Text pre-processing} On the text side, we rely entirely on the same dataset deployed in NLLB \citep{nllb2022}. The same data sources, cleaning, and filtering steps are used and run at scale with our \stopes library.

\paragraph{Audio pre-processing}
We start with \DataRawAudioText hours of raw audio originating from a publicly available repository of crawled web data. Table~\ref{tab:appendix:data:mined_details} provides statistics on the amount of raw audio for each language. Approximately 1 million hours in this collection are in English. 
We then applied a series of pre-processing steps to curate and improve the overall speech quality. Firstly, we deduplicated the audio file URLs found in the repository, downloaded the audio files, and resampled at 16KHz. Subsequently, we filtered out the non-speech data with a bespoke audio event detection (AED) model.

\paragraph{Audio segmentation}
To perform \st or \sst mining, it is desirable to split audio files into smaller chunks that map as closely as possible to self-contained sentences, equivalent to sentences in a text corpus. However, genuine semantic segmentation in speech is an open-ended problem--pauses can be an integral part of a message and can naturally occur differently across languages. For mining purposes, it is impossible to prejudge what specific segments can maximize the overall quality of the mined pairs.

We thus followed the over-segmentation approach drawn from \citep{Duquenne:2021:neurips} (as depicted in Figure~\ref{fig:appendix:data:workflow}). 
First, we used an open Voice Activity Detection (VAD) model \citep{SileroVAD} to split audio files into shorter segments. Subsequently, our speech LID model was used on each file.
Finally, we created several possible overlapping splits of each segment and left the choice of the optimal split to the mining algorithm described in the next section.
This over-segmentation strategy roughly octuples the number of potential segments considered.

\subsection{Speech mining}

\begin{figure}[!b]
    \centering
    \includegraphics[width=0.9\textwidth]{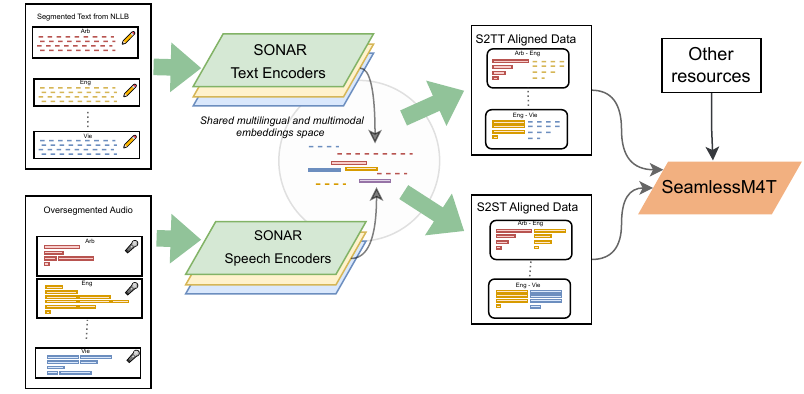}
    \caption{Workflow of the \sonar encoding and mining processes.
    }
    \label{fig:appendix:data:workflow-mining}
\end{figure}

The overall workflow of our mining process is shown in Figure~\ref{fig:appendix:data:workflow-mining}. 
First, we trained encoders for text (Section~\ref{sec:appendix:data:sonar}) and speech (Section~\ref{sec:appendix:data:encoders}). These are then used to project both modalities into a joint embedding space. We then mined speech segments against text sentences or speech segments in other languages to create large amounts of \st and \sst pairs. These corpora are subsequently combined with other resources to train the \mfourt model.

\subsubsection{\sonar text embedding space}
\label{sec:appendix:data:sonar}

\begin{figure}[!b]
    \centering
    \includegraphics[width=0.9\textwidth]{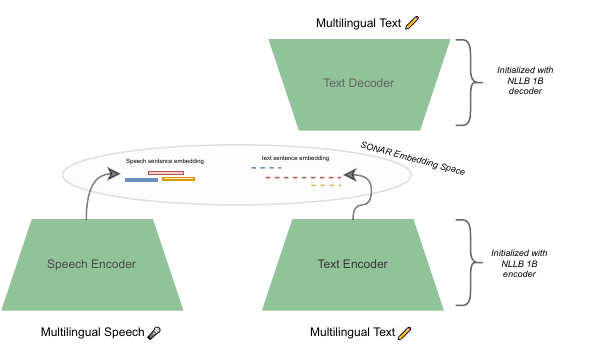}
    \caption{SONAR architecture.}
    \label{fig:appendix:data:sonar}
\end{figure}

\paragraph{Architecture and training setup}
We use a novel sentence embedding space developed by \citet{Duquenne:2023:sonar_arxiv}, named \textbf{S}entence-level multim\textbf{O}dal and la\textbf{N}guage-\textbf{A}gnostic \textbf{R}epresentations—in short, \sonar. \sonar substantially outperforms the previous \laser space. It follows the same two-step approach: we first trained a text embedding space and then relied on a teacher-student training strategy to extend it to the speech modality. Similarly to \laser, the initial text \sonar space uses an encoder-decoder architecture, but is based on the \nllbsmall model, capable of translating across 200 languages \citep{nllb2022}. The intermediate representation was replaced with a fixed-size vector using mean-pooling (i.e., the decoder thus attends to a single vector). This architecture is fine-tuned using all of NLLB's \mt training data, and several training objectives were explored. A detailed ablation study can be found in \citet{Duquenne:2023:sonar_arxiv}. This yields a powerful, massively multilingual sentence representation that can be decoded into all 200 languages of the NLLB project. Figure~\ref{fig:appendix:data:sonar} provides an illustration of the \sonar architecture and 
Table \ref{tab:appendix:data:sonar_bleu} summarizes the translation evaluation of the \sonar framework.%

\paragraph{Evaluation for mining}
\begin{table}[!t]
    \centering
    \small
    \begin{tabular}{l*{5}{c}}
        \toprule
        & \multicolumn{2}{c}{ {\bf $\uparrow$spBLEU}} &
        \multicolumn{2}{c}{ {\bf $\uparrow$COMET}  } \\\cmidrule(r){2-3}\cmidrule(l){4-5}
       {\bf Model}  & \makecell[c]{\xeng\\ {\it (n=200)}} & \makecell[c]{\engx\\{\it (n=200)}} & \makecell[c]{\xeng\\ {\it (n=89)}}  & \makecell[c]{\engx \\{\it (n=89)}}\\ %
        \midrule
        \sonar  &32.7 & 21.6 & 85.9 & 84.2 \\
        NLLB-1.3B (MT topline)
         & 35.2 & 24.9 & 86.5 & 85.2 \\ %
        \bottomrule
    \end{tabular}
    \caption{Average performance on \flores devtest set over the 200 NLLB languages and 89 languages supported by COMET: translation spBLEU and COMET scores, auto-encoding spBLEU. } %
    \label{tab:appendix:data:sonar_bleu}
\end{table}
On pure translation performance, we observe that the fixed-size representation bottleneck leads to a 7\% and 13\% decrease in BLEU score when translating into English (35.2$\ra$32.7) and out of English, respectively (24.9$\ra$21.6).
This is a rather interesting result, given that the use of attention is commonly considered mandatory to achieve reasonable performance.

On mining performance, we rely on the multilingual similarity search \xsim metric, which measures the percentage of sentences in the \flores dataset which are not correctly aligned when searching for the closest vector in the embedding space.
The improved version \xsimpp \citep{xsimpp:2023:acl} added challenging English sentences on the target side. Both of these metrics are a good proxy to the actual \mt mining task while being much faster to compute.

As summarized in Table~\ref{tab:appendix:data:sonar_xsim}, \sonar substantially outperforms other popular multilingual sentence representations like \laserthree \citep{laser3:2022:emnlp} or LaBSE \citep{google:2020:labse}.

\begin{table}[htb]
    \centering
    \small
    \begin{tabular}{lcccc}
        \toprule
        & \MC{2}{c}{ {\bf Overall}} & \MC{2}{c}{ {\bf Intersection} } \\\cmidrule(r){2-3}\cmidrule(l){4-5}
        & \makecell[c]{$\downarrow$\xsim \\{\it (n=200)}} & \makecell[c]{$\downarrow$\xsimpp\\{\it (n=200)}} & \makecell[c]{$\downarrow$\xsim \\{\it  (n=\nbLangsLaserLabse)}} &\makecell[c]{$\downarrow$\xsimpp \\{\it  (n=\nbLangsLaserLabse)}}\\
        \midrule
        \sonar  & 1.4 & 15.2 & 0.1 & 9.3 \\
        \laserthree & 5.1 & 36.4 & 1.1 & 27.5 \\
        LaBSE & 10.7 & 36.1 & 1.5 & 15.4 \\
        \bottomrule
    \end{tabular}
    \caption{Comparison of similarity search results (error rates) on all 200 \flores languages, and limited to the intersection of $\nbLangsLaserLabse$ languages on which each model has been trained on. %
    }
    \label{tab:appendix:data:sonar_xsim}
\end{table}

\subsubsection{Training speech encoders}
\label{sec:appendix:data:encoders}

\paragraph{Architecture and training setup}
As a second step and following \citep{Duquenne:2021:neurips}, the new \sonar text embedding space is extended to the speech modality through teacher-student training. %
In that work, a fixed-size speech representation was obtained by taking the BOS output of a pretrained \xlsr model \citep{babu2021xls}. This model was then fine-tuned to maximize the cosine loss between this pooled speech representation and sentence embeddings in the same languages (ASR transcriptions) or in English (speech translations).
We improved this initial recipe by doing the following:
\begin{itemize}
    \itemsep 0em 
    \item MSE loss instead of a cosine loss was used. This enables us to use the \sonar text decoder on speech input,
    \item \wvbert speech front-end instead of \xlsr. \wvbert was optimized on \wvbertlangs languages (see \Cref{sec:modeling:x2t:w2vbert} for details),
    \item Attention-pooling. Instead of the usual pooling methods (i.e., mean or max-pooling), we implemented a 3-layer sequence-to-sequence model to convert the variable length sequence of \wvbert to a fixed size vector,
    \item Training on human-performed ASR transcriptions only. We collected at least 100 hours of ASR transcriptions for most of the languages (see Table~\ref{tab:appendix:data:mined_details} column \textit{``train''}) and trained the speech encoders exclusively on them, %
    \item Following \citep{laser3:2022:emnlp,nllb2022}, we grouped languages by linguistic families (i.e., Germanic or Indian languages) and trained them together in one speech encoder. Alternative language groupings, which might consider the acoustic characteristics of each language, are left open for future research.
\end{itemize}

\begin{table*}[p]
    \small
    
\begin{tabular}[t]{lrrrr*{3}{r}}
\toprule
\multirow{2}{*}{\bf ISO} & \bf Raw & \bf Train & \multicolumn{2}{c}{\bf \xeng  ($\uparrow$BLEU) } & \multicolumn{3}{c}{\bf Mined audio [h]} \\\cmidrule(r){2-3}\cmidrule(lr){4-5}\cmidrule(l){6-8}
 & \textbf{audio [h]} & \bf ASR [h] & \textbf{Ours} & \bf Whisper & \textbf{Sen2Txx} & \textbf{Sxx2Ten} & \textbf{Sxx2Sen} \\ 
\midrule
\textbf{arb}  & {106755} & {822} & {28.7} & {25.5} & {1568} & {8072} & {776}\\
\textbf{ben}  & {7012} & {335} & {18.9} & {13.2} & {606} & {1345} & {263}\\
\textbf{cat}  & {43531} & {1738} & {35.1} & {34.2} & {1570} & {4411} & {354}\\
\textbf{ces}  & {41318} & {181} & {29.2} & {27.8} & {1454} & {6905} & {602}\\
\textbf{cmn}  & {79772} & {9320} & {16.2} & {18.4} & {5440} & {18760} & {1570}\\
\textbf{cym}  & {24161} & {99} & {14.5} & {13.0} & {--} & {4411} & {278}\\
\textbf{dan}  & {34300} & {115} & {31.9} & {32.7} & {2499} & {6041} & {583}\\
\textbf{deu}  & {490604} & {3329} & {32.7} & {34.6} & {91715} & {17634} & {1921}\\
\textbf{est}  & {12691} & {131} & {23.8} & {18.7} & {1022} & {3346} & {607}\\
\textbf{fin}  & {32858} & {184} & {22.2} & {22.1} & {651} & {6086} & {526}\\
\textbf{fra}  & {282179} & {2057} & {31.2} & {32.2} & {21523} & {17380} & {3337}\\
\textbf{hin}  & {15118} & {150} & {19.2} & {22.0} & {1041} & {2977} & {530}\\
\textbf{ind}  & {11559} & {269} & {26.5} & {29.1} & {1938} & {2658} & {510}\\
\textbf{ita}  & {79480} & {588} & {25.3} & {23.6} & {4378} & {6508} & {817}\\
\textbf{jpn}  & {75863} & {17319} & {17.4} & {18.9} & {1973} & {21287} & {1141}\\
\textbf{kan}  & {1451} & {114} & {20.0} & {11.6} & {--} & {232} & {198}\\
\textbf{kor}  & {37854} & {316} & {15.0} & {21.3} & {--} & {8657} & {640}\\
\textbf{mlt}  & {2122} & {106} & {23.2} & {13.5} & {131} & {130} & {60}\\
\textbf{nld}  & {93933} & {1723} & {25.5} & {24.0} & {3720} & {6859} & {1210}\\
\textbf{pes}  & {43788} & {386} & {22.2} & {19.6} & {--} & {7122} & {693}\\
\textbf{pol}  & {53662} & {304} & {21.1} & {22.3} & {1324} & {9389} & {757}\\
\textbf{por}  & {141931} & {269} & {35.4} & {38.1} & {4853} & {8696} & {928}\\
\textbf{ron}  & {18719} & {135} & {32.1} & {31.5} & {2770} & {2878} & {716}\\
\textbf{rus}  & {103906} & {259} & {25.4} & {27.8} & {11296} & {13509} & {1252}\\
\textbf{slk}  & {16954} & {102} & {29.5} & {26.1} & {1267} & {3785} & {491}\\
\textbf{spa}  & {324086} & {1511} & {24.3} & {23.3} & {27778} & {17388} & {2727}\\
\textbf{swe}  & {125195} & {144} & {33.4} & {37.02} & {3438} & {2620} & {484}\\
\textbf{swh}  & {18393} & {361} & {22.6} & { 7.2} & {690} & {2620} & {484}\\
\textbf{tam}  & {100331} & {245} & {14.3} & { 9.2} & {--} & {1664} & {867}\\
\textbf{tel}  & {3303} & {84} & {15.8} & {12.5} & {--} & {985} & {536}\\
\textbf{tgl}  & {4497} & {108} & {13.3} & {24.4} & {--} & {633} & {266}\\
\textbf{tha}  & {13421} & {195} & {15.3} & {16.1} & {2577} & {3563} & {542}\\
\textbf{tur}  & {23275} & {174} & {21.0} & {26.6} & {1417} & {6545} & {426}\\
\textbf{ukr}  & {6396} & {105} & {27.9} & {29.4} & {1220} & {1717} & {392}\\
\textbf{urd}  & {16882} & {185} & {17.6} & {17.2} & {773} & {3416} & {652}\\
\textbf{uzn}  & {8105} & {115} & {17.9} & { 6.0} & {475} & {1846} & {157}\\
\textbf{vie}  & {34336} & {194} & {17.8} & {20.4} & {1689} & {7692} & {868}\\
\midrule \\
\multicolumn{1}{l}{\textbf{Total/avr}} & {2529741} & {43772} & {23.3} & {22.5} & {202796} & {239767} & {29161}
\\
\bottomrule
\end{tabular}

    \normalsize
    \caption{Statistics on speech encoders and amount of mined data. Sen2Txx, Sxx2Ten, and SxxSen correspond to English speech paired with foreign text, foreign speech paired with English Text, and foreign Speech paired with English speech, respectively. Dashes are unmined directions. We provide the amount of raw audio data for mining and the amount of human-provided ASR transcripts to train the speech encoders. The speech encoders are evaluated for \st{} using BLEU on the \fleurs test set. Our model performs zero-shot \st{}. Finally, the last three columns provide the amount of mined data.}
    \label{tab:appendix:data:mined_details}
\end{table*}

\paragraph{Evaluation of speech encoders}

The trained speech encoders are to be used in \st and \sst mining, and the resulting paired data is to be fed into the \mfourt system (see section~\ref{sec:appendix:modeling}). Consequently, an ideal evaluation would consist of testing various iterations of each speech encoder by using them in an end-to-end loop: performing mining, then training a \st or \sst translation system on the mined data, and potentially comparing different thresholds of the \sonar score. Unfortunately, this is a very compute-intensive recipe.

Instead, given that the \sonar embedding space comes with a text decoder, we chose to evaluate the individual speech encoders on a \st task. That is, following \citep{tmodules,modulars2t}, we decoded foreign speech embeddings into English texts.
Results are summarized in Table~\ref{tab:appendix:data:mined_details}, column \textit{``X-eng BLEU''}. For comparison, we also provide the performance of \whisperlarge \citep{whisper}.
It is important to emphasize that the \sonar speech encoders were trained on ASR transcriptions only and the \sonar text decoder has never been exposed to any speech input. Therefore, the reported results correspond to fully zero-shot speech translation.

Despite the zero-shot scenario, the \sonar speech encoders compare favorably to a model like \whisperlarge, which was trained on a massive amount of translated audio. Gaps in BLEU points can be observed in some high resource languages such as German, Russian or Portuguese,  %
However, zero-shot speech translation with our speech encoders outperforms \whisperlarge on several low-resource languages -- particularly for Swahili and several South Asian languages like Bengali, Kannada, Telugu, and Tamil.

\subsubsection{Speech mining}

\paragraph{Margin setting} Mining was performed using a margin criterion with our \stopes data processing library\footnote{\url{https://github.com/facebookresearch/stopes}} \citep{andrews-etal-2022-stopes}.
The overall processing is identical to that developed for \mt mining in NLLB \citep{nllb2022}.
We performed so-called \textit{global mining}, where all speech segments in one language are compared to all speech segments in another language. \textit{Local mining}, on the contrary, would try to leverage knowledge on longer speech chunks that are likely to contain many parallel segments. A typical example would be documentation on an international event in multiple languages. Such high-level information is very difficult to obtain at scale.

First, the embeddings for all speech segments and text sentences are calculated. These are then indexed with the FAISS library \citep{johnson2019billion}, enabling efficient large-scale similarity search on GPUs. Finally, nearest neighbors to all elements in both directions are retrieved, and margin scores are computed following the formula introduced in \citep{Artetxe:2019:mine_acl}:

\begin{equation}
    \text{score}(x, y) = \text{margin}\left( cos(x,y), 
    \sum_{z \in NN_k(x)} \frac{cos(x, z)}{2k} + \sum_{v \in NN_k(y)} \frac{cos(y, v)}{2k} \right)
\end{equation}

\noindent where $x$ and $y$ are the source and target sentences, and $NN_k(x)$ denotes the $k$ nearest neighbors of $x$ in the other language. We set $k$ to 16.

In past work, a threshold of 1.06 on the margin score was used for bitext mining based on \laser embeddings \citep{schwenk:2021:acl_ccmatrix,nllb2022}. The \sonar space, however, displayed different dynamics and the optimal threshold was adapted accordingly. Since full end-to-end evaluation with \st or \sst training is too compute-intensive, we set the new threshold at 1.15 after some human inspection. The statistics reported in Table~\ref{tab:appendix:data:mined_details} are based on this threshold.

\paragraph{Mined dataset} We performed mining of speech in foreign languages against English texts (column Sxx2Ten in Table~\ref{tab:appendix:data:mined_details}) and English speech (column Sxx2Sen in Table~\ref{tab:appendix:data:mined_details}).
Given the sheer size of our raw English speech (1 million hours) and foreign text collections (often more than 1 billion sentences), we carried out this operation only for some languages (column Sen2Txx in Table~\ref{tab:appendix:data:mined_details}). Other directions are left for future work.

Except for Maltese, for which we had access only to a small amount of raw audio, we were able to mine more than 100 hours of speech alignments with English speech for all languages. The alignments with English texts reached a thousand hours for most languages and exceeded ten thousand hours for six (i.e., German, French, Spanish, Japanese, Russian, and Mandarin Chinese).
Overall, \mineddata covers \NbLangsMined languages and a total of \totalminedhours hours:
\begin{itemize}
    \item English speech to non-English text (Sen2Txx)—approximately 200,000 hours
    \item Non-English speech to English text (Sxx2Ten)—approximately 240,000 hours
    \item Non-English speech to English speech (Sxx2Sen)—approximately 29,000 hours
\end{itemize}
Adding such huge amounts of data to train a massively multilingual \sst{} translation system represents a substantial computational challenge. As described in Section~\ref{sec:appendix:modeling}, not all of this data was used for modeling, but only a subset with the highest \sonar alignment scores.
Since our mined data can help support many different use cases, we are open-sourcing the meta-data for the full amount\footnote{available at \url{https://github.com/facebookresearch/seamless_communication}} (i.e., up to a \sonar threshold of 1.15), to allow the community to rebuild \mineddata and use it for their own purposes. The optimal threshold can thus be tuned based on the task, balancing dataset size and alignment quality. Our mining code is also open-sourced in the \stopes library.

\subsection{Related work}

\subsubsection{Speech LID}
Spoken language identification has been traditionally approached in a two-stage workflow: a classifier is trained on top of conventional representations like the i-vector or x-vector, extracted from the raw audio signal \citep{dehak2011language, snyder2018spoken}. The same idea has been revisited in end-to-end, integrated neural architectures \citep{cai2019utterance, miao2019new, wan2019tuplemax}. These approaches typically fall short as the input audio goes shorter, which can be an issue with speech recordings involving multiple speakers talking to each other in turn. New methods were developed to tackle this very problem. \citet{lopez2014automatic} show that a simple feed-forward network can outperform i-vectors on this task. More complex architectures such as convolutional neural networks or Bi-LSTMs prove to be more efficient in capturing information from the speech input \citep{lozano2015end, fernando2017bidirectional}. Some other approaches try to bridge the gap with models focused on longer segments through teacher-student training \citep{shen2018feature, shen2019interactive}.

Recent initiatives aimed at increasing language coverage to go beyond a handful of conventionally very high-resource languages. The ECAPA-TDNN architecture introduced in \citep{desplanques-ecapa-tdnn-2020} has proven effective to distinguish between the 107 languages of Voxlingua107 \citep{valk2021slt}. The \xlsr pretrained model \citep{babu2021xls} is also fine-tuned on a language identification task using the same dataset. \whisperlarge is another popular model that can perform this task for 99 languages \citep{whisper}. Very recently, the MMS project further broadened language support to 4,000 spoken languages \citep{mms}.

\subsubsection{Speech segmentation}
To achieve sentence-like speech segments, a commonly employed method is pause-based segmentation using Voice Activity Detection (VAD). This approach is widely utilized in various applications, including speech mining, ASR, and speech translation. In this work, we adopted the over-segmentation strategy proposed by \citet{Duquenne:2021:neurips} on top of the segments obtained through VAD segmentation. While this over-segmentation significantly improves the recall of the mining process, it does come with certain drawbacks. Specifically, it leads to a substantial increase (8x) in the number of segments, introducing noise in the embedding space, and raising the computational demand for the mining process.
Pause-based segments may not align with semantically coherent sentences; in fact, they tend to be too short because speaker pauses can extend beyond sentence boundaries. Consequently, for speech translation, researchers have put forward more sophisticated segmentation strategies with the potential to deliver higher-quality speech translation results. \citet{GallegoTEFC21} used a pretrained wav2vec 2.0 instead of VAD to detect speech segments. \citet{DAC_PotapczykP20} proposed a divide-and-conquer (DAC) algorithm that iteratively operates on top of VAD longest detected pauses until all segments become below a max-segment length parameter. \citet{Hybrid_segmentation_GaidoNCT21} further builds upon this through a hybrid approach. SHAS~\citep{tsiamas22_interspeech} train a classifier on top of wav2vec 2.0 using optimal segmentation from a manually segmented corpus. Similar to \citet{DAC_PotapczykP20}, it then applies a DAC algorithm on the splitting probabilities of the network to obtain final segmentation decisions. This approach demonstrated significant gains over simple pause-based segmentation and other baselines in speech-to-text translation tasks. 
These segmentation methods could be promising for speech mining, suggesting exciting avenues for future research.

\subsubsection{Multilingual and multimodal representations}

Several works have studied how to learn multilingual sentence representations. Well known approaches are \laser \citep{Artetxe:2019:tacl_massive_ml}, LaBSE \citep{google:2020:labse}, or \citep{yang2019improving,samantar:2022:tacl}.
While \laser was trained with an MT translation objective, a decoder compatible with the \laser embedding space is not freely available. To the best of our knowledge, \sonar is the first sentence embedding space for which an efficient and multilingual decoder is available.
Another direction of research is to first train an English sentence representation (e.g., sentence-BERT \citep{reimers2019sentence}) %
and in a second step, extend it to more languages using teacher-student training \citep{Reimers:2020:emnlp_ts}. The same approach was used to extend \laser to 200 languages, named \laserthree \citep{laser3:2022:emnlp}.

Learning unsupervised representations of speech is the focus of several works, whether involving monolingual \citep{baevski:2022:arxiv_data2vec} or multilingual speech \citep{babu2021xls,hsu2021hubert,chung2021w2v}. Examples of joint text and speech pre-trained models are mSLAM \citep{mslam} and Mu$^2$SLAM~\citep{pmlr-v202-cheng23e}.
\citet{Duquenne:2021:neurips} were the first to introduce fixed-size text and speech representations that can be used to perform multimodal mining, followed by \citep{khurana2022samu}

\subsubsection{Speech mining}

The proof of concept of a joint text/speech representation that can be used to perform text/speech or speech/speech mining was presented by \citet{Duquenne:2021:neurips}. 
In follow-up work, this approach was used to align speech in 17 languages in the VoxPopuli corpus to give rise to the \SpeechMatrix corpus \citep{speechmatrix-acl23}. The authors mined for parallel speech segments in all 136 possible combinations of languages, yielding a total of 418 thousand hours of speech-to-speech alignments, out of which about 46 thousand hours are aligned with English. \SpeechMatrix is a large corpus, but the domain is rather limited since the raw audio of the VoxPopuli corpus is derived from European Parliament speeches. The corpus \SpeechMatrix is freely available.
\citet{khurana2022samu} use a joint text/speech embedding space, dubbed \textsc{Samu-Xlsr}, to evaluate the recall of text and speech retrieval in the corpora \covost, \mbox{MUST-C}, and \mbox{MTEDx}.

There are several works that indirectly create speech-to-speech corpora. One direction of research is to perform speech synthesis on corpora aligned at the text level, (e.g., the CVSS corpus \citep{cvss} which is based on the \covost speech-to-text translation corpus).

\FloatBarrier
\newpage

\section{\mfourt Models}
\label{sec:appendix:modeling}

Direct speech-to-text translation models 
have made significant progress in recent years \citep{Berard2016ListenAT,Weiss2017SequencetoSequenceMC,di2019must,agrawal-etal-2023-findings}, and achieved parity with cascaded models on academic benchmarks under specific situations (e.g., constrained data, in-domain settings, specific language pairs, etc.). However, with the arrival of massively multilingual translation models~\citep{nllb2022, siddhant2022towards, m2m100} and weakly supervised \asr models~\citep{whisper, google_usm, mms}, which leverage massive quantities of labeled data for training large foundation models, these comparisons have become outdated. To put it simply,  direct models now lag significantly behind strong cascaded models.

\begin{figure}[!htb]
\centering
\includegraphics[width=.8\linewidth]{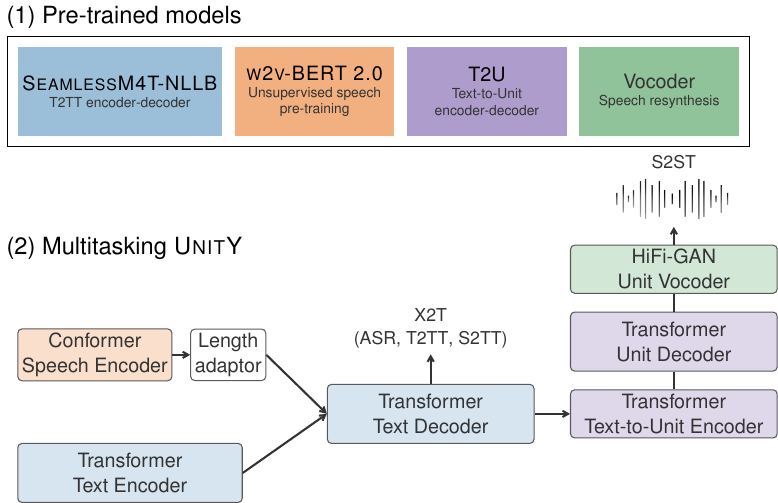}
\caption{\textbf{Overview of \mfourt.} (1) shows the pre-trained models used when finetuning multitasking \unity. (2) outlines multitasking \unity with its two encoders, text decoder, T2U encoder-decoder, and the supporting vocoders for synthesizing output speech in \sst. 
}\label{fig:modeling:overview}
\end{figure}

One of our goals with \mfourt is to bridge the gap between direct and cascaded models for \st in large multilingual and multimodal
settings by building a stronger direct \xt model (for translating both text and speech into text) that combines 
a strong speech representation learning model with a massively multilingual \mt model.
Beyond text outputs, our second goal builds on recent speech translation advancements, which have placed much emphasis on building systems that produce speech outputs~\citep{jia:interspeech:2019, lee-etal-2022-direct,unity}.
We enable speech-to-speech translation with \unity~\citep{unity}, a two-pass modeling framework that first generates text and subsequently predicts discrete acoustic units. 
Unlike cascaded models,
the different components in \unity (see \Cref{fig:modeling:overview}) can be jointly optimized.\footnote{There are two views of what constitutes a direct model in speech-to-speech translation literature: (1) A model that does not use intermediate text representation \citep{lee-etal-2022-direct} and (2) A model that directly predicts the target spectrogram~\citep{pmlr-v162-jia22b}}

The aforementioned approach alleviates the issue of cascaded error propagation and domain mismatch, while relying on an intermediate semantic representation to mitigate the problem of multi-modal source-target mapping.
 The vocoders for synthesizing speech are trained separately (see \Cref{sec:modeling:s2st:data}).
\Cref{fig:modeling:overview} provides an overview of the \mfourt model, including its four building blocks: 
(1) \nllbv a massively multilingual \mt model, 
(2) \wvbert, a speech representation learning model that leverages unlabeled speech audio data, 
(3) T2U, a text-to-unit sequence-to-sequence model, and 
(4) multilingual HiFi-GAN unit vocoder for synthesizing speech from units. 

The \mfourt multitask \unity model integrates components from the first three building blocks and is fine-tuned in three stages, starting from an \xt model (1,2) with English target only and ending with a full-fledged multitask \unity (1,2,3) system capable of performing \mt, \st and \sst, as well as \asr. 
In what follows, we first describe unsupervised speech pre-training (\wvbert) in \Cref{sec:modeling:x2t:w2vbert}.
We then introduce the \xt model in \Cref{sec:modeling:x2t}, starting with the data preparation pipeline in \Cref{sec:modeling:x2t:data}.
\Cref{sec:modeling:x2t:nllb} describes our multilingual \mt model, and \Cref{sec:modeling:x2t:stage12} details how the speech encoder and the \mt model are jointly fine-tuned for \xt with multimodal and multitask capabilities.
Next, we look at the \sst{} task, starting from the acoustic unit extraction pipeline and vocoder design to map units back to speech waveforms in \Cref{sec:modeling:s2st:data}
Then, we describe T2U pre-training in \Cref{sec:modeling:s2st:t2u}. 
\Cref{sec:modeling:s2st:stage3} ultimately outlines
how all these components come together in the third and final stage of fine-tuning. We evaluated our model using standard automatic metrics in \Cref{sec:modeling:main} and compared its performance with state-of-the-art speech translation models.

\subsection{Unsupervised Speech Pre-training}\label{sec:modeling:x2t:w2vbert}

Labels for speech recognition and translation tasks are scarce and expensive, especially for low-resource languages. It is challenging to train speech translation models with only limited access to supervision. 
Self-supervised pre-training with unlabeled speech audio data is, thus, a practical approach for reducing the need for supervision in model training. 
This method helps achieve the same recognition and translation quality with much less labeled data than models without pre-training. It also helps push the limits of model performance with the same amount of labeled data.
The most recent and publicly available state-of-the-art multilingual speech pre-trained model is MMS~\citep{mms}. It extends its predecessor, \xlsr~\citep{babu2021xls}, with additional 55K hours of training data and covers more than 1,300 new languages (see \Cref{tbl:modeling:sslsota}).
Besides MMS, USM~\citep{google_usm} is a proprietary SOTA multilingual speech pre-trained model that leverages the latest model architecture (BEST-RQ~\citep{best_rq} instead of wav2vec 2.0~\citep{baevski2020wav2vec}), has the largest scale of training data (12M hours), and covers more than 300 languages. 

\begin{table}[!t]
\small
    \centering
    \begin{tabular}{@{}lrrlc@{}}
        \toprule
        {\bf Model} & {\bf Languages} & {\bf Hours} & {\bf Model type} & {\bf Open model} \\
        \midrule
        \xlsrstot & 128 & 0.4M & wav2vec 2.0~\citep{baevski2020wav2vec} & $\checkmark$ \\
        USM & over $300^\dagger$ & 12M & BEST-RQ~\citep{best_rq} &  \\
        MMS & 1406 & 0.5M & wav2vec 2.0~\citep{baevski2020wav2vec} & $\checkmark$ \\
        \mfourtlg & over $\wvbertlangs^\dagger$ & 1M & \wvbert & $\checkmark$ \\
        \bottomrule
    \end{tabular}
    \caption{A comparison of multilingual speech pre-training in state-of-the-art \asr and \st models. $^\dagger$Estimated from the part of data that has language information.} 
    \label{tbl:modeling:sslsota}
\end{table}

\wvbert follows w2v-BERT~\citep{chung2021w2v} to combine contrastive learning and masked prediction learning, and improves w2v-BERT with additional codebooks in both learning objectives.
The contrastive learning module is used to learn Gumbel vector quantization (GVQ) codebooks and contextualized representations that are fed into the subsequent masked prediction learning module. The latter refines the contextualized representations by a different learning task of predicting the GVQ codes directly instead of polarizing the prediction probability of correct and incorrect codes at the masked positions. Instead of using a single GVQ codebook, \wvbert follows~\cite{baevski2020wav2vec} to use product quantization with two GVQ codebooks. Its contrastive learning loss $\mathcal{L}_c$ is the same as that in w2v-BERT, including a codebook diversity loss to encourage the uniform usage of codes. Following w2v-BERT, we use GVQ codebooks for masked prediction learning and denote the corresponding loss as $\mathcal{L}_{m_\text{GVQ}}$.
We also created an additional masked prediction task using random projection quantizers~\citep{best_rq} (RPQ), for which we denote the corresponding loss as $\mathcal{L}_{m_\text{RPQ}}$. The overall \wvbert training loss $\mathcal{L}$ is defined as follows:
\begin{align}
\mathcal{L} & = w_{c}\mathcal{L}_{c} +  w_{m_\text{GVQ}} \mathcal{L}_{m_\text{GVQ}} + 
 w_{m_\text{RPQ}} \mathcal{L}_{m_\text{RPQ}},
\end{align}
where loss weights $w_c$, $w_{m_\text{GVQ}}$ and $w_{m_\text{RPQ}}$ are set to $1.0$, $0.5$, and $0.5$, respectively. 

We follow the w2v-BERT XL architecture~\citep{chung2021w2v} for the \wvbert pre-trained speech encoder in \mfourtlg, which has 24 Conformer layers~\citep{gulati2020conformer} and approximately 600M model parameters. The \wvbert model is trained on 1 million hours of open speech audio data that covers over 143 languages.

\subsection{\xt: Into-Text Translation and Transcription}\label{sec:modeling:x2t}
\begin{figure}[!htb]
    \centering
    \includegraphics[width=\linewidth]{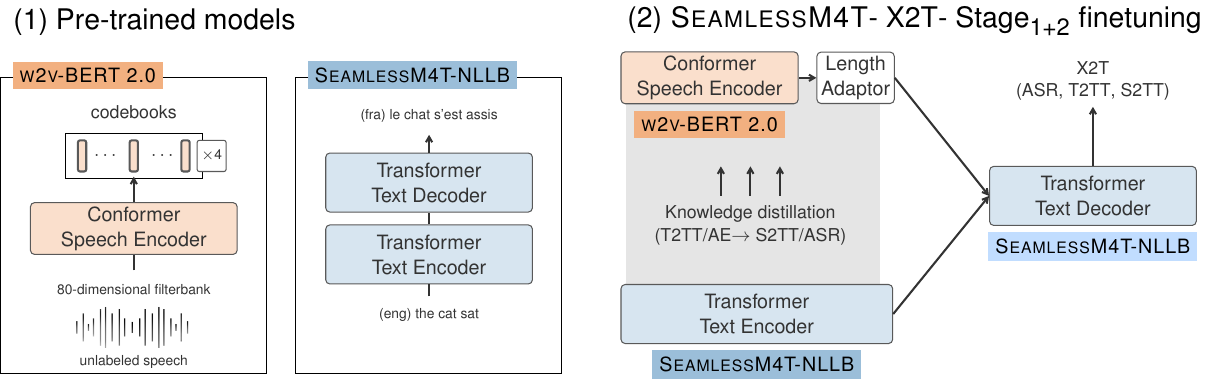}    \caption{\textbf{Overview of the \mfourt~\xt model.} (1) describes the main two building blocks: \wvbert and \nllbv. (2) describes the training of the \xt model. In $\text{Stage}_1$, the model is trained on \xeng directions and in $\text{Stage}_2$, \engx directions are added.}
    \label{fig:modeling:flow:x2t}
\end{figure}
The core of our multitask \unity framework is the \xt model, a 
multi-encoder
sequence-to-sequence models with a Conformer-based encoder~\citep{gulati2020conformer} for speech input and another for Transformer-based encoder~\citep{vaswani2017attention} for text input—both of which are joined 
with the same text decoder.
Our \xt model is trained on \st data pairing speech audio in a source language with text in a target language.

\subsubsection{Preparing \xt data}\label{sec:modeling:x2t:data}
\begin{figure}[tbh]
\centering
\includegraphics[width=\linewidth]{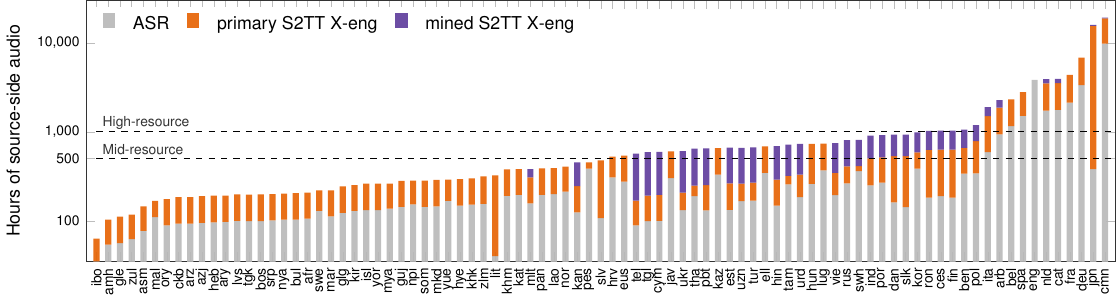}
\caption{Statistics of \asr and \xeng~\st data used to train our \mfourt model. We show the data size in hours of speech (log-scale) between \asr, \st primary and mined. Languages are sorted in ascending resource-level.
For numerical statistics see \Cref{tbl:modeling:s2tdata}}
\label{fig:s2tdata}
\end{figure}
\paragraph{Processing human-labeled data}
When using human-labeled data, we removed special tokens such as \texttt{<silence>} and \texttt{<no-speech>} from the verbatim transcriptions. 
We additionally perform length filtering to remove examples exceeding a maximum text length of 100 sub-word tokens 
(based on the text tokenizer described below) 
and pairs with a skewed text-to-audio length ratio that exceeds 5 sub-words per second. 
Doing so improves the batching efficiency when training and eliminates pairs that are likely to be noisy or misaligned.

\paragraph{Pseudo-labeling}
As with any sequence-to-sequence task, \st performance is dependent on the availability of high-quality training data. However, the amount of human-labeled \st data is scarce in comparison to its \mt or \asr counterparts. To address this shortage of labeled data, we resort to pseudo-labeling~\citep{jia2019leveraging, Pino2020SelfTrainingFE} the \asr data with a multilingual \mt model. In this case, we used NLLB-200-3.3B and generated pseudo-labels with the recommended decoding options from \cite{nllb2022}. Hereafter, we refer to human-labeled and pseudo-labeled data as \textit{primary} data. %

\paragraph{Parallel data mining}
Even with pseudo-labeled \asr data, the amount of \st data is insignificant compared to the scale of \mt data.
Consider for instance the English-Italian direction, one of the highly resourced pairs in \mt with over 128M parallel sentences—only 2M pairs of English text paired with Italian audio are available for \st.
Parallel data mining (see how \mineddata was built in \Cref{sec:appendix:data}) is another strategy we draw upon to collect more training data. 
This kind of mining, however, 
tends to produce noisy alignments and requires some filtering.
We use the top 400 hours 
of \mineddata (see \Cref{sec:appendix:data}) in each of 33 \xeng directions and the top 200 hours in each of 29 \engx directions based on \sonar alignment scores.
This amounts to an additional 18.3K hours of speech audio. 
We show in \Cref{{sec:modeling:ablation:data}} that these select amounts of mined data lead to a good trade-off between performance boosts and computational costs of training.

\paragraph{Filtering}\label{sec:modeling:filtering}
We perform additional filtering on the combined pool of \textit{primary} and \textit{mined} data.
Following \cite{nllb2022}, we implemented a toxicity filter. This removes pairs that have \emph{toxicity imbalance}, (i.e., when the difference in the number of toxic items detected in the source and target is above a certain threshold). For \st data, transcriptions are used as a proxy for the speech input when counting toxic items. We set the imbalance threshold at 1. 
In addition, we also applied a length filter. We removed pairs in which the utterance is shorter than 0.1 seconds or longer than 50 seconds. We also removed pairs in which the text is longer than 250 sub-words (based on the tokenizer described below). Lastly, we removed pairs in which the text contains more than 20\% of emojis, more than 50\% of punctuations, or more than 50\% of spaces.

\Cref{fig:s2tdata} shows the distribution of filtered \xeng~\st data used to train \mfourt models. Based on the total amount of speech audio hours in each language, we assessed its resource level: \textit{high-resource} are languages with more than 1000 hours of supervision, \textit{mid-resource} are those between 500 and 1000 hours, and \textit{low-resource} are those with less than 500 hours.

\paragraph{Training a Text Tokenizer.}
The tokenizer used in NLLB-200~\citep{nllb2022} is trained with \textit{SentencePiece}~\citep{sentencepiece} using the BPE algorithm~\citep{gage1994new,sennrich-etal-2016-neural}. 
These multilingual tokenizers, with their underlying vocabularies, are trained by sampling data from each language. Due to artifacts of sampling and the much larger number of unique symbols in logo-graphic writing systems, the result of this is that many key Chinese characters are missing from the original NLLB-200 vocabulary. To address this issue, we force the inclusion of these characters even in cases where they may not appear in the sampled \textit{SentencePiece} training data.
In order to decide which characters to include, we looked at the MTSU list\footnote{\url{https://lingua.mtsu.edu/chinese-computing/statistics/index.html}} and similar character frequency lists obtained from mined data in order to select the top 5000 Simplified Chinese characters, Traditional Chinese characters, and Japanese kanji characters. We then forced their inclusion, as long as they appeared at least 15 times in our training data to guarantee that the model would be able to learn how to embed these tokens.

We re-trained a 256K-sized \textit{SentencePiece} vocabulary on NLLB data~\citep{nllb2022} for \mfourt. The resulting tokenizer improves the coverage of the MTSU top 5K Chinese characters from 54\% to 84\%.

\subsubsection{Training a Large-Scale Multilingual Text-to-Text Translation Model}\label{sec:modeling:x2t:nllb}
We follow the same data preparation and training pipelines from \cite{nllb2022} using \stopes~\citep{andrews-etal-2022-stopes}. Having a smaller language coverage (100 instead of NLLB's 200 languages) allowed us to significantly decrease the size of the model. Whereas the full NLLB-200 model with mixture-of-experts is made up of 54.5B parameters (a number which can later be decreased via distillation), we opted for one of the smaller architectures proposed in \cite{nllb2022}, the 1.3B dense model.
We limited the NLLB-200 training data to the \ntextlangs \mfourt languages to be supported as target text. 
We additionally included over 75M bitexts from open-source \mt datasets that were not included in \cite{nllb2022}. These concern 
Modern Standard Arabic (arb), Mandarin Chinese (cmn), French (fra), Russian (rus), and Spanish (spa).
\begin{table}[!t]
\centering
\small
\begin{tabular}{@{}lcc@{}}
\toprule
& \multicolumn{2}{c}{\bf \mt ($\uparrow$\chrf)} \\\cmidrule{2-3}
{\bf Model} & \makecell[c] {\xeng\\ {\it (n=95)}}	&   \makecell[c] {\engx \\ {\it (n=95)}}\\
\midrule
\cite{nllb2022} & &\\
~- 3.3B  &	60.6 &	\bf 49.6 \\
~- 1.3B  &	59.3 &	48.2 \\
~- 1.3B-distil.  &	59.5 &	48.8 \\
\midrule
\nllbv-1.3B	& \bf 60.7	& \bf 49.6 \\ 
\bottomrule
\end{tabular}
\caption{Average \flores devtest \chrf over the \ntextlangs supported languages.}\label{tbl:nllb-v2-chrf}
\end{table}

We compare in \Cref{tbl:nllb-v2-chrf} the performance of \nllbv to that of comparably-sized NLLB models on \flores, averaging over our \ntextlangs languages when translating from English (\engx) and into English (\xeng). The model outperforms both smaller models from NLLB-200 (1.3B and 1.3B-distil) and is on par with the larger 3.3B model.

\subsubsection{Multimodal \& multitask into target text}\label{sec:modeling:x2t:stage12} 
In \mfourt, we leveraged foundational models either pre-trained on unlabeled data (w2v-BERT 2.0 for speech encoder pre-training) or trained on supervised high-resource tasks (NLLB model for \mt) to improve the quality of transfer tasks (speech-to-text and speech-to-speech). To fuse these pre-trained components and enable meaning transfer through multiple multimodal tasks, we trained an end-to-end model with (a) a speech encoder (\wvbert) postfixed with a length adapter, (b) text encoder (NLLB encoder), and (c) a text decoder (NLLB decoder). 
For the length adaptor, we used a modified version of M-adaptor~\citep{zhao22g_interspeech}, where we replaced the 3 independent pooling modules for Q, K, and V with a shared pooling module to improve efficiency.

The model is fine-tuned to jointly optimize the following objective functions:
\begin{align}
    \mathcal L_{\text{\st}} & = - \sum_{t=1}^{|y|} \log p(y^\text{text}_t | y^\text{text}_{<t}, x^\text{speech}),\\
    \mathcal L_{\text{\mt}} & = - \sum_{t=1}^{|y|} \log p(y^\text{text}_t | y^\text{text}_{<t}, x^\text{text}),
\end{align}
where $x^\text{text}$ and $x^\text{speech}$ are the source text and speech in the source language ${<}\ell_s{>}$ and $y^\text{text}$ is the target text in the target language ${<}\ell_t{>}$.
We additionally optimize an auxiliary objective function in the form of token-level knowledge distillation ($\mathcal{L}_{\text{KD}}$), to further transfer knowledge from the strong MT model 
to the student speech translation task (\st).
\begin{align}
    \mathcal{L}_{\text{KD}} & = \sum_{t=1}^{|y|} D_\text{KL}
    \left[
    p(. |y^\text{text}_{<t}, x^\text{text})\, \| \,
    p(. |y^\text{text}_{<t}, x^\text{speech})
\right].
\end{align}
The final loss is a weighted sum of all three losses:
    $\mathcal L = \alpha \mathcal{L}_{\text{\st}} + \beta \mathcal{L}_{\text{\mt}}+ \gamma \mathcal{L}_{\text{KD}},$
where $\alpha, \beta, \gamma$ are scalar hyperparameters tuned on the development data.
When the task does not fit the design of data triplets, we
then replaced translation tasks with auto-encoding—for example, on \asr $y^\text{text}$ is replaced by $x^\text{text}$ 
in which case the teacher distribution is from auto-encoding ($p(. |x^\text{text}_{<t}, x^\text{text})$).

We trained our \xt model in two stages. $\text{Stage}_1$ targeted training on supervised English \asr and into English \st data. We find that this step is necessary not only for improving the quality of \xeng translations but also \engx translations. In fact, we hypothesized that allowing the model to focus on one target language while fine-tuning multilingual speech representations shields it from the interference that can propagate back from the target side. In $\text{Stage}_2$, we add supervised \engx~\st and non-English \asr data to the mix.

\subsection{Speech-to-Speech Translation}\label{sec:modeling:s2st}
\begin{figure}[!htb]
    \centering
    \includegraphics[width=\linewidth]{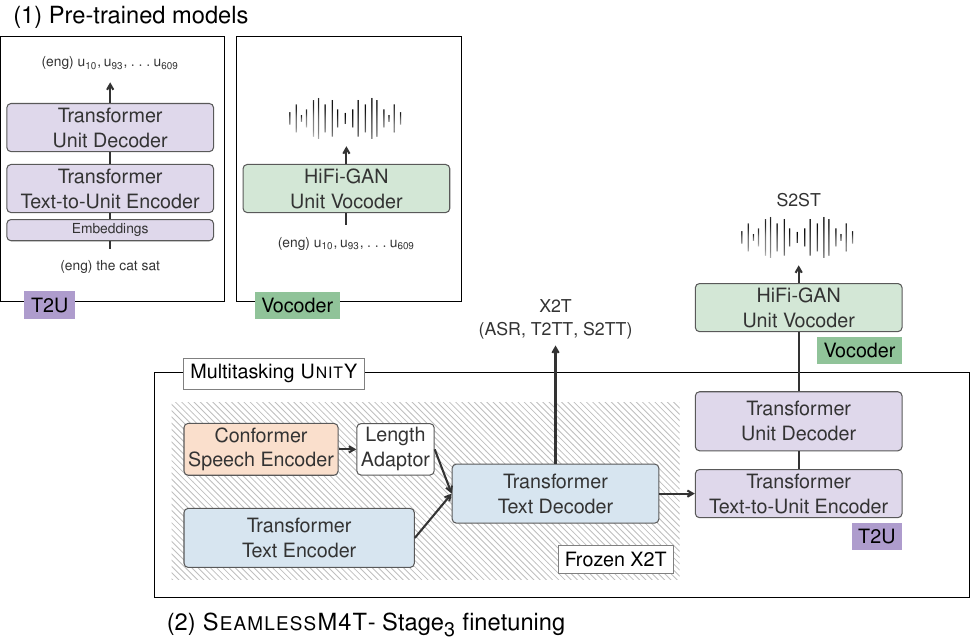}
\caption{
\textbf{Overview of the \mfourt multitask \unity model.} (1) describes the additional two building blocks on top of \xt: \tu encoder-decoder and unit vocoder. (2) describes the training of the \unity model. In $\text{Stage}_3$, the model is trained on \sst data.
}
\label{fig:modeling:flow:s2st}
\end{figure}
The key to our proposed speech-to-speech translation model is the use of self-supervised discrete acoustic units to represent target speech, thereby decomposing the S2ST problem into a speech-to-unit translation (S2UT) step and a unit-to-speech (U2S) conversion step.
For S2UT, the \mfourt model depicted in \Cref{fig:modeling:overview} uses \unity as a two-pass decoding framework which first generates text and
subsequently predicts discrete acoustic units.
Compared to the vanilla \unity model~\citep{unity}, 
(1) the core \st model initialized from scratch is replaced with an \xt model pre-trained to jointly optimize \mt, \st, and \asr,
(2) the shallow T2U model (referred to as T2U unit encoder and second-pass unit decoder in \cite{unity}) is replaced with a deeper Transformer-based encoder-decoder model with 6 transformer layers,
(3) the T2U model is also pre-trained on the T2U task rather than trained from scratch. 
The pre-training of \xt yields a stronger speech encoder and a higher quality first-pass text decoder,
while the scaling and pre-training of the T2U model allowed us to better handle multilingual unit generation without interference.

\subsubsection{Preparing \sst data}\label{sec:modeling:s2st:data}
\paragraph{Discrete acoustic units}
Recent works have achieved SOTA translation performance by using self-supervised discrete acoustic units as targets for building direct speech translation models~\citep{tjandra2019speech,lee-etal-2022-direct,lee-etal-2022-textless,zhang-etal-2022-speechut,chen-etal-2023-speech}. 
We extracted features from the 35$^\text{th}$ layer of \xlsr-1B~\citep{babu2021xls} for continuous speech representations at a 50Hz frame rate.
The mapping from \xlsr continuous representation space to discrete categories is required to map target speech into a sequence of discrete tokens. 
We randomly selected and encoded 10K unlabeled audio samples from each language of the \ntslangs supported target languages.
We then applied a $k$-means algorithm on these representations to estimate $K$ cluster centroids~\citep{lakhotia-etal-2021-generative,polyak21_interspeech,lee-etal-2022-direct}. 
These centroids resemble a codebook that is used to map a sequence of \xlsr speech representations into a sequence of centroid indices or acoustic units.
Experiments with different numbers of centroids ($K\in\{1000, 2000, 5000, 10000\}$) show that $K{=}10000$ with features from the 35$^\text{th}$ layer of \xlsr-1B achieves the best speech re-synthesis \wer~\citep{polyak21_interspeech}.

\xlsr has a broader language coverage than existing HuBERT~\citep{hsu2021hubert} models, and we found it provided similar speech re-synthesis performance to HuBERT on overlapping languages. We also experimented with \wvbert, which showed inferior performance. This can be attributed to w2v-BERT training with contrastive and MLM objectives, encouraging the model to only learn about semantic tokens rather than acoustic ones.

\paragraph{Synthesizing multilingual units with HiFi-GAN}
Following \cite{gong2023multilingual}, we built the multilingual vocoder for speech synthesis from the learned units. The HiFi-GAN vocoder~\citep{kong2020hifi} is equipped with language embedding to model the language-specific acoustic information. Moreover, to mitigate cross-lingual interference, language identification is used as an auxiliary loss in multilingual training.
We used a combination of commissioned and publicly available datasets, including single-speaker and multi-speaker \tts datasets, to train the multilingual vocoder on \nvocoderlangs target languages capable of converting the discrete units predicted by our S2UT model into waveforms. 
Compared to monolingual vocoders, we increased the model capacity by doubling the embedding dimension for both 
the duration predictor 
and the speech-language identification (LID) classifier to reach 1280.
\begin{figure}[!t]
\centering
\includegraphics[width=\linewidth]{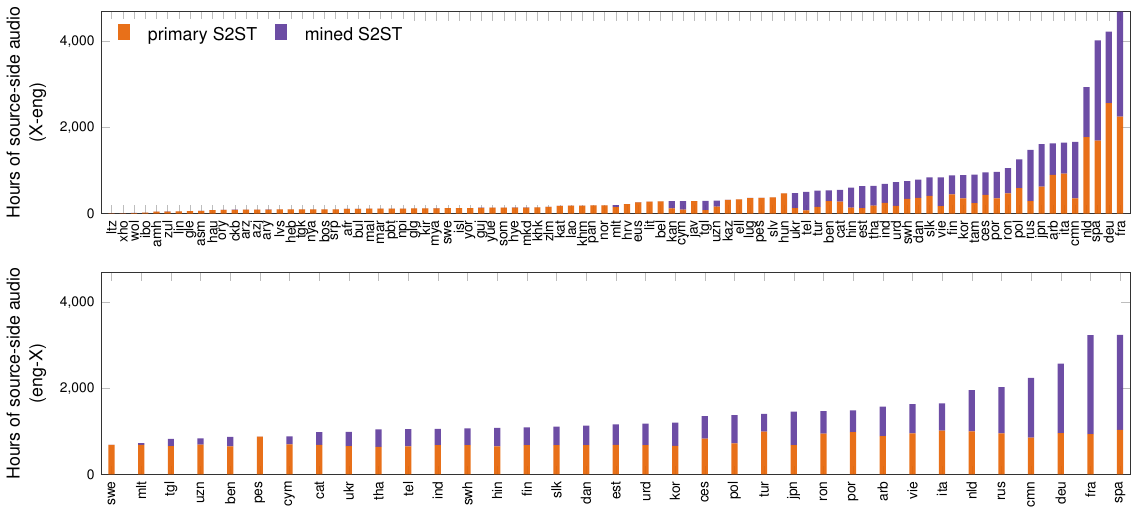}
\caption{Statistics of \sst data used in $\text{Stage}_3$ of training \mfourt model. We show the data size in hours of speech between primary and mined. Languages are sorted in ascending resource-level.
For numerical statistics see \Cref{tbl:modeling:s2stdata}}
\label{fig:s2stdata}
\end{figure}
\paragraph{Pseudo-labeling with text-to-unit}
The insufficient amount of parallel speech-to-speech training data significantly limits the training of high-quality S2UT models. To overcome this data scarcity, it is common practice to use \tts models to convert text from speech-to-text datasets (see \Cref{sec:modeling:x2t:data}) into synthetic speech~\citep{jia:interspeech:2019,lee-etal-2022-direct}.
This synthetic speech is in turn converted into units using the previously described unit extraction pipeline. This two-step unit extraction process is a slow process and is harder to scale given the dependencies on \tts models. 
High-quality off-the-shelf \tts models are hard to come by for all languages, especially for low-resource ones. Training reliable monolingual or multilingual in-house \tts models is also not scalable given the challenges around gathering high-quality clean speech data.
To overcome these challenges, we circumvented the need for synthesizing speech and trained multilingual text-to-unit (T2U) models on all the \nvocoderlangs target speech languages. 
These models can directly convert the text into target discrete units and can be trained on \asr datasets that are readily available. 
The multilingual training benefits from cross-lingual transfer between high-resource and low-resource languages, thereby also improving the quality of the pseudo-labeled data. To remove outlier samples from our paired data, we filtered based on the number of seconds of audio generated per text token length ratio, discarding any pair with a ratio exceeding 0.5.

\paragraph{Parallel data mining: \mineddata}
We added up to 2,500 hours of mined speech-to-speech data from \mineddata per language direction depending on its availability (see \Cref{sec:appendix:data}) . We used the XLSR-based unit extraction pipeline for extracting discrete acoustic units for target speech from the mined data. An in-house ASR model is then deployed to generate text transcriptions for the first pass decoder based on the target speech.

\Cref{fig:s2stdata} shows the distribution of all \sst data used to train \mfourt models between the primary and mined data.

\subsubsection{T2U modeling}\label{sec:modeling:s2st:t2u}
The T2U model is a Transformer-based encoder-decoder model trained on aligned text units from ASR data.
We trained T2U models for two purposes: (1) performing pseudo-labeling (\Cref{sec:modeling:s2st:data}) and (2) initializing the T2U component in \unity. 
For (1), 
we trained a model with 12 encoder and 12 decoder layers.
For (2), we trained a smaller T2U model with 6 encoder and 6 decoder layers.
Initial experiments showed that, although the smaller T2U model is of a lower quality than the larger one, fine-tuning the smaller T2U in \unity with labels from the larger one (i.e., distilling knowledge from the stronger T2U) can bridge the gap while being parameter-efficient.

\subsubsection{$\text{Stage}_3$ Finetuning for \sst}\label{sec:modeling:s2st:stage3}
In the last stage of fine-tuning, we initialized the multitask \unity model (see \cref{fig:modeling:overview}) with (1) the pre-trained \xt model and (2) the pre-trained T2U model and fine-tuned on a combination of \xeng and \engx~\sst translation data totaling 121K hours (see breakdown in \cref{fig:s2stdata}).
We froze the model weights corresponding to the \xt model and only fine-tuned the T2U component. This is to ensure that the performance of the model on tasks from the previous stages of fine-tuning remains unchanged. 

\subsection{The \mfourt Models}\label{sec:modeling:main}
With all the components laid out in the previous sections, we trained the \mfourtlg model in the outlined three stages. 
\mfourtlg has 2.3B parameters and is fine-tuned on
\mt for \ntextlangs languages paired with English, on \asr for \nasrlangs languages, 
on \st for 89 languages paired with English, and on \sst for 95 directions into English and \ntslangs target languages out of English. 
The amount of supervised data per direction is detailed in \cref{tbl:modeling:s2stdata,tbl:modeling:s2tdata}. This means that, for some source languages, our models are evaluated zero-shot to reach the coverage described in \cref{tbl:coverage} of \nsslangs-eng.

To provide a reasonably sized model, we followed the same recipe to train \mfourtmd. This model has 57\% fewer parameters than \mfourtlg and is intended to be an accessible test bed to either fine-tune, improve on, or engage in analysis with. \mfourtmd has the same coverage as \mfourtlg but builds on smaller and more parameter-efficient components (see \Cref{fig:modeling:overview}). 
We pre-trained a smaller \wvbert with 300M parameters and used the distilled model from \cite{nllb2022} (\nllbtinydistil) to initialize the \mt{} modules of the multitask \unity. See a comparison between \mfourtlg and \mfourtmd in \Cref{tbl:modeling:collection}. 
\begin{table}[!ht]
    \centering
    \small
    \begin{tabular}{ccccc}
    \toprule
          & {\bf $\text{\wvbert}^*$} & {\bf \mt} & {\bf T2U}  & {\bf Total} \\
    \midrule
\mfourtlg & 669M & 1370M & 287M & 2326M  \\
\mfourtmd & 366M & 615M & 170M & 1151M \\
    \bottomrule
    \end{tabular}
    \caption{\#parameters of the building components used  in \mfourt models.\\
    *: includes the parameters of the length adaptor .\\
    }
    \label{tbl:modeling:collection}
\end{table}

We evaluated our models on all four supervised tasks (\mt, \asr, \st, and \sst) as well as the zero-shot task of text-to-speech translation (\ttst, also referred to as cross-lingual text to speech synthesis~\citep{Zhang2023SpeakFL}). 
To generate text hypotheses, we decoded with beam-search (width${=}5$).
We scored with \chrf for \mt
and SacreBLEU for \st (default 13a tokenizer and character-level tokenizer for Mandarin Chinese (cmn), Japanese (jpn), Thai (tha), Lao (lao), and Burmese (mya); see signatures in \Cref{tbl:eval:metrics}).
For \asr, we scored with \wer on normalized transcriptions and references following \cite{whisper}.

During \sst and \ttst inference, we performed two-pass beam-search decoding— 
the best hypothesis out of the first-pass decoding is embedded with the text decoder and is sent to T2U to search for the best unit sequence hypothesis. We use a beam-width of 5 for both searches. We evaluated \sst and \ttst accuracy with \asrbleu~\citep{lee-etal-2022-direct} with \whisperlarge as the underlying \asr model for \engx directions and with \whispermedium for \xeng directions. We set the decoding temperature of Whisper at zero and used greedy decoding to ensure a deterministic behavior of the ASR model.
The transcribed hypotheses, as well as the references, are normalized following \cite{whisper} before computing \bleu scores in the same manner we did for \st.

\subsubsection{Comparison to cascaded approaches.}
On the set of languages supported by both \mfourt and Whisper, we compare in \Cref{tbl:modeling:cascaded:x2t} the performance of our direct \st model to that of cascaded models, namely combinations of Whisper \asr models and NLLB \mt models.
\mfourtlg surpasses the cascaded models with less than 3B of parameters  in \xeng directions by 2 \bleu points and in \engx directions by 0.5 \bleu points. We also add to the comparison in \Cref{tbl:modeling:cascaded:x2t} cascaded models with the large \nllbmedium~\mt model. These models exceed 4B parameters and only outperform \mfourtlg in \engx directions. \mfourtlg improves on \whisperlarge+\nllbmedium by 1.3 \bleu points on average in \xeng directions.

\Cref{tbl:modeling:cascaded:s2st} compares \sst between \mfourtlg and cascaded models. For \sst, we look at two options for cascading: (1) 3-stage with \asr, \mt, and \tts and (2) 2-stage with \st and \tts. 
Our \mfourtlg outperforms 2-stage cascaded models on \fleurs~\xeng directions by 9 \asrbleu points. It also outperforms stronger 3-stage cascaded models (\whisperlarge + \nllbmedium  + \yourtts) by 2.6 \asrbleu points. On \cvss, \mfourtlg outperforms the 2-stage cascaded model (\whisperlarge + \yourtts) by a large margin of 14 \asrbleu points. 
On \fleurs~\engx directions, \mfourtlg has an average \asrbleu of 21.5 on 32 \xeng directions excluding target languages where \whisperlarge (the \asr model used for \asrbleu) has a \wer higher than 100. Comparably, the medium-size model (\mfourtmd) scores an average \asrbleu of 15.4 on \sst~\engx directions.

\begin{table}[!ht]
    \centering
    \small
    \begin{tabular}{@{}lcccc@{}}
        \toprule
         &  &  & \multicolumn{2}{c}{\makecell[c]{\bf \st ($\uparrow$\bleu)}} \\
        \cmidrule{4-5}
         {\bf Model} & {\bf  type} &  { \bf size}  & \makecell[c]{\xeng \\{\it (n=81)}} & \makecell[c]{\engx\\{\it (n=88)}} \\
         \midrule
        \whispermedium (\asr)\phantom{a} + \nllbsmall & \multirow{4}{*}{cascaded} &  2B &  19.7 & 20.5 \\
        \whispermedium (\asr)\phantom{a}  + \nllbmedium &  &  4B &  20.4 & 21.8 \\
        \whisperlarge (\asr)+ \nllbsmall &  & 2.8B & 22.0 & 21.0 \\
        \whisperlarge (\asr)+ \nllbmedium &  & 4.8B & 22.7 & \bf 22.2  \\
        \midrule
        \whisperlarge & \multirow{2}{*}{direct} & 1.5B & 17.9 & - \\
        AudioPaLM-2-8B-AST &  & 8B & 19.7 & - \\
\midrule
        \mfourtmd & \multirow{2}{*}{direct} & 1B & 20.9 & 19.2 \\
        \mfourtlg &  & 2B & \bf 24.0 & 21.5 \\
        \bottomrule
    \end{tabular}
\caption{Comparison against cascaded \asr+\mt models on  \fleurs~\st.}\label{tbl:modeling:cascaded:x2t}
\end{table}

\begin{table}[ht]
    \small
    \centering
    \begin{tabular}{@{}lcccc@{}}
        \toprule
         &  &  & \multicolumn{2}{c}{\makecell[c]{\bf S2ST~\xeng \\\bf($\uparrow$\asr-BLEU)}} \\
        \cmidrule{4-5}
        {\bf Model} & {\bf type} & {\bf  size } & \makecell[c]{\fleurs \\{\it (n=81)}} & \makecell[c]{\cvss \\{\it (n=21)}}\\
        \midrule
       \multicolumn{4}{l}{\yourtts~\citep{casanova2022yourtts}}\\
       \phantom{abcd}+\whisperlarge (\st)  & \makecell[c]{2-stage\\cascaded} & 1.6B & 17.3 & 22.6 \\
       \midrule
       \phantom{abcd}+\whispermedium (\asr) \phantom{a}+ \nllbsmall &  \multirow{4}{*}{\makecell[c]{3-stage\\cascaded}}  &  2.1B &  19.9 & \\  
        \phantom{abcd}+\whispermedium (\asr)\phantom{a} + \nllbmedium &  &  4.1B &  20.6 & \\
        \phantom{abcd}+\whisperlarge (\asr)+ \nllbsmall &  & 2.9B & 22.1 & \\
        \phantom{abcd}+\whisperlarge (\asr)+ \nllbmedium &  & 4.9B & 23.2 &  \\
        \midrule
        \mfourtmd & unified & \sizemd & 20.4  & 28.1 \\
        \mfourtlg & unified & \sizelg & \bf 25.8  & \bf 36.5\\
        \bottomrule
    \end{tabular}
    \caption{Comparison against 2/3-stage cascaded models on \fleurs and \cvss~\sst~\xeng.}\label{tbl:modeling:cascaded:s2st}
\end{table}

\subsubsection{Multitasking \xt results.} \label{subsec:multitaskingxtresults}
\begin{table}[!t]
\small
    \centering
    \begin{tabular}{@{}lHccccc@{}}
        \toprule
         &  &  & \multicolumn{4}{c}{ \bf \st ($\uparrow$\bleu) }\\
        \cmidrule{4-7}
        
        {\bf Model}  & {\bf type} & \multirow{2}{*}{\makecell[c]{\bf size}} & \makecell[c]{\fleurs\\\xeng \\{\it (n=81)}} & \makecell[c]{\fleurs\\\engx \\{\it (n=88)}} & \makecell[c]{\covost\\\xeng \\{\it (n=21)}} & \makecell[c]{\covost\\\engx\\ {\it (n=15)}} \\
        \midrule
        \makecell[l]{\xlsrstot} & direct & 2.6B &  & x & 22.1 & 27.8 \\
        \makecell[l]{\whisperlarge} & direct & 1.5B & 17.9 & x & 29.1 & x \\
        \makecell[l]{\audiopalmast}  & direct & 8.0B & 19.7 & x & \bf 37.8 & x \\
        \midrule
        \mfourtmd & direct & \sizemd & 20.9 & 19.2 & 29.8 &  26.6 \\
        \mfourtlg & direct & \sizelg & \bf 24.0 & \bf 21.5 & 34.1 & \bf 30.6  \\  
        \bottomrule
    \end{tabular}\\[10pt]
     \begin{tabular}{@{}lHccccc@{}}
        \toprule
         &  &  & \multicolumn{2}{c}{\bf \asr ($\downarrow$WER) } & \multicolumn{2}{c}{\bf \mt ($\uparrow$\chrf)}\\
        \cmidrule(r){4-5}
        \cmidrule(l){6-7}
        {\bf Model}  & Model type & \multirow{2}{*}{\makecell[c]{\bf size}} & \makecell[c]{\fleurs \\ {\it (n=77)}} & \makecell[c]{ \fleurs-54\\ {\it (n=54)} }  & \makecell[c]{\flores\\\xeng\\{\it (n=\ntextlangs)}} & \makecell[c]{\flores\\\engx\\{\it (n=\ntextlangs)}} \\
        \midrule
        \nllbmedium & direct & 3.3B & x & x & 60.7 & 49.6 \\
        \midrule
        \makecell[l]{\whisperlarge} & direct & 1.5B & 41.7 & 43.7 & x & x \\
        \makecell[l]{MMS-L61-noLM-LSAH} & direct & 1.0B & x & 31.0 & x & x \\
        \makecell[l]{MMS-L1107-CCLM-LSAH} & direct & 1.0B$^*$ & x & \bf 18.7 & x & x \\
        \midrule
        \mfourtmd & direct & \sizemd & \bf 21.9 & 22.0 & 55.4  & 48.4 \\
        \mfourtlg & direct & \sizelg & 23.1 & 23.7 &  \bf 60.8 & \bf 50.9 \\
        \bottomrule
    \end{tabular}
    \caption{\textbf{Multitasking \xt results.} Performance of \mfourtlg on \xt tasks (\st, \asr and \mt) compared to SOTA direct translation models. 
    For \fleurs~\st~\xeng, we report the average \bleu scores over languages Whisper supports.
    For \fleurs~\asr, we report the average normalized \wer over languages supported by both \mfourt and Whisper.
    For MT, we average \chrf~scores over the supported written languages in \mfourt. 
    *: MMS is CTC-based, and this version decodes with an n-gram language model for each language.
    Note that for all external models included in this comparison, we lifted the results reported in their respective papers and matched their evaluation and scoring pipeline for a fair comparison.\protect\footnotemark
    }
    \label{tbl:modeling:main:x2t}
\end{table}
\footnotetext{
Scoring \whisperlarge, using \url{https://github.com/openai/whisper} with the recommended decoding options, results in \bleu scores lower by 0.3 \bleu points on average than what is reported in~\cite{whisper}.
}
\begin{table}[!htb]
    \centering
    \small
    \begin{tabular}{@{}lcccc@{}}
        \toprule
         & \multicolumn{4}{c}{\bf \fleurs~\st~\xeng ($\uparrow$\bleu)}  \\\cmidrule{2-5}
          {\bf Model} & \makecell[c]{High\\ {\it (n=15)}} &\makecell[c]{ Medium\\ {\it (n=25)}} & \makecell[c]{Low \\ {\it (n=34)}}& \makecell[c]{Low$^\dagger$ \\ {\it (n=23)}} \\
        \midrule

        \whisperlarge & 24.2 & 19.4 & 16.1 & 18.1 \\
        \audiopalmast & \bf 27.9 & 20.9 & 18.0 & 22.0 \\
        \midrule
        \mfourtmd & 23.9 & 21.8 & 22.2 & 23.5 \\
        \mfourtlg & 26.9 & \bf 25.2 & \bf 25.4 & \bf 27 \\        
        \bottomrule
    \end{tabular}
    \caption{\textbf{\fleurs~\st~\xeng by resource-level.} In each resource-level (high, medium and low), we average over languages that are covered by all 3 models. In low$^\dagger$, we exclude low-resource languages that are evaluated as zero-shot by \audiopalmast.}
    \label{tbl:modeling:s2t:res}
\end{table}
\begin{table}[!t]
\small
    \centering
    \begin{tabular}{@{}lHccccccc@{}}
        \toprule
         {\bf Model} & Model type 
         & \multirow{2}{*}{\makecell[c]{\bf size}} 
         & \multicolumn{3}{c}{\makecell[c]{\fleurs~\xeng {\it (n=81)}}}
         & \multicolumn{3}{c}{\makecell[c]{\fleurs~\engx {\it (n=88)}}} \\  
         \cmidrule(l){4-6}  \cmidrule(r){7-9}
        &  &  
        & \scriptsize$\uparrow$\bleu 
        &\scriptsize$\uparrow$\spbleu  
        &\scriptsize$\uparrow$\blaser
        &\scriptsize$\uparrow$\bleu 
        & \scriptsize$\uparrow$\spbleu  
        & \scriptsize$\uparrow$\blaser \\
        \midrule
        \makecell[l]{\whisperlarge} & direct 
                & 1.5B & 17.9 & 19.9 & 3.29  & x & x & x \\
        \midrule
        \mfourtmd & direct & \sizemd 
        & 20.9 & 23.1 & 3.56 
        & 19.2 & 26.0 & 3.68 \\
        \mfourtlg & direct & \sizelg 
        & \bf 24.0 &\bf 26.4 & \bf 3.66 
        & \bf 21.5 & \bf 28.9 & \bf 3.71\\
        \bottomrule
    \end{tabular}
    \caption{\textbf{\st results with \spbleu and \blaser} we report here the performance of \whisperlarge and \mfourtlg measured with \spbleu \& \blaser. Note that unlike \bleu scores copied from \cite{whisper}, the \spbleu and \blaser scores are based on our evaluation using \url{https://github.com/openai/whisper} with the recommended decoding options.
    }
    \label{tbl:modeling:spbleu}
\end{table}
\begin{table}[!t]
\small
    \centering
    \begin{tabular}{@{}lHHccccccc@{}}
        \toprule
         &  &  
         & \multicolumn{4}{c}{\bf \sst ($\uparrow$\asrbleu)}
        & \multicolumn{3}{c}{\bf \sst ($\uparrow$\blaser)}\\
        \cmidrule(l{2pt}){4-7} \cmidrule(r){8-10}
         {\bf Model} & Model type 
         & \multirow{2}{*}{\makecell[c]{\bf size}} 
         & \makecell[c]{\fleurs\\\xeng \\ {\it (n=101)}} 
          & \makecell[c]{\fleurs\\\xeng \\ {\it (n=82)}} 
         & \makecell[c]{\fleurs\\\engx\\ {\it (n=35)}}
         & \makecell[c]{\fleurs\\\engx\\ {\it (n=32)}}
         & \makecell[c]{\fleurs\\\xeng \\ {\it (n=82)}} 
         & \makecell[c]{\fleurs\\\engx\\ {\it (n=35)}}
         & \makecell[c]{\fleurs\\\engx\\ {\it (n=32)}}  \\
        \midrule
        \mfourtmd & & \sizemd  & 17.9 & 20.8 & 14.3 & 15.4 
        & 3.62 & 3.63 & 3.63
        \\ 
        \mfourtlg & & \sizelg & 22.7 & 26.3 & 19.8 & 21.5 
        & 3.85 & 3.94 & 3.95
        \\
        \bottomrule
    \end{tabular}
    \caption{\textbf{\sst results with \asrbleu and \blaser} we report here the performance of \mfourtlg and \mfourtmd measured with \asrbleu \& \blaser.
    }
    \label{tbl:modeling:sst:blaser}
\end{table}

We report in \Cref{tbl:modeling:main:x2t} results on the \fleurs benchmark for the tasks of \asr and \st (\xeng and \engx), and the related \flores benchmark for \mt (\xeng and \engx).
We also report results on the evaluation test set of \covost (\xeng and \engx)
The \mfourt model outperforms the previous direct SOTA model (AudioPaLM-2 8B AST~\citep{Rubenstein2023AudioPaLMAL}) by 4.2 \bleu points in \st \xeng directions (i.e., an improvement of 20\%).
In \covost~\engx, \mfourtlg improves upon the previous SOTA (\xlsr) by 2.8 \bleu points. However, in \xeng, it lags behind AudioPaLM by 3.7 \bleu points. 
For \asr, \mfourt outperforms Whisper~\citep{whisper} on the overlapping 77 supported languages with a \wer reduction of 45\%. 
We additionally compared against MMS~\citep{mms} on \fleurs-54, a subset of \fleurs languages that both MMS and Whisper support. \mfourtlg outperforms the MMS variants evaluated with CTC by more than 6\% WER, but it is surpassed by the variants that leverage
monolingual n-gram language models (5\% WER better). 

In the \mt support task, our \mfourt model matches the performance of NLLB-3.3B~\citep{nllb2022} in \xeng directions and improves on \engx directions by 1 \chrf point.
To further understand where the improvements in 
\fleurs~\st~\xeng directions are coming from, we bucket languages by resource-level (see the exact list of languages in \Cref{tbl:modeling:s2tdata}) and report average \bleu scores per resource-level in \Cref{tbl:modeling:s2t:res}. The results show that \mfourtlg strongly improves the quality of translating from low-resourced languages with an improvement of +7.4 \bleu (i.e., 40\% improvement over AudioPaLM-2-8B-AST). We also average in column low$^\dagger$ over low-resource directions that are supervised in AudioPaLM-2-8B-AST—the gain of +5 \bleu in that subset of directions suggests that this improvement goes beyond sheer supervision, but instead should be attributed to the quality of supervised data and the training recipes.

\subsubsection{Zero-shot Text-to-Speech Translation}\label{sec:modeling:t2st}
We evaluate \fleurs~\st on the reverse task of \ttst. We report in \Cref{tbl:modeling:t2st}
the average \asrbleu scores on 87 \xeng directions (the overlap between \fleurs and the languages supported by \mfourt text encoders). We also report the average \asrbleu on 32 \engx directions (excluding Bengali, Telugu and Northern Uzbek where \whisperlarge~\asr~\wer is above 100).
The \xeng average \asrbleu is higher than the \asrbleu of \sst~\xeng (34.9 vs. 24.6) where the \engx average is similar to that of \sst (22.5 vs. 21.5). This result demonstrates that (1) the quality of \mfourt on zero-shot \ttst is on-par with the supervised tasks and (2) that non-English speech source is the most challenging input to translate with our model.

\begin{table}[!tbh]
    \centering
    \small
    \begin{tabular}{@{}lccc@{}}
        \toprule
         & \multicolumn{2}{c}{\bf \fleurs~\ttst ($\uparrow$\asrbleu)}  \\
        \cmidrule(lr){2-4}
        {\bf Model} 
        &\makecell[c]{\xeng\\ {\it (n=88)}} 
        &\makecell[c]{\engx\\ {\it (n=35)}} 
        &\makecell[c]{\engx\\ {\it (n=32)}} \\
        \midrule
        \mfourtlg &  34.9 & 20.7 & 22.5  \\
        \bottomrule
    \end{tabular}
    \caption{\textbf{zero-shot \fleurs~\ttst} we report the average \asrbleu of \mfourtlg on \fleurs~\ttst.
    }
    \label{tbl:modeling:t2st}
\end{table}

\subsubsection{Evaluation with \spbleu and \blaser.}
To avoid expanding the set of special case languages evaluated with character-level tokenization, we evaluated with \spbleu using the \flores-200 sentence piece tokenizer. \Cref{tbl:modeling:spbleu} reports \spbleu scores on \fleurs~\st~\xeng and \engx. 
We also report in the same table the average \blaser scores (for more on \blaser see \Cref{sec:eval:blaser}).
Since \blaser is modality-agnostic, we can also score the task of \sst with \blaser. 
\Cref{tbl:modeling:sst:blaser} provides the average \blaser scores of \mfourtlg and \mfourtmd on \sst~\xeng and \engx directions. Since \blaser supports 83 languages (including English), we average over 82 \xeng directions. For \engx, we show averages of 35 languages, then averages excluding 3 languages with a WER exceeding 100\%. Since \blaser supports all 35 target languages the scores are more reliable and less affected by the noisiness of the ASR model underlying \asrbleu (a difference of -1.7\asrbleu points with the addition of 3 directions).
The full results and metrics per evaluation direction can be found at \url{https://github.com/facebookresearch/seamless_communication}.

\subsubsection{Evaluation of \xy directions with \spbleu.}
Since \mfourt models support multiple languages on both the source and target sides, we can evaluate non-English centric directions (labeled \xy) in a zero-shot manner. 
\begin{figure}[!htb]
\centering
\includegraphics[width=\linewidth]{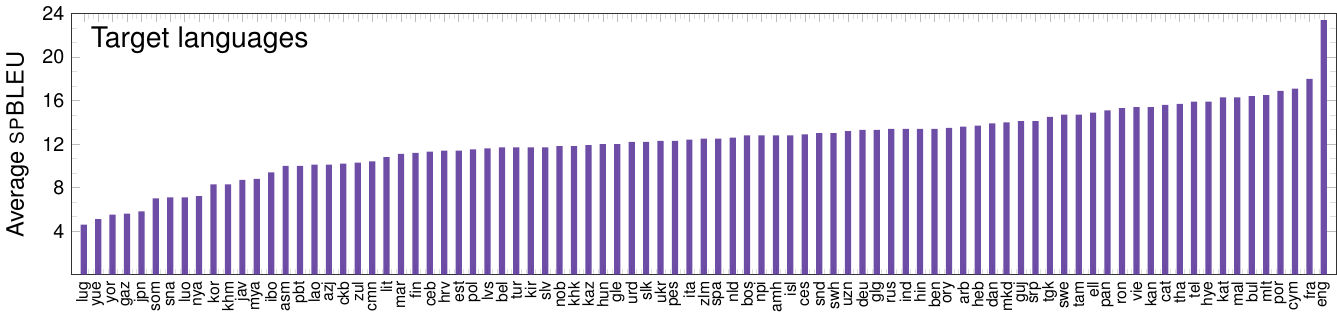}
\caption{\textbf{\st~\fleurs~\xy results}. We evaluate \xy directions from \fleurs and average \spbleu scores. 
For a given target text language, we average scores over 100 source languages.
}\label{fig:modeling:x2t:xxresults}
\end{figure}
\FloatBarrier
\subsection{Analysis and Ablations}
\subsubsection{Unsupervised speech pre-training}
We explored various techniques to enhance the quality of our encoders’ representations, including 
algorithm-wise improvements and pre-training data scaling.

\begin{table}[t]
    \centering
    \small
    \begin{tabular}{@{}clHc@{}}
        \toprule
        \multirow{1}{*}{\bf ID} & \multirow{1}{*}{\bf Configuration} & &\multicolumn{1}{c}{\makecell[c]{\bf \fleurs~\xeng\\\bf($\uparrow$\bleu)}}\\
        \midrule
        A & w2v-BERT baseline with updated XLS-R data (400K hrs, 143 langs) & 16.1 & 12.4 \\
        B & A + product quantization with 2 GVQ codebooks & 16.3 & 12.5 \\
        C & B + increased open training data from 400K hours to 1M hours & 16.8 & 12.7 \\
        D & C + 2 RPQ codebooks for masked prediction objective & 17.0 & \bf 12.8 \\
        \bottomrule
    \end{tabular}
    \caption{Ablation on w2v-BERT variants and training data scaling.}\label{tbl:w2vbert}
\end{table}

\paragraph{Experimental setup}
In our ablation, we aimed to evaluate the w2v-BERT variants by their performance on the downstream \st task. All pre-trained w2v-BERT speech encoders are composed of 24 Conformer layers~\citep{gulati2020conformer} 
with approximately 600M of parameters. 
Each speech encoder was used to initialize an \st model.
The text decoder was initialized with the decoder from \nllbsmall, a large multilingual machine translation model covering 200 languages \citep{nllb2022} with 1.3B parameters.  
We fine-tuned the \st models on the task of speech translation into English (\xeng~\st) 
on
67 languages.
We fine-tuned all the speech encoder parameters and only fine-tuned LayerNorms and Self-attention in the text decoder (LNA-D~\citep{li-etal-2021-multilingual}). 
Our learning rate increased up to 3e-4 through 4000 warm-up updates and subsequently followed the inverse square root learning rate schedule. We trained on 32 GPUs with a batch size of 960K frames in each for 100K updates.
We report \bleu scores (SacreBLEU\footnote{see \Cref{tbl:eval:metrics}} \citep{post-2018-call}) evaluated on the test set of all 101 \xeng directions from \fleurs~\citep{fleurs2022arxiv}. 
Given the coverage of our training data, this means that 34 of the directions were evaluated as zero-shot.

\paragraph{Results}
We summarize our ablation results in~\Cref{tbl:w2vbert}. 
We see that product quantization with 2 GVQ codebooks outperforms normal quantization with a single GVQ codebook (A vs. B). Scaling training data leads to performance gains (B vs. C). Adding additional masked prediction learning objectives with 2 RPQ codebooks helps improve performance (C vs. D).

\begin{table}[!hb]
\centering
\footnotesize
\begin{tabular}{@{}cccccccccc@{}}
    \toprule
    \multirow{2}{*}{\makecell{\bf Language\\(code)}}  & 
   \bf \multirow{2}{*}{ASR} &
   \multicolumn{4}{c}{\bf \st} &
   \multicolumn{4}{c}{\bf \sst} \\
   \cmidrule(lr){3-6}
   \cmidrule(lr){7-10}
   & &
    \multicolumn{2}{c}{\xeng}&
    \multicolumn{2}{c}{\engx}&
    \multicolumn{2}{c}{\xeng}&
    \multicolumn{2}{c}{\engx} \\
     \cmidrule(l){2-2}
     \cmidrule(lr){3-4}
     \cmidrule(lr){5-6}
     \cmidrule(lr){7-8}
     \cmidrule(lr){9-10}
     & Primary & Primary & Mined & Primary & Mined & Primary & Mined & Primary & Mined \\
    
    \midrule
    arb & 934 
        & 942 & 600 
        & 1,959 & 600 
        & 899  & 736 
        & 895 & 681 \\
    ben & 338 
        & 320 & 600 
        & 1,987 & 499 
        & 292 & 246 
        & 652 & 221 \\
    eng & 3,845 & - & - &-  & - & - & - & - & - \\
    hin & 148 
        & 143 & 600 
        & 2,066 & 600 
        & 138 & 466 
        & 656 & 430 \\
    ind & 250 
        & 254 & 600 
        & 1,818 & 596 
        & 248 & 443 
        & 684 & 375 \\
    ita & 591 
    & 910 & 600 
    & 2,279 & 600 
    & 930 & 716 
    & 1,020 & 636 \\
    jpn & 381 
        & 15,141 & 600 
        & 1,798 & 259 
        & 624 & 993 
        & 681 & 779 \\
    por & 269 
        & 246 & 600 
        & 2,250 & 600 
        & 355 & 606 
        & 983 & 508 \\
    rus & 264 
        & 144 & 600 
        & 2,161 & 600 
        & 290 & 1,093 
        & 959 & 1,075 \\
    spa & 1,515 
        & 1,285 & - 
        & 2,505 & 574 
        & 1,694 & 2,335 
        & 1,035 & 2,209 \\
    swh & 361 
        & 50 & 600 
        & 1,930 & 596 
        & 342 & 411 
        & 682 & 392 \\
    tha & 190 
        & 59 & 600 
        & 1,941 & 101 
        & 184 & 462 
        & 641 & 408 \\
    tur & 169 
        & 100 & 600 
        & 2,135 & 600 
        & 156 & 375 
        & 998 & 411 \\
    urd & 185 
        & 145 & 600 
        & 1,844 & 507 
        & 179 & 555 
        & 682 & 502 \\
    vie & 194 
        & 151 & 600 
        & 2,396 & 600 
        & 176 & 666 
        & 954 & 684 \\
    \midrule
    \bf Total & 9,633  & 19,890 & 7,800 & 29,068 & 2,701 & 6,508 & 10,103 & 11,523 & 9,312 \\
    \bottomrule
\end{tabular}
\caption{Hours of data in the ablation dataset for the tasks of \asr, \st and \sst, split between \engx and \xeng when relevant. For each task, we report hours of training data between primary and mined. By default, \st mined data is capped at 400 hours in \xeng and at 200 hours in \engx.
}\label{tbl:s2t-ablation-dataset}
\end{table}
 
\subsubsection{Multimodal \& multitasking \xt}\label{sec:modeling:ablation:multitask}
\paragraph{Ablation dataset}
To iterate on different multitasking recipes, we constructed a smaller multilingual speech translation benchmark with 14 languages paired with English. 
The supervised \st data comes from two sources: primary (open-source or licensed) and mined, whereas the \asr data is either from open-sourced or licensed datasets.
The \mt data we used in our multitasking fine-tuning is limited to bitexts produced in the pseudo-labeling process, i.e., translated transcriptions in the \asr datasets (see \Cref{sec:modeling:x2t:data}).  
For a breakdown of the ablation dataset, see \Cref{tbl:s2t-ablation-dataset}.

\paragraph{Experimental setup}
We fine-tuned multilingual translation models on our ablation dataset with different mixes of tasks.
As a baseline, we only trained on primary \st data (\engx + \xeng), optimizing L1: $\mathcal L_\text{\st}$ exclusively.
With the data fixed, we experimented with two other objectives to optimize: (L2) with joint optimization of \mt and \st ($\mathcal L_\text{\st}+\mathcal L_\text{\mt}$) and (L3) with the additional knowledge distillation objective with \mt as the teacher and \st as the student.
We then added more data, namely \asr data and mined data respectively, and compared the performance of models trained with different objectives in the three data setups.

We initialized \xt models with our \wvbert speech encoder and  \nllbv \mt model. We fine-tuned all parameters in the speech encoder and text encoder, while only fine-tuning LayerNorms and Self-attentions in the text encoder (LNA~\citep{li2020multilingual}). We trained all models for 100K updates (corresponding to 5-7 epochs). To regularize our models, we applied LayerDrop ($p{=}0.1$) to the speech encoder with masking ($p{=}0.1$). For the text encoder-decoder, we applied regular dropouts ($p=0.1$). We evaluated the last checkpoint on development data and evaluated \bleu scores on \fleurs dev for translation tasks (including \mt) and Whisper-style normalized \wer for \asr.
\paragraph{Results}
Within each data setup (D1, D2 or D3), we see in \Cref{tbl:modeling:ablation:multitask} that adding \mt to the multitasking loss, as expected, helps the performance on \mt (+1.8 \bleu on average D1,2,3). Without adding this loss, fine-tuning exclusively on \st leads to catastrophic forgetting of the pre-training \mt task (comparing L1 to L2). However, the accuracy of \st is seldom affected by this joint training with \mt. Knowledge distillation is proving to be a necessary ingredient to leverage joint fine-tuning with \mt. After adding knowledge distillation (L1 to L3), \st's performance improves by 0.6 \bleu points on average (D1,2,3).

If we compare the three different data setups, adding \asr data is crucial to supporting the \asr task as evaluating it as zero-shot leads to $3\times$ higher error rates. Joint fine-tuning with \mt and the auxiliary knowledge distillation loss has no negative effect on \asr given that for \asr data, the teacher task is auto-encoding (see \Cref{sec:modeling:x2t:stage12}). Adding mined \st data for which the source text is not available for \mt to teach \st, still helps \st in the M3 task mix.
We note, however, that the accuracy of \mt drops as we add more speech-text only data (\asr and mined \st) without the aligned text-text data.

\begin{table}[tb]
    \centering
    \footnotesize
    \begin{tabular}{@{}lccccccccc@{}}
        \toprule
        Data & \multicolumn{3}{c}{\bf D1: \st data} & 
        \multicolumn{3}{c}{\bf D2: D1+\asr data} & 
        \multicolumn{3}{c}{\bf D3: D2+Mined data}  \\
        \cmidrule(r){2-4}
        \cmidrule(lr){5-7}
        \cmidrule(l){8-10}
        Task & \st & $\text{\asr}^*$ & \mt & \st & \asr & \mt & \st & \asr & \mt \\
         & \it (n=28) & \it (n=15) & \it (n=28)
         & \it (n=28) & \it (n=15) & \it (n=28)
         & \it (n=28) & \it (n=15) & \it (n=28)\\  
        Metric & 
        \scriptsize $\uparrow$\bleu & 
        \scriptsize $\downarrow$\wer & 
        \scriptsize $\uparrow$\bleu 
        &  \scriptsize $\uparrow$ \bleu &
        \scriptsize $\downarrow$\wer & 
        \scriptsize $\uparrow$\bleu
        &  \scriptsize $\uparrow$\bleu
        &\scriptsize $\downarrow$\wer &
        \scriptsize $\uparrow$\bleu\\
        \midrule
        L1: $\mathcal L_\text{\st}$ & 26.5 & 36.5 & 34.1 & 26.7 & 16.4 & 34.2 & 27.6 & \bf 15.8 & 34.7 \\
        L2: L1 + $\mathcal L_\text{\mt}$ & 26.6 & 36.4 & \bf 36.8 & 26.7 & 16.8 & \bf 36.1 & 27.6 & 16.3 & \bf 35.4 \\
        L3: L2 + $\mathcal L_\text{KD}$ & \bf 27.1 & \bf 35.9 & 36.7 & \bf 27.2 & \bf 16.2 & \bf 36.1 &\bf 28.3 & \bf 15.8 & 35.3 \\
        \bottomrule
    \end{tabular}
    
    \caption{Ablations on multitasking objectives in three different data setups. Results are reported on \fleurs dev. 
    }
    \label{tbl:modeling:ablation:multitask}
\end{table}
       
\subsubsection{Leveraging Mined Speech-Text Data}\label{sec:modeling:ablation:data}
\paragraph{Experimental setup}
We fine-tuned \st models on increasing amounts of mined data from \mineddata. On top of the primary \st data, in the first model, we add 200 hours of mined data in each direction, 400 hours in the second and 600 hours in the last.
\mineddata is ranked based on \sonar scores and we selected the top ranking pairs up to the desired amount of additional data.
\paragraph{Results}
\Cref{tbl:modeling:ablation:mined} reports the results of models trained with increasing amounts of mined data. The model trained with at most 400 hours in each direction achieves the best average \bleu score. This signals that some filtering of \mineddata—e.g., based on \sonar similarity scores—can improve the quality of the model's translations without inflating the computational cost of training. 
\begin{table}[htb]
    \centering
    \begin{tabular}{@{}lcccc@{}}
        \toprule
         
        & \multicolumn{2}{c}{\makecell[c]{\bf\xeng\\\it(n=14)}} 
        & \multicolumn{2}{c}{\makecell[c]{\bf \engx\\\it(n=14)}} \\
        \cmidrule(r){2-3}
        \cmidrule(l){4-5}
        {\bf Data setting} & \scriptsize $\uparrow$\bleu &      \scriptsize  $\Delta$  
               & \scriptsize $\uparrow$\bleu & \scriptsize  $\Delta$ \\
        \midrule
        Baseline  & 23.9 & & 29.0 & \\
        \phantom{ab} + 200H mined & 25.9 & +2.0  
                                    & 29.4 & +0.4 \\
        \phantom{ab} + 400H mined & \bf 26.6 & \bf +2.7 
                                    & \bf 29.8 & \bf +0.8 \\
        \phantom{ab} + 600H mined & 26.0 & +2.1 
                                    & 29.5 & +0.5 \\
        \bottomrule
    \end{tabular}
    \caption{Ablations on the use of mined data. Results are reported on \fleurs dev. 
    }
    \label{tbl:modeling:ablation:mined}
\end{table}
\subsubsection{T2U pre-training in \unity}\label{sec:modeling:ablation:t2u}
\paragraph{Experimental setup}
Similar to the ablation dataset described in \Cref{sec:modeling:ablation:multitask}, we built an \sst ablation dataset with pseudo-labeled \sst data (\engx + \xeng) to fine-tune multilingual \unity models. 
With the data fixed, we compare two options for using pre-trained components when fine-tuning \unity. 
In the first (M1), we initialized the speech encoder with its adaptor and the first pass decoder with a pre-trained \xt model.
In the second (M2), we additionally initialized the T2U of \unity with a pre-trained T2U model.
In both setups, we only fine-tuned the weights of the T2U model on \sst data.

\paragraph{Results}
We evaluated our models on \fleurs dev for \sst and report \asrbleu scores in \Cref{tbl:modeling:ablation:s2st}. We note that T2U pre-training is beneficial for the fine-tuning of \unity as it converges faster (comparing \asrbleu scores after 10K updates) and is, therefore, more computationally efficient.

\subsubsection{Leveraging mined speech-to-speech data}
To measure the impact of adding mined \sst data to $\text{Stage}_3$ of \unity fine-tuning, we compared model M2 from \Cref{sec:modeling:ablation:t2u} to a model trained following the same training procedure, but with more mined data from \mineddata (see amounts of additional data per direction in \Cref{tbl:s2t-ablation-dataset}.
\paragraph{Results}
The results in \Cref{tbl:modeling:ablation:s2st} show that adding mined data improves \engx translation accuracy by 0.5 \asrbleu points, but it decreases that of \xeng by 0.2. However, we do notice slight improvements in the quality of the speech generated and hence add \mineddata for the final model training.

\begin{table}[htb]
    \centering
    \small
    \begin{tabular}{@{}ccccc@{}}
        \toprule
        & & & \multicolumn{2}{c}{\makecell[c]{\bf \fleurs S2ST\\\bf($\uparrow$\asr-BLEU)}}\\
        \cmidrule{4-5}
        & {\bf Model} & {\bf updates} & \xeng & \engx \\
        \midrule
        \multirow{2}{*}{M1} & \multirow{2}{*}{T2U from scratch} &10K & 6.9 & 1.8 \\
         & &  20K & 23.3 & 12.4 \\
        \midrule
        \multirow{3}{*}{M2} &  \multirow{3}{*}{pre-trained T2U}  & 10K & 18.1 & 8.8 \\
         &  & 20K & 24.2 & 15.2 \\
         &  & 50K$^\dagger$ & \bf 26.5  & 18.6   \\
        \midrule
        M3 & pre-trained T2U + Mined data & 80K$^\dagger$ & 26.3  & \bf 19.1  \\
        \bottomrule
    \end{tabular}
    \caption{Ablations on pre-training \unity's T2U and use of \sst mined data. Results are reported on \fleurs dev. 
    $^\dagger$ 80K and 50K correspond to 2 epoches in the two different data settings.
    }
    \label{tbl:modeling:ablation:s2st}
\end{table}

\subsection{Related work}

\paragraph{Two-pass sequence generation}
Two-pass decoding has the advantage of maintaining end-to-end optimization capability while inheriting the benefits of a cascading approach. \cite{NIPS2017_c6036a69} and \cite{9053606} incorporate an additional search process to find a better output. \cite{dalmia-etal-2021-searchable} re-ranks the intermediate hypotheses using an external module such as a language model.
\cite{zhao19d_interspeech} injects specific information in the intermediate decoder to bias the output toward the desired domain.
\cite{sainath2019two} provides an intermediate output to users before generating the final output for streaming applications. The two-pass approach makes the optimization tractable and results in better speech translation performance~\citep{8682801,anastasopoulos-chiang-2018-tied}.

\paragraph{Codec-based audio modeling}
In contrast to acoustic units extracted from SSL-based  audio representation models (e.g., \xlsr in this work),
recent advances in quantized, audio codec auto-encoders enabled successful research combining large, autoregressive language models and audio data.
Open-source EnCodec~\citep{Defossez2022HighFN} and proprietary SoundStream~\citep{Zeghidour2022SoundStreamAE} models are widely known examples of quantized audio auto-encoders.
One advantage of codec-based units is that they can be converted back to the waveform without needing an externally trained vocoder. 

In speech translation research,
VaLLE~\citep{Wang2023NeuralCL} introduced the conditional autoregressive modeling of EnCodec-based audio data to perform text-to-speech synthesis. VaLLE-X~\citep{Zhang2023SpeakFL} subsequently built upon VaLLE to scale language coverage and enable language translation using a model cascade. VIOLA~\citep{Wang2023-iv} later explored the ability of decoder-only codec-based LM to translate without cascades.

\paragraph{Multimodality \& multitask for speech \& text}
Multimodality and multitask on the source side are orthogonal to multitask learning with two-pass decoding, where the goal is to provide the second task with higher-level representations produced from the first task decoder~\cite{anastasopoulos-chiang-2018-tied}.

In general, multitask learning aims to improve generalization by leveraging domain-specific information contained in the training signals of related tasks \citep{caruana1997multitask,vandenhende2021multi}. Compared with single tasks, multitasking has the potential to improve performance by sharing complementary information or acting as a regularizer.  \cite{maninis2019attentive}, \cite{liu2019end}, and \cite{pfeiffer-etal-2020-mad} introduced task-dependent components to enhance individual task performance. \cite{weiss2017sequence} explored different multitask training strategies for speech translation, and they find the one-to-many strategy, in which an encoder is shared between the speech translation and \asr tasks, is more effective.  \cite{Bahar2019ACS} and \cite{tang-etal-2021-improving} compared different multitask strategies for \st, and confirmed the effectiveness of many-to-one training, in which \mt and \st are trained together and the decoder is shared between two tasks.

Recent works have also trained multitask and multimodal encoders by learning joint representations of multiple modalities. The motivation is that the learned features will be richer and that inter-modal tasks can benefit from such joint training.
These techniques were explored in audio~\citep{Maestro,mslam,google_usm,Rubenstein2023AudioPaLMAL}, in vision~\citep{chen2020uniter,gan2020large,fu2021violet}, as well as audiovisual~\citep{shi2022learning,anwar2023muavic}.

\section{Automatic and Human Evaluation}
\label{sec:appendix:evaluation}

Up to this point, to evaluate our model, we have used standard automatic evaluation metrics for each particular task as reported in Table \ref{tbl:eval:metrics}. In this section, for the tasks of \st and \sst, we extend beyond these standard automatic metrics to include additional automatic and human evaluations to further assess our contributions. Automatic evaluations in this section reflect the robustness of our models in terms of noise and domains. Human assessment focuses on the preservation of speaker intention, as well as the subjective quality of the audio generated. To start, we introduce \blaser, a new, modality-agnostic evaluation metric that enables quality estimation for both speech and text.

\subsection{Modality-Agnostic Automatic Metric: \blaser} 
\label{sec:eval:blaser}

\paragraph{Description} \blaser is the new version of BLASER~\citep{chen-etal-2023-blaser}, which works with both speech and text modalities—hence being modality-agnostic. Like the first version, our approach leverages the similarity between input and output sentence embeddings. The new version uses SONAR embeddings (\ref{sec:appendix:data:sonar}), supports 83 languages in the speech modality and 200 in text, and is extendable to future encoders for new languages or modalities that share the same embedding space.
For the purposes of evaluating speech outputs (and unlike ASR-based metrics), BLASER offers the advantage of being text-free. 

More specifically, in \blaser, we take the source input, the translated output from any \sst, \st, or \mt model, and the reference speech segment or text, and convert them into SONAR embedding vectors ($h_\text{src}$, $h_\text{mt}$, and $h_\text{ref}$, respectively). For the supervised version of \blaser, these embeddings are combined and fed into a small, dense neural network that predicts an XSTS score for each translation output. For the unsupervised version, we use, similar to \cite{chen-etal-2023-blaser}, the average of source-translation and reference-translation cosine similarities.

In addition, we trained a reference-free version of the system called \blaser-QE (for Quality Estimation). \blaser-QE is a supervised model trained only with source and translation embeddings. It can be applied in settings where reference translations are missing or if their quality is questionable.

\paragraph{Data} The supervised version of \blaser was trained on the XSTS-annotated data (\cite{licht2022xsts}), which includes the same \sst annotations as in the original BLASER (\cite{chen-etal-2023-blaser}). Additional \sst, \st, and T2ST annotations come from a variety of other internal studies, including NLLB human evaluations \cite{nllb2022}, and 
\mt annotations are drawn from NLLB (\cite{nllb2022}). We filtered out all audio longer than 30 seconds because the SONAR encoders were not trained on long audio.

For the original BLASER data, train/test splits were reused. The other datasets were split randomly in 80/20 proportion so that the same source audio or text always goes to the same partition. Details on the data are reported in Table \ref{tab:blaser-data}.

\begin{table}[htb]
    \centering
    \small
    \begin{tabular}{lrrrrrrr}
\toprule
{\bf Data part} &  {\bf test size} &  {\bf train size} &  {\bf systems} &  {\bf langs} &   $\mathbf{\rho_{unsup.}}$&   $\mathbf{\rho_{sup.}}$ &  $\mathbf{\rho_{QE}}$ \\
\midrule
BLASER 1.0 \sst data    &       9804 &       10690 &       10 &      9 &                  0.51 &                0.56 & 0.53 \\
Other \sst data     &       5453 &       15904 &        8 &     13 &                  0.47 &                0.48  & 0.38\\
\st and T2ST data &       5205 &       10246 &        7 &      8 &                  0.49 &                0.54 & 0.51 \\

\mt data           &      20311 &       86776 &        2 &     59 &                  0.49 &                0.61 & 0.60\\
\midrule
{\bf All data}          &      40773 &      123616 &       24 &     62 &                  0.51 &                0.59 & 0.56 \\
\bottomrule
\end{tabular}
\caption{The data for \blaser: test and train size, number of systems and languages, Spearman correlation of unsupervised, supervised, and reference-free \blaser scores with XSTS labels on the test subset.}
\label{tab:blaser-data}
\end{table}

\paragraph{Training} 

For the supervised model, the architecture is the same as for the BLASER 1.0 model: a 3-layer perceptron with $tanh$ activations on top of 6 concatenated vectors of normalized embeddings and their derivatives: $[h_{ref};h_{mt};h_{src}\odot h_{mt};|h_{src}-h_{mt}|; h_{ref}\odot h_{mt};|h_{ref}-h_{mt}|]$. For the QE version, we used the same settings but with reference-free inputs: $[h_{src};h_{mt};h_{src}\odot h_{mt};|h_{src}-h_{mt}|]$.

We used the training code for BLASER 1.0 with a few modifications in the hyperparameters intended to mitigate overfitting: 50\% dropout, 0.1 weight decay, batch size of 1024, and full linear decay of the learning rate by the end of the training. To compensate for the increased batch size, we trained for 50 instead of 20 epochs.

\begin{table}[htb]
    \centering
    \small
\begin{tabular}{lrrrrrrr}
\toprule
& \multicolumn{6}{c}{\bf $\uparrow$Pearson Correlation}\\\cmidrule{2-7}
{\bf Model} &  eng-deu &  eng-spa &  eng-fra &  spa-eng &  fra-eng &  rus-eng &  {\bf average} \\ 
\midrule
BLASER 1.0 unsup & 0.32 & 0.58 & 0.64 & 0.50 & 0.48 & 0.43 & 0.49 \\
\blaser unsup & \textbf{0.37} & 0.75 & 0.71 & 0.59 & 0.57 & 0.49 & 0.58 \\
\blaser QE    & 0.34 & 0.73 & 0.71 & 0.54 & 0.48 & 0.45 & 0.54 \\
BLASER 1.0 sup   & 0.33 & 0.75 & 0.71 & 0.58 & \textbf{0.57} & \textbf{0.53} & 0.58 \\
\blaser sup   & 0.36 & \textbf{0.75} & \textbf{0.73} & \textbf{0.58} & 0.56 & 0.50 & \textbf{0.58} \\
\bottomrule
\end{tabular}
\caption{Pearson correlations of unsupervised and supervised BLASER models with XSTS scores on the BLASER 1.0 test data.}
\label{tab:blaser-corrs}
\end{table}

\paragraph{Results} Table \ref{tab:blaser-corrs} presents the performance of unsupervised and supervised BLASER on the BLASER 1.0 test data. The unsupervised 2.0 model slightly outperforms its predecessor. The supervised v1.0 and v2.0 models have the same average correlation with human judgments. Because \blaser supports more languages, we used this for evaluations.
 
The last three columns in Table \ref{tab:blaser-data} present correlations of the 2.0 model's predictions with XSTS scores for all data partitions. Based on the results, the supervised model outperforms the unsupervised on each partition. The reference-free model scores between them in most cases, but for the new \sst data, its performance is below that of the unsupervised model. We hypothesized that on this subset, references sometimes diverge from the sources, either due to errors of speech segmentation or synthesis, or due to non-literal translation that makes sense only in the context. A manual examination of a few samples corroborates this hypothesis, but more analysis of the role of reference in BLASER models is required in the future.
Full \blaser scores for \mfourt models are reported in table \ref{tbl:modeling:spbleu}. Additionally, the next section \ref{subsec:appendix:humanmetrics} reports the corresponding correlations of \blaser scores with human scores.

\subsection{Human Evaluation}
\label{subsec:appendix:humanmetrics}

Human evaluation is a vital tool in assessing the quality of our systems. We first briefly describe related work in the area, followed by a detailed description of the entire human evaluation process, including protocols, data, and calibration. 

\paragraph{Related work.} Human evaluation has been widely applied to machine translation in the scientific community. Two of the most popular models of human evaluation are deployed within the context of International Evaluation Campaigns. The WMT conference \citep{kocmi-EtAl:2022:WMT} asks participants to evaluate the outputs of translation systems using a pre-defined protocol, typically that of Direct Assessment \citep{grahametal2013directassessment}.  
Beyond this text-based evaluation, the IWSLT Evaluation Campaign covers speech translation. As an example, the speech-to-speech track\footnote{https://iwslt.org/2023/s2s} evaluates speech output quality in four dimensions. The first one is translation quality, which focuses on capturing meaning, and annotators rank target audio between 1 and 5. The rest of the dimensions cover naturalness, including voice and pronunciation, clarity of speech for comprehension, and sound quality, which takes into account noise and other artifacts. These criteria are used as an alternative to the Mean Opinion Score (MOS).

\subsubsection{Human Evaluation Protocols} 

Similar to the related work aforementioned, for \st evaluation, we used the XSTS protocol to assess translation quality. For the \sst task, we evaluate using two protocols: XSTS for translation quality, and MOS to assess naturalness. 

\paragraph{XSTS.}  XSTS \citep{licht2022xsts} evaluates translation quality in terms of semantic meaning preservation, and has previously been used to evaluate the NLLB models \citep{nllb2022}.  While XSTS was originally designed to evaluate text, the protocol is effectively modality agnostic, and we required only small adaptations in order to support \sst and \st tasks.   For instance, the \sst and \st versions of the protocol required additional instructions for annotators regarding the treatment of non-speech tags (e.g. \texttt{<laugh>}), which the annotators were instructed to ignore, and how they should consider pauses and non-speech noises (they are instructed to ignore these as well).

On the execution logistics side, conversations with vendors used for our evaluation work indicated that the evaluation of \sst translations seemed to require a higher cognitive load for the annotators than \mt (as a result of not being able to experience the source and target simultaneously), and thus was slower to conduct.

\paragraph{XSTS annotation and calibration process.} During annotation, 3 annotators examined each source-target audio pair (or audio-text pair) and evaluated the item for semantic similarity using the XSTS protocol. Prior to annotating, all annotators went through a set of monolingual English `practice' evaluations with score justifications.  To expedite evaluation, more than 3 (up to 30 in our case) annotators were used per language pair; each evaluated sentence pair was shown to 3 annotators, assigned essentially randomly, and with calibration set items intermixed in the evaluation. In cases where the 3 annotators had a disagreement of score values of 2 or more, 2 additional annotators evaluated the same item again, bringing the total to 5 evaluator scores for those items. The median score over annotators of the same audio pair was then taken for each evaluation sentence pair; the median is used for robustness.  The process was the same for both \sst and \st evaluations. For overall direction scores, we report the mean of this median score (or some aggregate, such as the fraction of sentences with a median XSTS score above a given threshold) across all evaluated items in the dataset generated by a particular system in a language direction. Calibration set items received the exact same treatment, resulting in 1 score per sentence pair per annotator pool, and language-level scores were calibrated using the mean score on the calibration set for the crew of annotators evaluating a given language direction; the calibration set and methodology is described below.

To enable interlanguage comparison of model quality, a mono-lingual "cross-lingual calibration set" \citep{licht2022xsts} was generated and included in the evaluation, and scores were calibrated using the `moderated calibration' methodology established previously \citep{licht2022xsts,nllb2022}.  The calibration process was found to reduce language-level annotator biases and has been shown to improve correlation with automatic metrics as a result. %
Running a calibration set or `gold set' of items with a known score, even one much-reduced in size (e.g., 50–100 items instead of the 500 here) is useful as a diagnostic tool for ensuring annotation quality, even if one is not intending on doing interlanguage calibration.  Annotator crews sufficiently `out of calibration' can be identified, and their results excluded, or additional training can be conducted to improve their performance.  

\paragraph{MOS.} To evaluate and compare generated speech quality for the \sst task, we used a standard Mean Opinion Score (MOS) protocol to evaluate naturalness, quality of sound, and clarity of speech of our generated audio. MOS methodology has been in use in the telecommunications industry for decades and it was standardized in Recommendation ITU-T P.800. 
In particular, the adapted guidelines that we used for the MOS evaluation include the following instructions:
\begin{enumerate}
\item \textbf{How clear is the speech? }
Recordings with clear speech and no mumbling and unclear phrases should be given a high score. Recordings with a large amount of mumbling and unclear phrases should be given a low score.
\item \textbf{How good is the sound quality?}
Recordings with clean audio and no noise and static in the background should be given a high score. Recordings with a large amount of noise and static in the background should be given a low score.
\item \textbf{How natural is the speech?}
Recordings that sound human-like, with natural-sounding pauses, stress, and intonation, should be given a high score. Recordings that sound robotic, flat, or otherwise unnatural should be given a low score. 
\end{enumerate}

For each of these questions, annotators are instructed to provide a Likert score (1-5). The MOS evaluations were conducted on the same generations as those evaluated with the XSTS protocol, with the exclusion of the cross-lingual calibration set items, and we elected to only evaluate a more limited sample of language directions into English (as MOS characteristics were not expected to be strongly dependent on the input language). Similar to XSTS, 3 annotators scored each generation for most languages evaluated, with some languages (\texttt{cat}, \texttt{fin}, \texttt{hin}, \texttt{ind}, \texttt{ita}, \texttt{kor}, \texttt{por}, \texttt{ron}, \texttt{swh}, \texttt{tel}, \texttt{tha}, \texttt{tur}, \texttt{urd}, \texttt{vie}) only receiving scores from a single annotator per item (this balanced the need for language coverage with budgetary constraints). The median of annotator scores is collected for each item, and aggregate scores are presented in the paper. Unlike XSTS, additional annotators were not consulted when scores disagreed and calibration was not performed.

\subsubsection{Evaluation Framework}

\paragraph{Dataset} Human evaluations were conducted utilizing the `test' partition of the \fleurs dataset \citep{fleurs2022arxiv}.  The \fleurs `test' partition provides up to 350 sentences sourced from the FLORES-101 dataset \citep{flores101:2021} for each supported language (\fleurs supports 102 languages). Each sentence has up to 3 recorded audios spoken by different speakers (depending on which recordings passed quality review), along with the associated FLORES-101 text. The quality review requirement means that each language may not have a recording for all 350 sentences and that for those sentences that do have recordings, not all three speaker recordings may be present.

When conducting an evaluation of a translation system for a particular language direction, we filtered down the \fleurs data to a subset of sentences that have recordings in both languages in order to have a common, bidirectional evaluation set per-language pair. We do this to ensure \st and \sst evaluations both use an identical set of sentences. Because the coverage of \fleurs varies per language, the subset of items present in the evaluation set varies per language and thus also per language pair; though there is a majority of items common across languages, and we believe the scores to be largely comparable as they were drawn from the same domain.  

When preparing \fleurs to be used as a human reference set, pairings had to be made between distinct readers in the source language and readers of the equivalent \fleurs item in the target language.  When possible, these pairings were made to match user gender (53\% of the time over the entirety of the \fleurs test partition, varying significantly between languages paired with English), and mixed-gender matches had to be made for the remaining 47\% of items. We elected to limit human evaluation to 2 unique readings per \fleurs sentence at most.  %

\newcommand{\evallangs}{{24}}
\begin{table}
\centering

\small
\begin{threeparttable}
\begin{tabular}{llllp{3cm}}
\toprule
{\bf Modality} & {\bf Protocol} & {\bf Direction} & {\bf Systems} & {\bf Languages} \\
\midrule
\st & XSTS & \xeng & 
\makecell[l]{\whisperlarge\\\mfourtlg$^*$\tnote{1}} &
\evallangs\tnote{2}
\\ 
\st & XSTS & \engx & \mfourtlg$^*$ &
\evallangs
\\
\midrule
\sst & XSTS &\xeng &
\makecell[l]{\whisperlarge+\yourtts\\\mfourtlg} & 
\evallangs
\\
\sst & XSTS & \engx  & \mfourtlg &
\evallangs
\\
\midrule 

\sst & MOS  & \xeng & 
\makecell[l]{\whisperlarge+\yourtts\\\mfourtlg} & 
8 (arb, cmn, fra, hin, rus, spa, tel, tur) \\
\sst & MOS  & \engx & \mfourtlg &
\evallangs
\\
\bottomrule
\end{tabular}
\begin{tablenotes}
\item[1] \mfourtlg$^*$ refers to the \mfourtlg model using \texttt{fairseq} for generations instead of \fairseq, but \st performance was on average within 0.5 \bleu between the two.
\item[2] Bengali, Catalan, Dutch, Finnish, French, German, Hindi, Indonesian, Italian, Japanese, Korean, Mandarin Chinese, Modern Standard Arabic, Portuguese, Romanian, Russian, Spanish, Swahili, Tagalog, Telugu, Thai, Turkish, Urdu, and Vietnamese.
\end{tablenotes}
\end{threeparttable}
\caption{\label{table:evalsummary} Summary of evaluations: languages, modalities, models, and protocols used in human evaluations. Modalities and protocols in parentheses are not presented in this paper but will be shared in a later update.}
\end{table}

\paragraph{Language directions, modalities, and systems evaluated} We list the languages and modalities evaluated with each protocol in Table \ref{table:evalsummary}. Language selections were made by balancing a mix of resource availability for human annotations. The final language sample captures a large human population while also representing a mix of high and mid-resource languages.

For the \st and \sst tasks, we have XSTS evaluations of \evallangs{} languages in the \xeng direction for both the \mfourtlg and \whisperlarge models, where generations from the \mfourtlg model were made using a slightly earlier version via \texttt{fairseq} (instead of \fairseq) but \st performance has less than 0.5 \bleu average difference. For \engx, we have evaluations for the same languages but only for the \mfourtlg model (with \texttt{fairseq} generations). Additionally, we have evaluations for a human reference system (i.e. the \fleurs data itself) for all languages in each direction.

We only evaluated direct models for \st and plan to extend the benchmarking to 2-stage cascaded systems in future work for \engx. For \sst, we do not evaluate \engx benchmarks due to the complexity involved in running monolingual TTS models for all the target directions. However, in the future, we plan to build such baselines using systems like MMS-TTS \cite{mms}; we have experimented using these same systems in other sections of this paper, e.g. for the purpose of extending text-based responsible AI datasets to speech (Section \ref{subsec:biasexperimental}). 

For the \sst task, we used the MOS protocol to evaluate for all \evallangs{} in the \engx direction. Since we do not expect MOS scores to be strongly sensitive to input language when evaluating English output speech, we evaluated a smaller language set in the \xeng direction (only 8 languages).

\subsubsection{Human Evaluation Results}

\paragraph{XSTS results for the \st task} %

\begin{table}
\begin{tabular}{rrccc}
\toprule
{\bf Direction} & {\bf System} & {\bf Avg. XSTS (\st)} & {\bf \% 3+} & {\bf \% 4+} \\
\midrule
\engx & Human reference & 4.69 & 95.98 & 78.66 \\
 & \mfourtlg & 4.53 & 87.69 & 73.28 \\
\midrule
& Human reference & 4.67 & 95.23 & 76.86 \\
\xeng & \whisperlarge & 4.05 & 70.11 & 58.00 \\
& \mfourtlg & {\bf 4.16} & {\bf 72.51} & {\bf 59.86} \\
\bottomrule
\end{tabular}
\caption{For \st task: overall average XSTS human evaluation results into and out of English, over all \evallangs{} evaluated languages.  Results were computed for each language direction (see Table \ref{table:fullXSTSresultsst} for full language-level results).  \%3+ and  \%4+ refer to the percent of a language's evaluated sentences with median scores equal to or greater than 3 and 4 respectively.}
\label{tab:XSTShighlevelsummaryst}
\end{table}

\begin{table}
\centering
\begin{threeparttable}
\scriptsize
\begin{tabular}{ll|ccc|ccc|ccc|c}
\toprule
\textbf{Direction} & \textbf{Lang} & \multicolumn{3}{c}{\textbf{XSTS (calibrated)}} & \multicolumn{3}{c}{\textbf{\% XSTS 3+}} & \multicolumn{3}{c}{\textbf{\% XSTS 4+}} & \textbf{Items} \\
 &  & Seamls\tnote{1} & Wspr\tnote{2} & Hum\tnote{3} & Seamls & Wspr & Hum & Seamls & Wspr & Hum &  \\
\midrule
\engx & arb & 4.53 & -- & 4.61 & 90.2 & -- & 96.8 & 71.7 & -- & 72.9 & 410 \\
 & ben & 4.35 & -- & 4.49 & 92.1 & -- & 96.8 & 64.2 & -- & 73.0 & 629 \\
 & cat & 4.73 & -- & 4.82 & 90.1 & -- & 95.9 & 82.6 & -- & 85.0 & 638 \\
 & cmn & 4.03 & -- & 4.69 & 70.1 & -- & 98.0 & 41.7 & -- & 69.7 & 636 \\
 & deu & 4.61 & -- & 4.75 & 88.4 & -- & 96.1 & 72.4 & -- & 77.5 & 612 \\
 & fin & 4.37 & -- & 4.66 & 75.5 & -- & 93.0 & 52.1 & -- & 61.2 & 632 \\
 & fra & 4.76 & -- & 4.83 & 92.0 & -- & 98.5 & 83.3 & -- & 83.3 & 538 \\
 & hin & 4.55 & -- & 4.51 & 98.2 & -- & 97.4 & 77.6 & -- & 74.5 & 388 \\
 & ind & 4.81 & -- & 4.84 & 97.6 & -- & 98.5 & 83.8 & -- & 85.8 & 544 \\
 & ita & 4.59 & -- & 4.56 & 97.1 & -- & 97.4 & 93.8 & -- & 90.2 & 612 \\
 & jpn & 4.09 & -- & 4.74 & 66.7 & -- & 92.6 & 53.9 & -- & 79.8 & 514 \\
 & kor & 4.58 & -- & 4.76 & 85.7 & -- & 94.7 & 66.3 & -- & 76.7 & 356 \\
 & nld & 4.72 & -- & 4.67 & 88.4 & -- & 94.0 & 78.2 & -- & 62.7 & 335 \\
 & por & 4.79 & -- & 4.8 & 95.7 & -- & 97.6 & 92.2 & -- & 90.8 & 632 \\
 & ron & 4.71 & -- & 4.88 & 89.3 & -- & 98.4 & 77.9 & -- & 86.8 & 619 \\
 & rus & 4.56 & -- & 4.76 & 82.9 & -- & 94.3 & 73.2 & -- & 78.7 & 597 \\
 & spa & 4.69 & -- & 4.59 & 89.1 & -- & 96.5 & 82.0 & -- & 62.9 & 623 \\
 & swh & 4.52 & -- & 4.81 & 82.0 & -- & 93.1 & 76.6 & -- & 90.8 & 466 \\
 & tel & 4.5 & -- & 4.49 & 96.6 & -- & 97.3 & 89.8 & -- & 88.0 & 442 \\
 & tha & 4.18 & -- & 4.65 & 78.2 & -- & 94.7 & 46.5 & -- & 70.0 & 643 \\
 & tur & 4.61 & -- & 4.84 & 88.7 & -- & 97.9 & 74.7 & -- & 87.5 & 566 \\
 & urd & 4.39 & -- & 4.52 & 91.2 & -- & 95.1 & 68.9 & -- & 76.0 & 283 \\
 & vie & 4.63 & -- & 4.65 & 93.0 & -- & 95.6 & 85.1 & -- & 84.6 & 611 \\
 \midrule
\xeng & arb & 4.29 & 3.82 & 4.59 & 81.7 & 62.7 & 95.1 & 62.7 & 43.4 & 72.7 & 410 \\
 & ben & 4.03 & 2.84 & 4.51 & 78.1 & 37.0 & 97.6 & 51.8 & 16.2 & 74.7 & 629 \\
 & cat & 4.72 & 4.59 & 4.79 & 89.3 & 84.2 & 95.3 & 81.8 & 76.8 & 82.0 & 638 \\
 & cmn & 3.75 & 4.04 & 4.55 & 58.5 & 67.8 & 94.0 & 37.1 & 48.1 & 59.7 & 636 \\
 & deu & 4.64 & 4.69 & 4.81 & 88.4 & 90.2 & 97.5 & 76.6 & 79.7 & 82.7 & 612 \\
 & fin & 4.07 & 3.71 & 4.66 & 61.2 & 46.7 & 92.4 & 44.8 & 30.4 & 61.6 & 632 \\
 & fra & 4.73 & 4.74 & 4.82 & 90.0 & 90.5 & 97.8 & 81.4 & 82.0 & 81.6 & 538 \\
 & hin & 4.31 & 4.25 & 4.55 & 87.4 & 85.8 & 96.6 & 71.9 & 70.4 & 80.7 & 388 \\
 & ind & 4.57 & 4.53 & 4.81 & 88.2 & 87.5 & 98.0 & 74.3 & 72.2 & 84.0 & 544 \\
 & ita & 4.55 & 4.61 & 4.61 & 94.8 & 97.1 & 98.5 & 92.6 & 95.9 & 94.0 & 612 \\
 & jpn & 3.18 & 3.77 & 4.68 & 33.3 & 53.1 & 88.9 & 26.3 & 45.1 & 76.8 & 514 \\
 & kor & 4.26 & 4.65 & 4.74 & 72.8 & 90.4 & 96.3 & 50.0 & 70.8 & 72.2 & 356 \\
 & nld & 4.57 & 4.53 & 4.63 & 80.9 & 80.0 & 89.9 & 73.1 & 69.9 & 63.3 & 335 \\
 & por & 4.71 & 4.84 & 4.82 & 92.1 & 97.2 & 98.4 & 89.1 & 96.2 & 92.1 & 632 \\
 & ron & 4.46 & 4.46 & 4.84 & 77.2 & 76.6 & 98.1 & 66.4 & 66.7 & 81.7 & 619 \\
 & rus & 4.46 & 4.69 & 4.7 & 78.6 & 87.6 & 93.1 & 69.7 & 81.1 & 72.7 & 597 \\
 & spa & 4.59 & 4.82 & 4.53 & 86.8 & 94.7 & 95.5 & 77.2 & 88.8 & 58.4 & 623 \\
 & swh & 4.03 & 1.65 & 4.86 & 60.9 & 2.6 & 95.7 & 56.7 & 2.6 & 92.3 & 466 \\
 & tel & 3.84 & 3.14 & 4.49 & 74.4 & 52.9 & 97.3 & 60.4 & 38.7 & 88.2 & 442 \\
 & tha & 3.49 & 3.39 & 4.5 & 47.3 & 43.2 & 92.1 & 27.4 & 23.3 & 57.7 & 643 \\
 & tur & 4.16 & 4.48 & 4.8 & 70.5 & 84.1 & 97.7 & 53.9 & 68.7 & 82.9 & 566 \\
 & urd & 3.83 & 3.5 & 4.5 & 67.5 & 58.7 & 95.1 & 49.1 & 40.3 & 75.3 & 283 \\
 & vie & 3.45 & 3.63 & 4.53 & 46.3 & 56.1 & 91.2 & 38.3 & 44.8 & 77.7 & 611 \\
\bottomrule
\end{tabular}
\begin{tablenotes}
\item[1] \mfourtlg using \texttt{fairseq} generations
\item[2] \whisperlarge
\item[3] Human reference
\end{tablenotes}
\end{threeparttable}
\caption{\label{table:fullXSTSresultsst} Full calibrated XSTS \st results; bootstrapped 95\% CI widths are $\sim\pm 0.1$ on average.  \%3+ and \%4+ refer to the percent of a language’s evaluated sentences with (uncalibrated) median scores equal to or greater than 3 and 4 respectively.}
\end{table}

\begin{figure}
\centering
\includegraphics[width=0.8\textwidth]{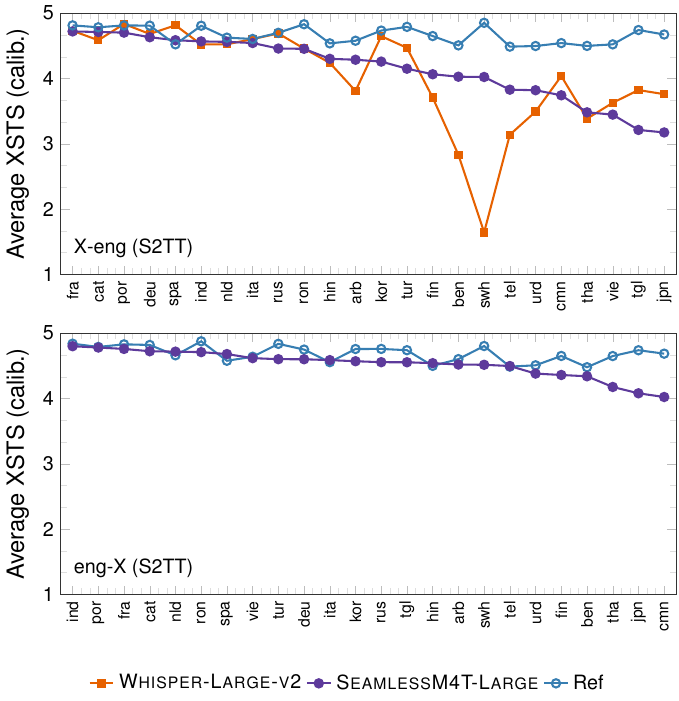}
\caption{\label{fig:XSTS_scores_cal_S2T} Language Direction level mean XSTS scores per direction for \st modality, after calibration. Bootstrapped 95\% CI is typically within $\sim\pm0.1$.}
\end{figure}

We present results for the \st task using the XSTS protocol (see Table \ref{table:evalsummary}). Figure \ref{fig:XSTS_scores_cal_S2T} shows calibrated XSTS scores on the language level for all models and languages evaluated (for both \xeng and \engx), which are also enumerated in Table \ref{table:fullXSTSresultsst} and summarized in Table \ref{tab:XSTShighlevelsummaryst}. For \xeng language directions, \mfourtlg quality was above an XSTS score of 3\footnote{The instructions surrounding an XSTS score of 3 are: ``The two sentences are mostly equivalent, but some unimportant details can differ.  There cannot be any significant conflicts in intent or meaning between the sentences, no matter how long the sentences are.''} for all \evallangs{} evaluated language directions. For \engx language directions, \mfourtlg was above an XSTS score of 4\footnote{The instructions for an XSTS score of 4 are: ``The two sentences are paraphrases of each other. Their meanings are near-equivalent,  with no major differences or information missing. There can only be minor differences in meaning due to differences in expression (e.g., formality level, style, emphasis, potential implication, idioms, common metaphors).''} for all \evallangs{} evaluated language directions.

Notably, in the \xeng direction, we see that \mfourtlg improves translation quality considerably over the \whisperlarge baseline for Swahili (an XSTS improvement of 2.38), Bengali (an XSTS improvement of 1.19), Telugu (an XSTS improvement of 0.69, and Modern Standard Arabic (an XSTS improvement of 0.47). \mfourtlg has significant improvements in language quality over \whisperlarge for 10 out of the \evallangs{} languages evaluated \xeng with regressions in 10, but only 2 languages, Japanese and Tagalog, have regressions of more than 0.5 XSTS.

When averaging over language directions, \mfourtlg demonstrates superior performance on both average XSTS score and percentage of sentences above XSTS thresholds of 3 and 4 compared to the \whisperlarge's baseline on \xeng (see Table \ref{tab:XSTShighlevelsummaryst}).

We also note generally higher performance in the \engx direction compared to the \xeng direction. From the automatic results in section \ref{subsec:multitaskingxtresults}, we observed that the higher performance in one direction or the other varies depending on the task (\st, \sst, or \mt). For \st, in terms of \spbleu and \blaser (see Table \ref{tbl:modeling:spbleu}), even when averaging over a larger set of languages, superior performance in \engx compared to \xeng holds. We offer a few possible explanations for this phenomenon. For example, speech encoding may be a more complicated task than speech or text decoding. If this is the case, better performance in English speech encoding could contribute to higher performance in the \engx direction. Data-wise, a plausible explanation could be a difference in audio quality of \fleurs recordings for different languages (e.g., English source sentence audio quality may have been higher, inflating the \engx scores), though evidence for this is only anecdotal. %

\paragraph{XSTS results for \sst task}

\begin{figure}
\centering
\includegraphics[width=0.8\textwidth]{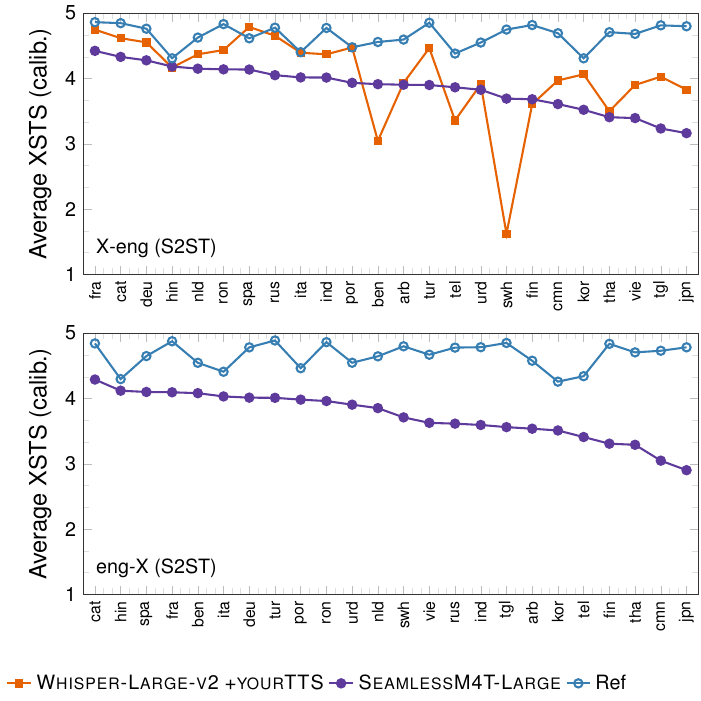}
\caption{\label{fig:XSTS_scores_cal_S2ST} Language Direction level mean XSTS scores per direction for \sst modality, after calibration. Bootstrapped 95\% CI is typically within $\sim\pm0.1$.}
\end{figure}

\begin{table}
\begin{tabular}{rrccc}
\toprule
{\bf Direction} & {\bf System} & {\bf Avg. XSTS (\sst)} & {\bf \% 3+} & {\bf \% 4+} \\
\midrule
\engx & Human reference & 4.67 & 95.38 & 79.45 \\
 & \mfourtlg & 3.73 & 58.10 & 36.90 \\
\midrule
 & Human reference & 4.66 & 95.20 & 78.89 \\
\xeng & \whisperlarge+\yourtts & {\bf 4.04} & {\bf 71.52} & {\bf 56.18} \\
 & \mfourtlg & 3.87 & 60.61 & 46.05 \\
\bottomrule
\end{tabular}
\caption{For \sst task: overall average XSTS human evaluation results into and out of English, over all \evallangs{} evaluated languages.  Results were computed for each language direction (see Table \ref{table:fullXSTSresultssst} for full language-level results).  \%3+ and  \%4+ refer to the percent of a language's evaluated sentences with median scores equal to or greater than 3 and 4 respectively.  Avg number of items is identical to those presented in Table \ref{tab:XSTShighlevelsummaryst}.}
\label{tab:XSTShighlevelsummarysst}
\end{table}

\begin{table}
\centering
\begin{threeparttable}
\scriptsize
\begin{tabular}{ll|ccc|ccc|ccc|c}
\toprule
\textbf{Direction} & \textbf{Lang} & \multicolumn{3}{c}{\textbf{XSTS (calibrated)}} & \multicolumn{3}{c}{\textbf{\% XSTS 3+}} & \multicolumn{3}{c}{\textbf{\% XSTS 4+}} & \textbf{Items} \\
 &  & Seamls\tnote{1} & Wspr\tnote{2} & Hum\tnote{3} & Seamls & Wspr & Hum & Seamls & Wspr & Hum &  \\
\midrule
\engx & arb & 3.55 & -- & 4.59 & 51.5 & -- & 95.4 & 24.9 & -- & 74.6 & 410 \\
 & ben & 4.09 & -- & 4.56 & 80.4 & -- & 96.8 & 41.0 & -- & 67.9 & 629 \\
 & cat & 4.3 & -- & 4.85 & 77.7 & -- & 98.3 & 64.3 & -- & 90.9 & 638 \\
 & cmn & 3.06 & -- & 4.74 & 31.4 & -- & 98.3 & 7.7 & -- & 72.2 & 636 \\
 & deu & 4.02 & -- & 4.79 & 63.4 & -- & 97.9 & 29.4 & -- & 72.4 & 612 \\
 & fin & 3.32 & -- & 4.84 & 32.3 & -- & 95.4 & 16.0 & -- & 85.9 & 632 \\
 & fra & 4.1 & -- & 4.88 & 64.5 & -- & 98.5 & 44.2 & -- & 87.2 & 538 \\
 & hin & 4.13 & -- & 4.3 & 94.6 & -- & 98.7 & 83.0 & -- & 95.1 & 388 \\
 & ind & 3.6 & -- & 4.8 & 58.1 & -- & 98.2 & 35.5 & -- & 93.0 & 544 \\
 & ita & 4.04 & -- & 4.42 & 86.8 & -- & 100.0 & 75.8 & -- & 96.1 & 612 \\
 & jpn & 2.91 & -- & 4.79 & 18.5 & -- & 91.6 & 11.1 & -- & 80.5 & 514 \\
 & kor & 3.52 & -- & 4.27 & 40.2 & -- & 77.0 & 13.8 & -- & 39.9 & 356 \\
 & nld & 3.86 & -- & 4.65 & 52.2 & -- & 90.4 & 32.8 & -- & 58.8 & 335 \\
 & por & 3.99 & -- & 4.47 & 81.6 & -- & 99.1 & 70.7 & -- & 94.6 & 632 \\
 & ron & 3.97 & -- & 4.87 & 63.3 & -- & 97.1 & 42.2 & -- & 89.2 & 619 \\
 & rus & 3.62 & -- & 4.78 & 40.4 & -- & 93.6 & 29.5 & -- & 75.0 & 597 \\
 & spa & 4.11 & -- & 4.66 & 64.2 & -- & 96.5 & 47.2 & -- & 63.7 & 623 \\
 & swh & 3.72 & -- & 4.8 & 45.1 & -- & 91.6 & 34.8 & -- & 84.5 & 466 \\
 & tel & 3.42 & -- & 4.35 & 60.0 & -- & 91.9 & 24.0 & -- & 61.8 & 442 \\
 & tha & 3.3 & -- & 4.72 & 46.7 & -- & 95.6 & 21.5 & -- & 86.2 & 643 \\
 & tur & 4.02 & -- & 4.89 & 65.4 & -- & 98.8 & 32.9 & -- & 87.8 & 566 \\
 & urd & 3.91 & -- & 4.56 & 71.7 & -- & 94.3 & 40.3 & -- & 71.0 & 283 \\
 & vie & 3.64 & -- & 4.68 & 64.5 & -- & 98.4 & 43.2 & -- & 91.0 & 611 \\
 \midrule
\xeng & arb & 3.91 & 3.94 & 4.61 & 62.2 & 68.0 & 96.1 & 48.5 & 50.7 & 75.6 & 410 \\
 & ben & 3.92 & 3.05 & 4.57 & 68.0 & 43.7 & 96.5 & 38.2 & 15.1 & 69.6 & 629 \\
 & cat & 4.34 & 4.63 & 4.86 & 77.7 & 88.7 & 98.4 & 68.8 & 82.4 & 91.4 & 638 \\
 & cmn & 3.62 & 3.98 & 4.7 & 51.1 & 67.6 & 97.0 & 26.7 & 36.6 & 68.6 & 636 \\
 & deu & 4.29 & 4.56 & 4.77 & 69.9 & 85.0 & 95.6 & 49.3 & 60.0 & 72.5 & 612 \\
 & fin & 3.69 & 3.62 & 4.82 & 44.3 & 44.9 & 94.6 & 33.1 & 29.1 & 84.7 & 632 \\
 & fra & 4.43 & 4.75 & 4.87 & 75.1 & 92.2 & 98.3 & 64.5 & 80.9 & 85.7 & 538 \\
 & hin & 4.19 & 4.18 & 4.32 & 95.6 & 95.6 & 99.0 & 88.4 & 89.7 & 95.4 & 388 \\
 & ind & 4.02 & 4.38 & 4.78 & 67.3 & 84.2 & 97.6 & 57.2 & 76.1 & 92.5 & 544 \\
 & ita & 4.02 & 4.4 & 4.41 & 82.0 & 98.4 & 99.3 & 76.3 & 96.4 & 96.1 & 612 \\
 & jpn & 3.17 & 3.84 & 4.81 & 26.7 & 51.0 & 92.8 & 20.2 & 39.9 & 81.3 & 514 \\
 & kor & 3.53 & 4.07 & 4.32 & 40.4 & 66.0 & 82.0 & 18.5 & 37.1 & 40.7 & 356 \\
 & nld & 4.16 & 4.38 & 4.63 & 58.5 & 71.9 & 89.6 & 49.9 & 54.6 & 55.8 & 335 \\
 & por & 3.94 & 4.48 & 4.49 & 74.7 & 98.7 & 99.4 & 70.1 & 96.2 & 95.9 & 632 \\
 & ron & 4.15 & 4.44 & 4.84 & 67.9 & 81.1 & 96.4 & 52.0 & 65.9 & 86.9 & 619 \\
 & rus & 4.06 & 4.66 & 4.78 & 55.4 & 82.6 & 92.5 & 46.1 & 74.2 & 76.2 & 597 \\
 & spa & 4.15 & 4.79 & 4.62 & 65.2 & 93.4 & 95.7 & 52.5 & 83.6 & 61.2 & 623 \\
 & swh & 3.7 & 1.63 & 4.76 & 41.2 & 1.3 & 89.7 & 32.6 & 0.9 & 80.0 & 466 \\
 & tel & 3.87 & 3.37 & 4.39 & 72.6 & 58.1 & 94.8 & 44.3 & 30.1 & 62.9 & 442 \\
 & tha & 3.42 & 3.51 & 4.72 & 48.8 & 56.0 & 95.6 & 31.9 & 33.4 & 86.6 & 643 \\
 & tur & 3.91 & 4.47 & 4.86 & 57.8 & 81.3 & 97.3 & 37.1 & 62.4 & 85.5 & 566 \\
 & urd & 3.84 & 3.92 & 4.56 & 66.8 & 72.4 & 92.6 & 40.6 & 48.4 & 72.1 & 283 \\
 & vie & 3.4 & 3.91 & 4.69 & 53.7 & 73.0 & 98.4 & 37.6 & 57.8 & 92.5 & 611 \\
\bottomrule
\end{tabular}
\begin{tablenotes}
\item[1] \mfourtlg
\item[2] \whisperlarge
\item[3] Human reference
\end{tablenotes}
\end{threeparttable}
\caption{\label{table:fullXSTSresultssst} Full calibrated XSTS \sst results; bootstrapped 95\% CI widths are $\sim\pm 0.1$ on average.  \%3+ and \%4+ refer to the percent of a language’s evaluated sentences with median (uncalibrated) XSTS scores equal to or greater than 3 and 4 respectively.}
\end{table}

Figure \ref{fig:XSTS_scores_cal_S2ST} shows aggregate, language-level XSTS scores for the \sst task, which are also presented in Table \ref{table:fullXSTSresultssst} and summarized in Table \ref{tab:XSTShighlevelsummarysst}. In the \xeng direction, where a \whisperlarge+\yourtts baseline has been evaluated, we see that XSTS scores for \mfourtlg tend to lag behind the baseline, with the notable exceptions of Bengali, Telugu, and Swahili (where \mfourtlg scores significantly higher than the \whisperlarge+\yourtts baseline). In aggregate, \mfourtlg still performs reasonably well, with the majority of language directions scoring at or above an XSTS score of 3.

XSTS \sst performance of \mfourtlg is in sharp contrast to two results: one is \st performance, where \mfourtlg is roughly on par with \whisperlarge and far surpasses performance in a few tested languages (most notably Swahili) in both XSTS and automated metrics (\blaser and \bleu), the other is \sst performance in terms automated metrics (\blaser and \asrbleu), where \mfourtlg consistently outperforms \whisperlarge+\yourtts baseline in the \xeng direction.

When moving from \st to the \sst task, language-level XSTS scores decreased by an average of 0.29 points for the \mfourtlg model but remained virtually unchanged (a decrease of only 0.01 points) for the \whisperlarge+\yourtts baseline.

Additionally, XSTS annotators report ``audio issues'' for \mfourtlg generations much more frequently, by almost an order of magnitude, than \whisperlarge+\yourtts generations (on \xeng direction, 206 \mfourtlg \sst generations were reported as having ``audio issues'' by at least one annotator, compared to only 22 \whisperlarge+\yourtts \sst generations and 13 human reference items; in \texttt{fra} (\xeng) for example, more than 10\% of \mfourtlg items were marked as having audio issues by at least one annotator). Excluding items marked as having audio issues by at least one annotator does not increase \mfourtlg XSTS scores appreciably (only 3 languages \xeng have XSTS score increases of slightly more than 0.1 when audio issue items are removed, and the rest have XSTS increases of less than 0.04), but high incidence of reported ``audio issues'' may be a symptom of a  problem that does correspond to the observed performance decrease of \mfourtlg on the \sst task in XSTS.

\paragraph{Truncation and omission in \sst output}

\begin{figure}
\centering
\includegraphics[width=0.9\textwidth]{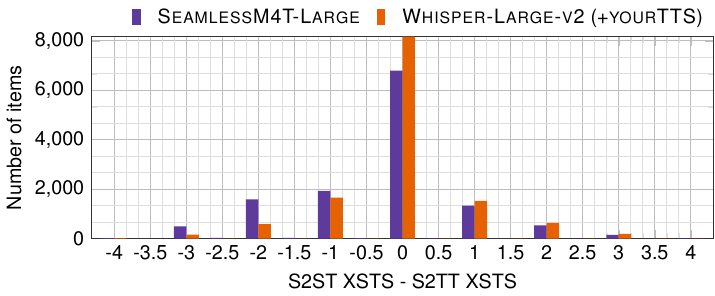}
\caption{\label{fig:xsts_delta_histogram} XSTS score differences between \sst and \st generations at the item level. A higher proportion of \mfourtlg \sst generations received XSTS scores that are 2 or more points behind the \st generation compared to \whisperlarge(+\yourtts).}

\end{figure}

Figure \ref{fig:xsts_delta_histogram} shows a histogram of performance deltas between \sst and \st generations for both \mfourtlg and \whisperlarge(+\yourtts). We see that \mfourtlg has a higher proportion of items with \sst generations 2 or more XSTS points behind the \st generation.

These observations motivated a small-scale inspection of 50 randomly sampled items (25 generations from \mfourtlg and 25 generations from \whisperlarge+\yourtts) where the median XSTS score was 5 for the \st generation but less than 4 for the \sst generation. We discovered that half of the sampled items in the \whisperlarge+\yourtts generations had poor \st translation quality but still received high marks. In the \mfourtlg sample, we similarly found that about half of the sampled items had a poor \st translation quality but still received a high score for the \st translation. Note that such a high proportion of XSTS scores not reflecting underlying performance in this inspection is not particularly surprising; we are conditioning on a score mismatch between modalities, and one common reason for a score mismatch is an errant score. However, in addition to this, half of the samples had the end of the \st translation truncated in the generated \sst audio (which did not occur in the sample of \whisperlarge+\yourtts generations). Truncations may have had a small impact on automated measures like \asrbleu but a large impact on the underlying meaning, and this may be reflected in the XSTS scores. The impact of a truncation or omission likely varies depending on which words are truncated or omitted; the distribution of performance deltas shown in Figure \ref{fig:xsts_delta_histogram} seem consistent with a mix of both minor and major impacts to underlying meaning.

If truncations are the predominant reason behind the drop in XSTS scores for \mfourtlg in the \sst task, one possible aggravating factor may be recency bias: due to truncations being the most recent effect an annotator observes, their perceived importance on translation quality may be higher. This may be especially true in the \sst task, where simultaneous comparison of source and target is difficult due to both being in the speech modality; if cognitive load is too high this may reduce an annotator's ability to holistically reflect on the quality of the entire translation and may further exacerbate the effect of recency bias.

\paragraph{\st and \sst performance as a function of source audio duration}
\begin{figure}
\centering
\includegraphics[width=0.8\textwidth]{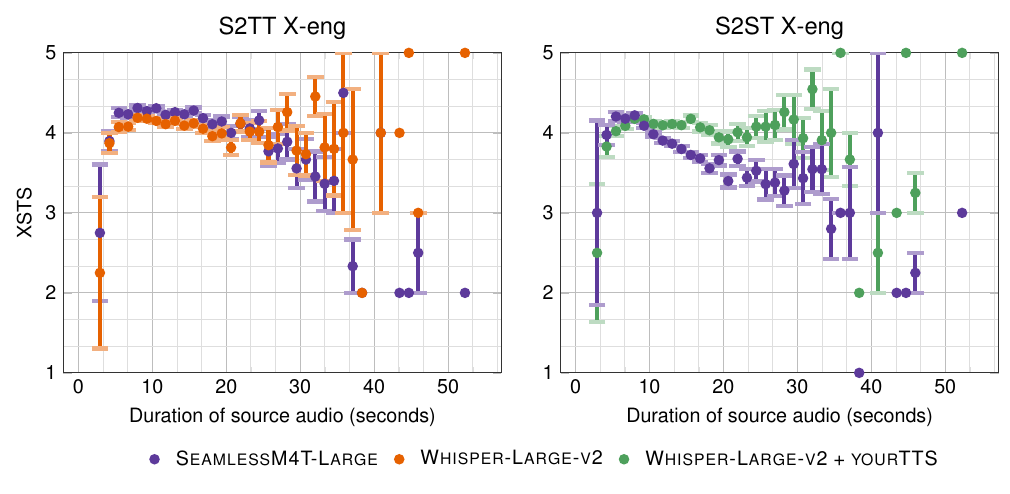}
\caption{\label{fig:perf_vs_src_duration} \st and \sst item-level XSTS scores as a function of the duration of the source speech audio file. Error-bars are standard errors around the mean in each source duration bucket.}
\end{figure}

Figure \ref{fig:perf_vs_src_duration} shows a breakdown of XSTS score by duration of source audio for both \st and \sst tasks for \mfourtlg and \whisperlarge(+\yourtts). We see that for the \st task, \mfourtlg provides superior performance over \whisperlarge on average item-level XSTS for most source audio durations, though performance for both models tends to decline for as a function of source audio duration for audios longer than $\sim$10 seconds. On the \sst task, \mfourtlg provides superior performance over \whisperlarge+\yourtts for shorter audio durations ($\lesssim$ 8 s) but \mfourtlg performance falls off rapidly for longer source audio durations compared to \whisperlarge+\yourtts performance which remains relatively stable for a broader range of source audio durations. This is consistent with the hypothesis that truncations and omissions are primarily responsible for the XSTS performance gap between \mfourtlg and \whisperlarge+\yourtts despite clear superiority of \mfourtlg on automated metrics, as longer source audios may more likely to suffer from truncations and omissions in the target audio.

\paragraph{MOS Results}

\begin{figure}
\centering
\includegraphics[width=0.8\textwidth]{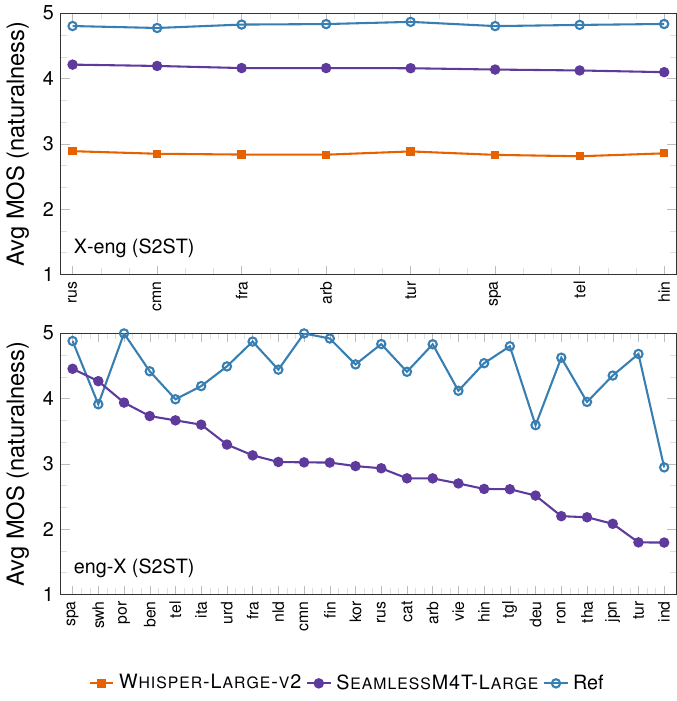}
\caption{\label{fig:MOS_naturalness} Language Direction level mean MOS (naturalness) scores per direction for \sst modality, after calibration. Bootstrapped 95\% CI widths are $\sim\pm 0.1$ on average.}
\end{figure}

Figure \ref{fig:MOS_naturalness} presents language-level results for both \engx and \xeng translation directions on the ``naturalness'' aspect of the MOS protocol. 

In the \xeng direction, \mfourtlg shows superior performance in the naturalness MOS aspect compared to \whisperlarge+\yourtts. The consistency of MOS score across input languages in the \xeng direction is likely due to two major factors. One, the input speech is unlikely to affect output speech quality with respect to any of the aspects measured by MOS. Two, since MOS requires only a review of the target audio, the items from each input language were evaluated by the same set of 31 annotators. With each annotator scoring a similar fraction of items for each input language, the annotator bias effects are similar across source languages.

In the \engx direction, \texttt{spa} and \texttt{swh} generations received MOS naturalness scores above 4, and 11 directions received naturalness scores above 3. Additionally, we also measured ``Clarity of Speech'' and ``Sound Quality'' MOS aspects. We find that for languages where both \whisperlarge+\yourtts and \mfourtlg were measured (all \xeng), \mfourtlg averaged 0.67 points higher in sound quality and 0.79 points higher in clarity of speech over \whisperlarge+\yourtts (\mfourtlg surpassed the human reference in the ``clarity of speech'' and ``sound quality'' aspects for all 8 \xeng languages measured). In the \engx direction, for the ``clarity of speech'' aspect, 20 languages scored above a 3 for \mfourtlg, with 4 languages scoring above a 4; for the ``sound quality'' aspect, all languages scored above a 3, with 13 languages scoring above a 4.

\paragraph{Correlations between \asrbleu, \bleu and XSTS}
\begin{table}
\centering
\begin{tabular}{lll|cc}
\toprule
\textbf{Modality} & \textbf{Direction} & \textbf{Metric} & \textbf{Pearson} & \textbf{Spearman} \\
\toprule
\st & \engx & \blaser & \textbf{0.750} & \textbf{0.505} \\
 &  & \bleu & 0.053 & 0.355 \\

& \xeng & \blaser & \textbf{0.923} & \textbf{0.871} \\
&  & \bleu & 0.822 & 0.776 \\
& All & \blaser & \textbf{0.913} & \textbf{0.827} \\
&  & \bleu & 0.626 & 0.625 \\
\midrule
\sst & \engx & \blaser & \textbf{0.727} & \textbf{0.675} \\
 &  & \asrbleu & 0.154 & 0.292 \\

& \xeng & \blaser & \textbf{0.854} & \textbf{0.756} \\
&  & \asrbleu & 0.692 & 0.651 \\

& All & \blaser & \textbf{0.810} & \textbf{0.736} \\
&  & \asrbleu & 0.514 & 0.561 \\
\bottomrule
\end{tabular}
\caption{\label{tab:automated_metric_xsts_correlations} Comparison of language-level correlations between automated metrics and XSTS. \blaser provides superior correlations with human metrics in all directions, and modalities, with a particular improvement in correlations for \engx directions. These results hold for both Pearson and Spearman correlation coefficients.}
\end{table}

\begin{figure}
\centering
\includegraphics[width=0.8\textwidth]{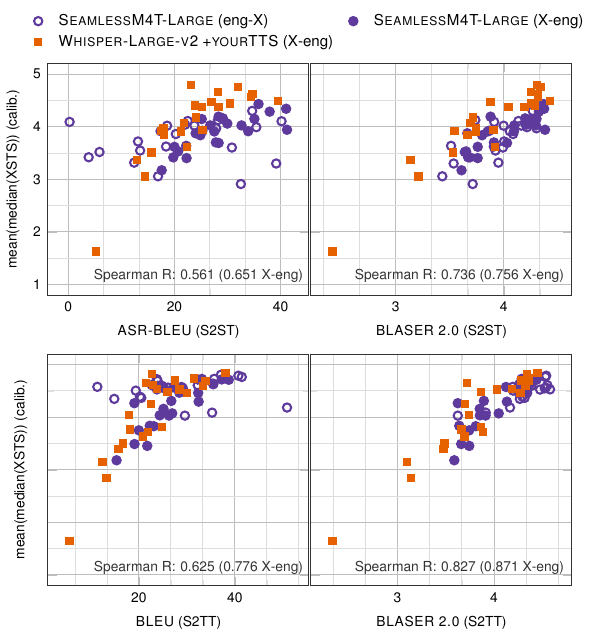}
\caption{\label{fig:auto_metric_corrs} Correlations between XSTS and \asrbleu and \blaser for \sst task and between XSTS and \bleu and \blaser for \st task. \blaser offers superior correlation with XSTS, most notably for \engx directions where \asrbleu and \bleu correlations are much weaker than for \xeng directions.}
\end{figure}

In Table \ref{tab:automated_metric_xsts_correlations} and visualized in Figure \ref{fig:auto_metric_corrs}, we compare the strength of correlations between automated metrics (\asrbleu for \sst and \bleu for \st) and XSTS at the language level. \asrbleu and \bleu are computed using corpus scores, and language level \blaser scores are computed by averaging sentence-level (supervised) \blaser scores over the evaluation set. All automated metrics are calculated using only the items sent for human evaluation with XSTS in order to make consistent comparisons, however \asrbleu and \blaser scores on the language level differed only slightly from those used to generate Tables \ref{tbl:modeling:cascaded:s2st} and \ref{tbl:modeling:cascaded:x2t}.

We find that, both with Spearman and Pearson correlation coefficients, \blaser has a superior correlation with XSTS over \asrbleu (for \sst task) and \bleu (for \st task), and \blaser correlates much better with XSTS in the \engx direction in particular.

\subsubsection{Limitations}

\textbf{Test set limitations} The \fleurs \citep{fleurs2022arxiv} test set is limited in that the different language pairs contain slightly different sets of sentences. Due to the limitations in both the dataset (which contains a maximum of 3 speakers) and timing and cost considerations on the human evaluation front (we evaluated a maximum of two speakers per sentence), we have a lack of diversity in our speaker set per language, which may introduce bias relative to a test set with a larger number of speakers.

\noindent \textbf{Limited sample size of human annotators per language} We only have a maximum of 5 (but typically 3) annotator evaluations per sentence for each language in our XSTS evaluations. Relatively small samples of annotators mean annotator bias is important to consider. We try to mitigate this by (1) using the median score per sentence for each language to be robust to outliers, (2) using bootstrap re-sampling of annotator scores to estimate language score uncertainty due to finite annotators, and (3) approximate and correct annotator bias with a cross-lingual calibration set. For MOS, several language directions only have a single annotator per item, with no calibration set—this exacerbates issues related to annotator bias.

\noindent \textbf{Challenges of implementing reliable human measures of performance} Obtaining human measures of translation performance offers several advantages over automated metrics (e.g. while \bleu-like metrics are trivial to ``game'', humans are not as easily fooled, and humans are better judges of target translations of high quality but for which word content differs from available reference translations), but human measures of translation quality suffer from their own limitations that are not shared by automated metrics. XSTS was designed to mitigate some of these limitations (annotator bias and variance) but, e.g., recency bias and other human biases may still persist. We continue to develop additional recommendations and clarifications, improve annotator trainings and reduce cognitive load, but obtaining reliable human measures of translation quality continues to be an active area of research and development.

\subsection{Automatic Robustness Evaluation}
\label{subsec:appendix:robustness}

We evaluated model robustness against non-linguistic perturbations in the real-world speech inputs, including background noises and speaker variations. As reported in several other sections, we compare our model to \whisperlarge.%

\subsubsection{Robustness Against Background Noises}

\begin{figure}[t]
    \centering
    \includegraphics[width=1.1\linewidth]{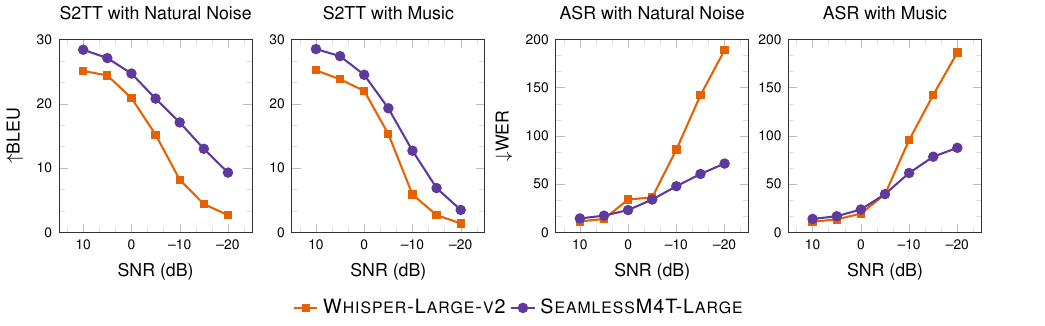}
    \caption{\textbf{Evaluation results of model robustness against background noises.} We report average test \bleu and test \wer over 4 languages (3 language families) for \xeng~\st and ASR on \fleurs with low-to-high input noise level (high-to-low SNR). Simulated noises are sampled from MUSAN~\citep{snyder2015musan} on the ``noise'' and ``music'' categories.}
    \label{fig:robustness_noise}
\end{figure}

\paragraph{Related work} The analysis of speech model robustness across different background noise levels has been conducted in prior work~\citep{wang2022wav2vec,zhu2022noise,whisper} on simulated noisy audios. However, existing simulation-based evaluations are either limited by the noise types (e.g., simple white noise), task coverage (e.g., ASR only), language coverage (e.g., English only), or the replicability of benchmark data. This calls for an open, versatile benchmark to overcome these limitations.

\paragraph{Experimental framework} We build a replicable noise-robustness evaluation benchmark based on \fleurs ("noisy \fleurs''), which covers 102 languages, 2 speech tasks (\st and ASR), and various noise types (natural noises and music). To create simulated noisy audios, we sampled audio clips from MUSAN~\citep{snyder2015musan} on the ``noise'' and ``music'' categories, and mixed them with the original \fleurs speech audios under different signal-to-noise ratio (SNR): 10, 5, 0, -5, -10, -15 and -20. We compare models by \bleu-SNR curves (for \st) or \wer-SNR curves (for ASR), which illustrate the degree of model performance degradation when the noise level of speech inputs increases (i.e., when SNR decreases). Both \mfourtlg and \whisperlarge achieve high performance mostly in high-resource languages, where stress tests in the noisy speech setup are more necessary and informative. For low-resource languages, the clean speech setup is already challenging, let alone the noisy one. We hence focus on 4 high-resource languages (French, Spanish, Modern Standard Arabic, and Russian) from 3 different language families for our noise-robustness analysis on \mfourtlg and \whisperlarge.

\paragraph{Results} Figure~\ref{fig:robustness_noise} shows the average test \bleu and test \wer over the 4 languages for \xeng{} \st and \asr on \fleurs with low-to-high input noise level (high-to-low SNR). We see that both \bleu-SNR curves for \mfourtlg are consistently above those for \whisperlarge. Similarly, \mfourtlg's \wer-SNR curves are consistently below \whisperlarge's ones. These suggest the superior robustness of \mfourtlg in noisy speaking environments. \mfourtlg outperforms \whisperlarge by an average of 33.3\% and 42.2\% over various noise types and noise levels for \xeng{} \st and \asr, respectively.

\subsubsection{Robustness Against Speaker Variations}
\label{subsec:robustnessspeakervariation}

\paragraph{Related work} \asr and \st systems are expected to minimize the effects of speaker variations that are irrelevant to the input content of interest. Fairness of \asr systems to different speaker subgroups (by race, gender, country, etc.) has been studied in prior work~\citep{liu2022towards,dheram2022toward}, which requires the availability of accurate speaker demographics labels~\citep{hazirbas2021casual,porgali2023casual} for speaker grouping and group-wise scoring. However, these labels are rare in existing \asr benchmarks, limiting the applications of such analysis. To overcome label scarcity, \cite{wang-etal-2020-covost} proposed a set of label-free metrics that do not rely on speaker grouping for analyzing the effects of speaker variations.

\paragraph{Experimental setup} We follow~\cite{wang-etal-2020-covost} to evaluate model robustness against speaker variations by calculating average by-group mean score and by-group coefficient of variation of an utterance-level quality metric. Instead of using \bleu as the quality metric, we used chrF, which has better stability at the utterance level. The calculation of both robustness metrics does not require explicit speaker subgroup labels. We grouped evaluation samples and corresponding utterance-level chrF scores by content (transcript), and then calculated the average by-group mean score $\textrm{chrF}_{MS}$ and average by-group coefficient of variation $\textrm{CoefVar}_{MS}$ defined as follows:
\begin{align*}
    \textrm{chrF}_{MS}&=\frac{1}{|G|}\sum_{g\in G}\textrm{Mean}(g) \\
    \textrm{CoefVar}_{MS}&=\frac{1}{|G'|}\sum_{g\in G'}\frac{\textrm{StandardDeviation}(g)}{\textrm{Mean}(g)}
\end{align*}
where $G$ is the set of sentence-level chrF scores grouped by content (transcript) and
$G' = \{g | g\in G, |g|>1, \textrm{Mean}(g) > 0 \}$.
The two metrics are complementary: $\textrm{chrF}_{MS}$ provides a normalized quality metric that, unlike conventional corpus-level metrics, takes speaker variations into consideration, while $\textrm{CoefVar}_{MS}$ provides a standardized measure of quality variance under speaker variations. 
For robustness analysis of \mfourtlg and \whisperlarge, we conducted an out-of-domain evaluation on \fleurs on all its languages that have at least 40 content groups in the test sets.

\begin{table}[htb]
    \centering
    \small
    \begin{tabular}{lccccc}
        \toprule
        \multirow{1}{*}{\bf Languages} & \multirow{1}{*}{\bf Average \#} &  \multicolumn{2}{c}{\bf \whisperlarge} & \multicolumn{2}{c}{\bf \mfourtlg} \\
        \cmidrule(lr){3-4} \cmidrule(lr){5-6}
        ($\ge$ 40 content groups) & cont. groups & chrF$_{MS}$$\uparrow$ & CoefVar$_{MS}$$\downarrow$ & chrF$_{MS}$$\uparrow$ & CoefVar$_{MS}$$\downarrow$ \\
        \midrule[\heavyrulewidth]
        \xeng~\st for 77 langs & 278 & 40.8 & 13.7 & \textbf{45.3} & \textbf{9.1} \\
        \asr for 78 langs & 280 & 58.7 & 17.0 & \textbf{72.5} & \textbf{6.4} \\
        \bottomrule
    \end{tabular}
\caption{\textbf{Evaluation results of model robustness against speaker variations.} We report average by-group mean chrF (chrF$_{MS}$) and average by-group coefficient of variation on chrF (CoefVar$_{MS}$) on \fleurs \xeng \st and ASR test sets.}
\label{tbl:robustness-speaker-variation}
\end{table}

\paragraph{Results} Table~\ref{tbl:robustness-speaker-variation} shows the $\textrm{chrF}_{MS}$ and $\textrm{CoefVar}_{MS}$ scores of \mfourtlg and \whisperlarge on \fleurs~\xeng~\st and \asr test sets. We see that \mfourtlg outperforms \whisperlarge on $\textrm{CoefVar}_{MS}$ by an average of 49.4\% over the 2 tasks. Moreover, \mfourtlg outperforms \whisperlarge on chrF$_{MS}$ by an average of 18.3\%. These suggest the superior robustness of \mfourtlg when it comes to speaker variations.

\section{Responsible AI}
\label{sec:appendix:ra}

In line with our expectations to build systems responsibly, we focus our efforts on the evaluation of added toxicity and bias. Both of these dimensions of responsible AI have drawn significant scientific attention in recent times (e.g., \citep{kiritchenko-et-al-2021,bender-et-al-2021,costajussa:nature:2019}). Moreover, the occurrence of these unintended errors or translation faults could adversely impact user experiences. Sustained attention devoted to such issues is, thus, vital to the safe deployment of our systems. 

Beyond these dimensions, we are also concerned with the concept of fairness. In contrast to the idea of robustness (as conceptualized in section \ref{subsec:robustnessspeakervariation}), where the focus is on whether our system performance is affected by the varying qualities of a speaker's voice, fairness in this section is more concerned about the \textit{content} of the translation outputs. Fair outputs do not preference or skew towards particular demographics and tend to treat different groups somewhat equitably. We document the results of these evaluations to better direct mitigation efforts.

\subsection{Definitions}
\label{sec:rai-def-back}
We begin by detailing how we define errors that arise from \textit{added toxicity} and \textit{gender bias}.

\paragraph{Toxicity.}  
In their taxonomy of critical machine translation errors, \citep{sharou-specia-2022-taxonomy} define ``deviation in toxicity'' as ``instances where the translation may incite hate, violence, profanity, or abuse against an individual or a group (a religion, race, gender, etc.) due to incorrect translations,'' which ``covers cases where toxicity is introduced into the translation when it is not in the source, deleted in the translation when it is in the source, mistranslated into different (toxic or not) words, or not translated at all (i.e., the toxicity remains in the source language or transliterated).'' Our definition of \textit{added toxicity} departs slightly from theirs in that it does not cover instances of untranslated toxic source content or of toxic source content deleted in the translation. To put it simplistically, added toxicity is the introduction of toxic elements not present in a source utterance.

\paragraph{Gender Bias.} Another error with which responsible AI is concerned lies in the propagation and amplification of gender bias. In machine translation, gender bias is observed when translations show errors in linguistic gender determination despite the fact that there are sufficient gender clues in the source content for a system to infer the correct gendered forms. To illustrate this phenomenon, sentence (1) below does not contain enough linguistic clues for a translation system to decide which gendered form should be used when translating into a language where the word for \textit{doctor} is gendered. Sentence (2), however, includes a gendered pronoun which most likely has the word \textit{doctor} as its antecedent.
\begin{enumerate}
\item I didn't feel well, so I made an appointment with my doctor.
\item My doctor is very attentive to \textbf{her} patients' needs.
\end{enumerate}
Gender bias is observed when the system produces the wrong gendered form when translating sentence (2) into a language that uses distinct gendered forms for the word \textit{doctor}. A single error in the translation of an utterance the like of sentence (1) would not be sufficient to conclude that gender bias exists in the model; doing so would take consistently observing one linguistic gender over another.
It has previously been hypothesized that one possible source of gender bias is gender representation imbalance in large training and evaluation data
sets, e.g. \citep{NEURIPS2022_09933f07,qian-etal-2022-perturbation}.

\subsection{Toxicity}
\label{apx:toxicity}

Warning: this section contains examples that may be offensive to some.

\subsubsection{Motivation}

\paragraph{Context} As mentioned above, added toxicity means introducing toxicity in the translation output not present in the input. This can be classified as a critical error; one that could lead users to distrust a translation system. As such, it is important to quantify how much toxicity our models add. We are also interested in combining added toxicity analysis with demographic bias analysis to determine whether added toxicity is generated more in certain demographic axes than in others.

\paragraph{Related work} While related research in speech toxicity detection is quite limited \citep{10.1007/978-3-030-60276-5_19,DBLP:conf/eusipco/YousefiE21}, toxicity detection for text-based approaches has been widely explored in different contexts. Many examples of these efforts can be found in large evaluations like JigSaw Series Kaggle Competitions\footnote{https://www.kaggle.com/c/jigsaw-toxic-comment-classification-challenge} or WMT Critical Error detection \citep{specia-etal-2021-findings}. Recently, in the context of \mt, there has been a substantial push to scale toxicity detection by using a word-list-based detection method for models such as NLLB \citep{nllb2022}, which further spurred research into analyzing toxicity at scale \citep{costajussa2023toxicity} and mitigation strategies \citep{gilabert2023resetox}.  %
Using a dataset that covers different demographic axes can allow for further analysis of which demographic axes are most sensitive to toxicity \citep{costajussa2023toxicity}. So far, datasets that cover a wide range of demographic axes mostly focus on text and more attention needs to be directed at speech (an example of a text data is \holisticbias \citep{smith-etal-2022-im}).

\paragraph{Proposed methodology} 

Inspired by \asrbleu, this work proposes using \asretox as a new metric to detect added toxicity in speech and evaluate added toxicity for \mfourt's \sst capability. Essentially, this metric follows a cascaded framework by first deploying a standard ASR module (i.e., the same that it is used for \asrbleu as defined in Table \ref{tbl:eval:metrics}), then the toxicity detection module, ETOX \citep{costajussa2023toxicity}, which uses the Toxicity-200 word lists. %
For \st, the translated output can be directly evaluated with ETOX. In both cases (\sst and \st), we measure added toxicity at the utterance/sentence level. We first compute toxicity detection for each input in the evaluation dataset and the corresponding output. Then we compare them and count a case as containing added toxicity only when the output value exceeds the one displayed by the input. %

\subsubsection{Experimental Framework}
\label{sec:expframetoxicity}

\paragraph{Language directions and modalities} Similarly to the previous human evaluation framework in Section \ref{subsec:appendix:humanmetrics}, we evaluated \sst and \st on \fleurs. Distinctive from human evaluation, we extended toxicity evaluation to cover all languages for which we provide translations for as summarized in Table \ref{tab:all_languages}. Igbo, Burmese, Nepali, and Assamese have issues related to segmentation and consistencies in the toxicity word lists.%
With these problems, these languages tend to over-detect toxicity and we consider them to be outliers. Therefore, we excluded them from the analysis and results.

\paragraph{Datasets} We used two datasets to analyze added toxicity. One, we used \fleurs to better align with our human evaluation effort and other evaluative components of this work. %
In addition, %
we used the English-only \holisticbias framework \citep{smith-etal-2022-im}, which has been shown to trigger true added toxicity in previous studies \citep{costajussa2023toxicity}. %
\holisticbias comprises 26 templates, encompassing more than 600 descriptors across 13 demographic axes, along with 30 nouns. The dataset consists of over 472K English sentences utilized in the context of two-person conversations. Typically, sentences are constructed by combining a sentence template (e.g., "I am a [NOUN PHRASE]."), a noun (e.g., parent), and a descriptor (e.g., disabled). The nearly 600 descriptors cover various demographic aspects, including ability, race/ethnicity, and gender/sex. The nouns may indicate a specific gender (e.g., woman, man) or avoid gender references (e.g., child, kid). Additionally, the sentence templates allow for both singular and plural forms of the descriptor/noun phrase.

In this work, we extend \holisticbias to speech by applying the default “en” transformer-tts model from fairseq S{\^{}}2
(\citep{wang-etal-2021-fairseq}). %
It first converts input texts into IPA phonemes, then passes them to a mel spectrogram generator transformer model
, and finally feeds the outputs to a HiFi-Gan vocoder to create the waveform. %

\paragraph{Models} As a baseline system for \st\ \xeng, we employ \whisperlarge~\citep{whisper}. As for \sst~\xeng, we apply the \cite{casanova2022yourtts} to generate synthesized speech from the output of \whisperlarge\ \st. For \st\ \engx, we employ the cascade system of \whisperlarge + \nllbmedium~\citep{nllb2022}.
Below, we report results for \mfourtlg.%

\paragraph{Evaluation} We use the Github implementation of ETOX\footnote{https://github.com/facebookresearch/stopes/tree/main/demo/toxicity-alti-hb/ETOX} For languages without spaces, we use the spm tokenization option in the tool.
For \asr, we use the same implementation framework used for \asrbleu as reported in Table \ref{tbl:eval:metrics}. %

\subsubsection{Results}

\begin{figure}[h!]
    \centering
\noindent\includegraphics[width=.45\linewidth]{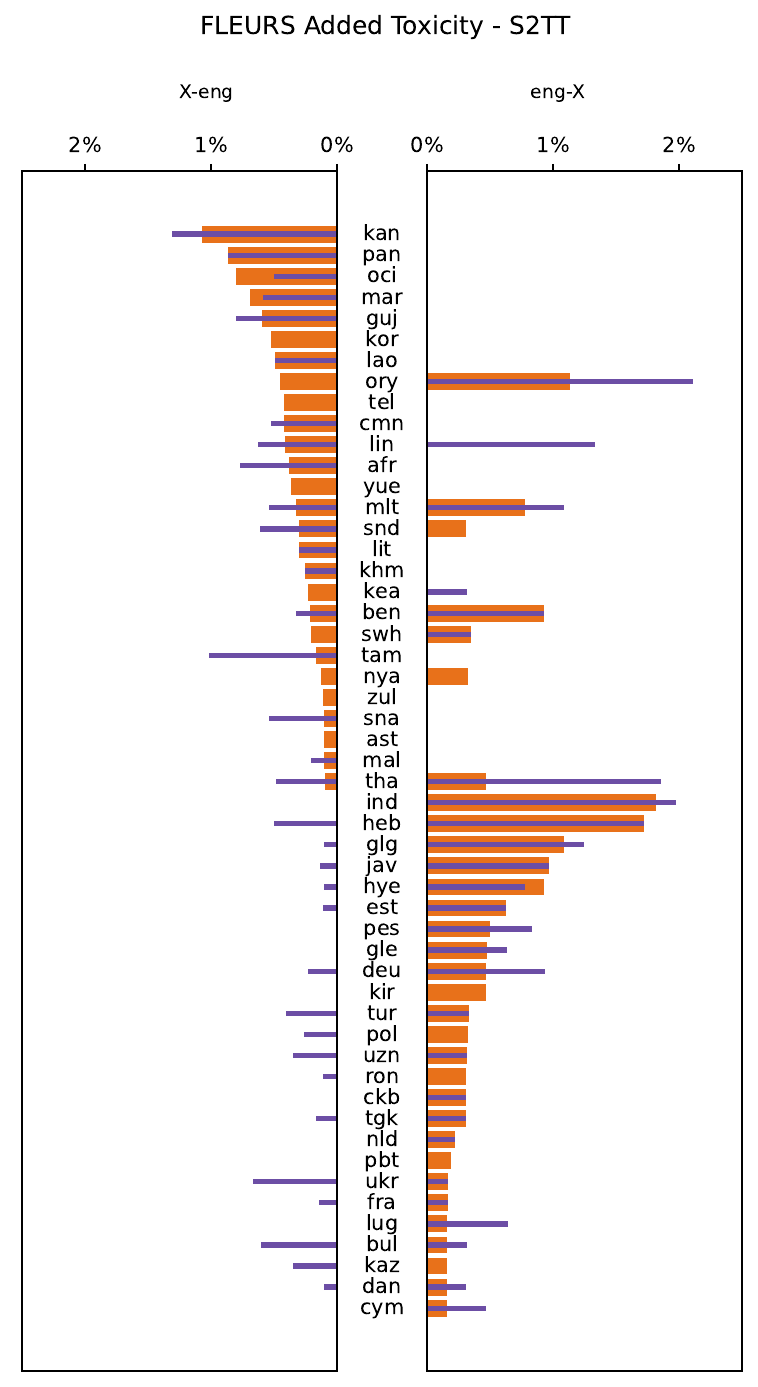}
\noindent\includegraphics[width=.45\linewidth]{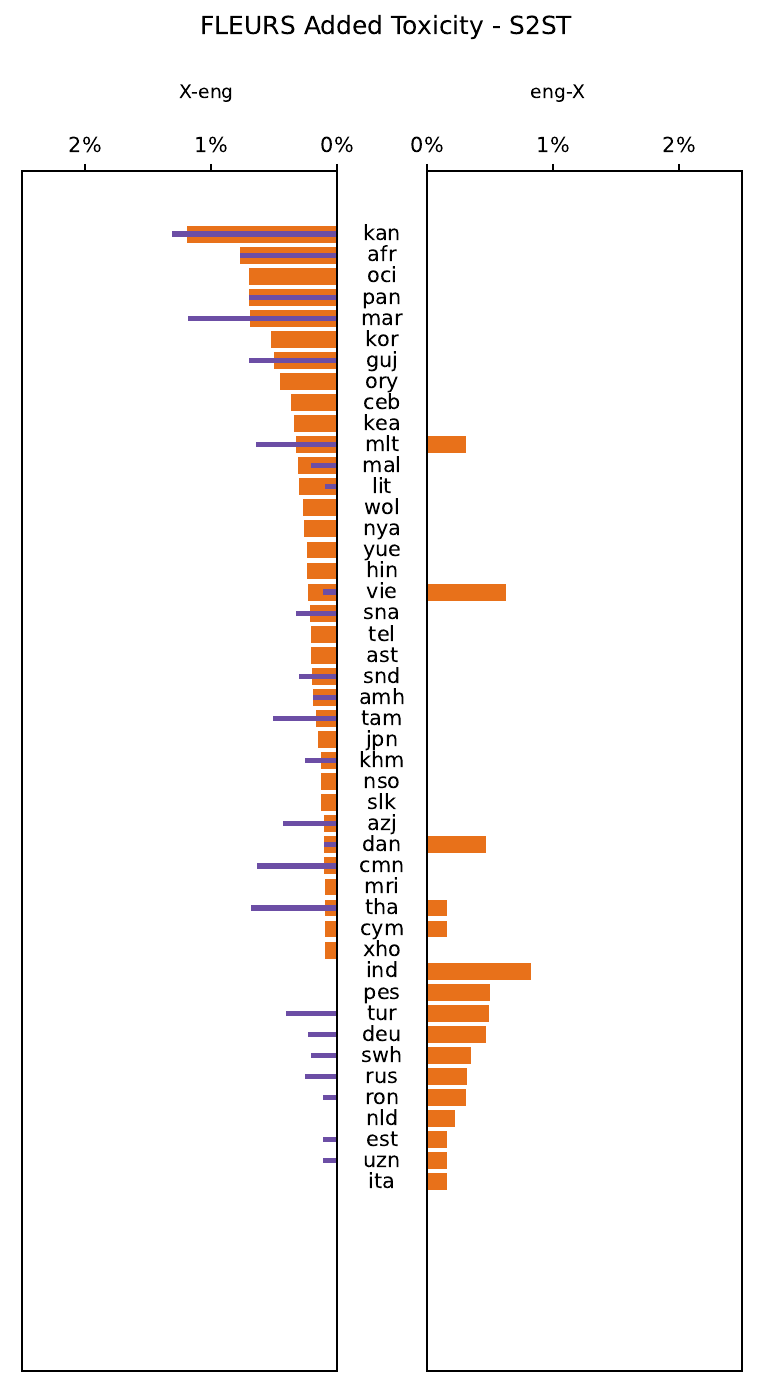}
    \caption{\label{xxtoen-toxicity-s2t} Added toxicity for \xeng and \engx for \st (left) and \sst (right) in \fleurs. The figure shows the number of outputs with added toxicity per language both for \mfourtlg (orange) and \whisperlarge and \whisperlarge + \yourtts systems when available (blue).}
\end{figure}

\paragraph{Automatic toxicity detection on \fleurs} We evaluated the output of \mfourtlg on the \fleurs dataset. 
\Cref{xxtoen-toxicity-s2t} presents results from \st and \sst for \xeng and \engx directions, where we show the number of sentences that contain added toxicity. When looking at the amount of added toxicity per sentence, less than 5\% of the cases contain more than 1 added toxicity token per sentence. Overall, \fleurs shows a relatively low prevalence of added toxicity of 0.15\%, averaging across languages, tasks, and translation directions. %

For \st  in \xeng  (Figure \ref{xxtoen-toxicity-s2t} (left)), added toxicity is 0.11\% averaged across languages, and 27 language pairs contain some added toxicity. For \sst (Figure \ref{xxtoen-toxicity-s2t} (right)), added toxicity is 0.12\% averaged across languages, and 35 language pairs contain added toxicity. 
  
For \st in \engx, (Figure \ref{xxtoen-toxicity-s2t} (left)), added toxicity is 0.21\% averaged across languages, and 32 language pairs contain added toxicity. For \sst (Figure \ref{xxtoen-toxicity-s2t} (right)), added toxicity is 0.16\% averaged across languages, and 16 language pairs contain added toxicity. The main difference across modalities is the reduced amount of added toxicity in \sst for the \engx translation direction. We comment on this difference alongside the results from the \holisticbias dataset later in this section.

By comparison, for \st in \xeng, \whisperlarge's added toxicity is 0.31\% averaged across languages and is prevalent in 58 languages. For overlapping languages in \whisperlarge and \mfourtlg, the latter shows an added toxicity reduction of 63\%. For \sst in \xeng , \whisperlarge + \yourtts's added toxicity lies at 0.27\% averaged across languages and is prevalent in 52 languages.%
Again, for overlapped languages in this cascaded \sst system and \mfourtlg, %
ours show a reduction of toxic tokens by 62\%. For \st in \engx, the \whisperlarge + \nllbmedium cascaded combination adds toxicity by 31\% averaged in languages and added toxicity is prevalent in 39 languages. For overlapping languages, \mfourtlg reduces this amount by 26\%. The filtering of imbalanced toxicity in the training data as reported in Section \ref{sec:modeling:x2t:data} may have contributed to this improvement.

\begin{figure}[h!]
    \centering
\noindent\includegraphics[width=.28\linewidth]{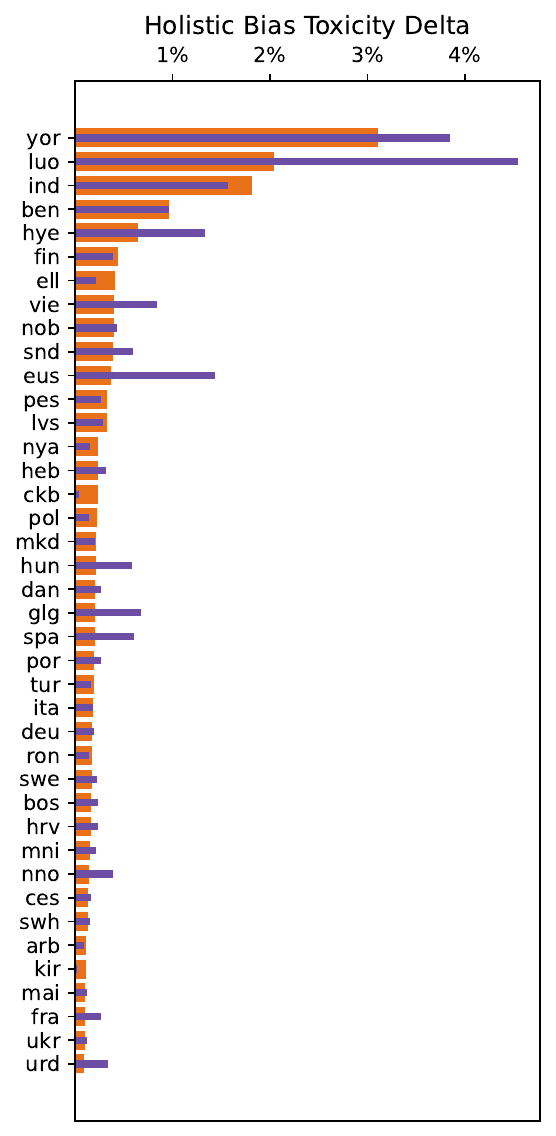}
\noindent\includegraphics[width=.70\linewidth]{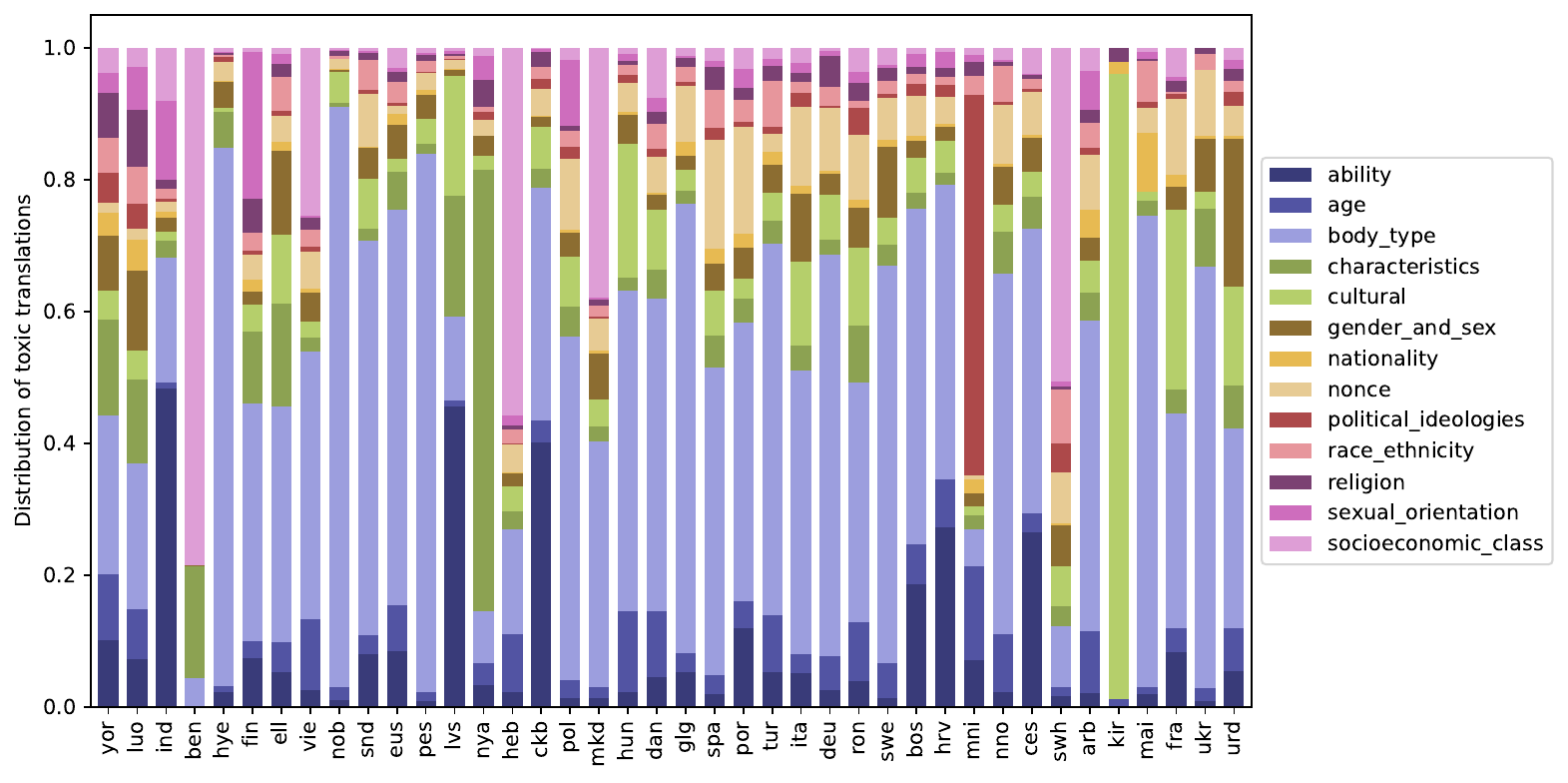}
    \caption{\label{entoxx-toxicity-hb-s2t} (left) Added toxicity for \engx, \st in \holisticbias. Showing top 40 languages. The plotted languages are above 500 samples of added toxicity—0.1\% of the dataset. (right) Different languages differ in distributions of toxic terms as a function of demographic axes, with some languages’ toxicity being dominated by only one or two axes.  The figure shows the number of outputs with added toxicity per language both for \mfourtlg (orange) and \whisperlarge + \nllbmedium cascaded systems (blue).
}
\end{figure}

\begin{figure}[h!]
    \centering
\noindent\includegraphics[width=.28\linewidth]{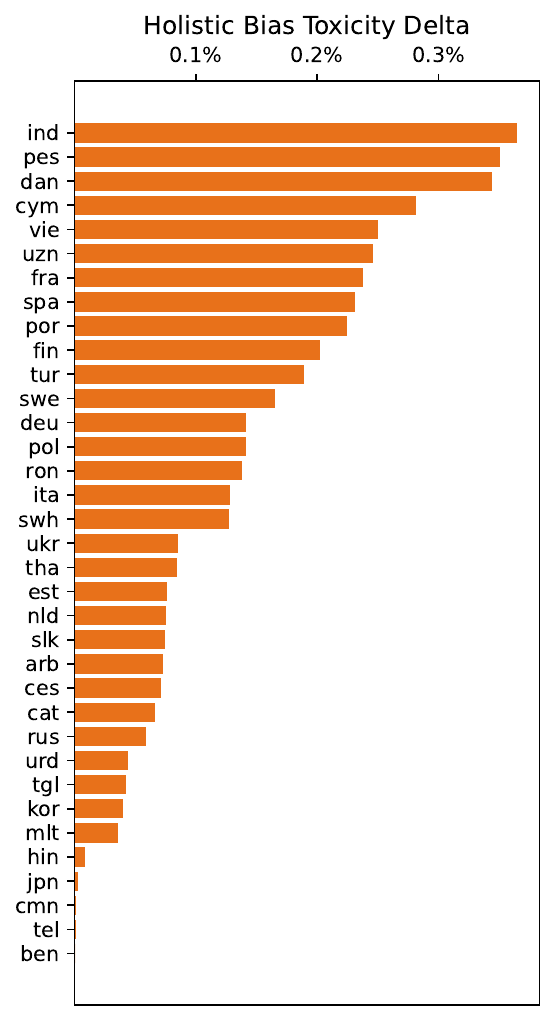}
\noindent\includegraphics[width=.70\linewidth]{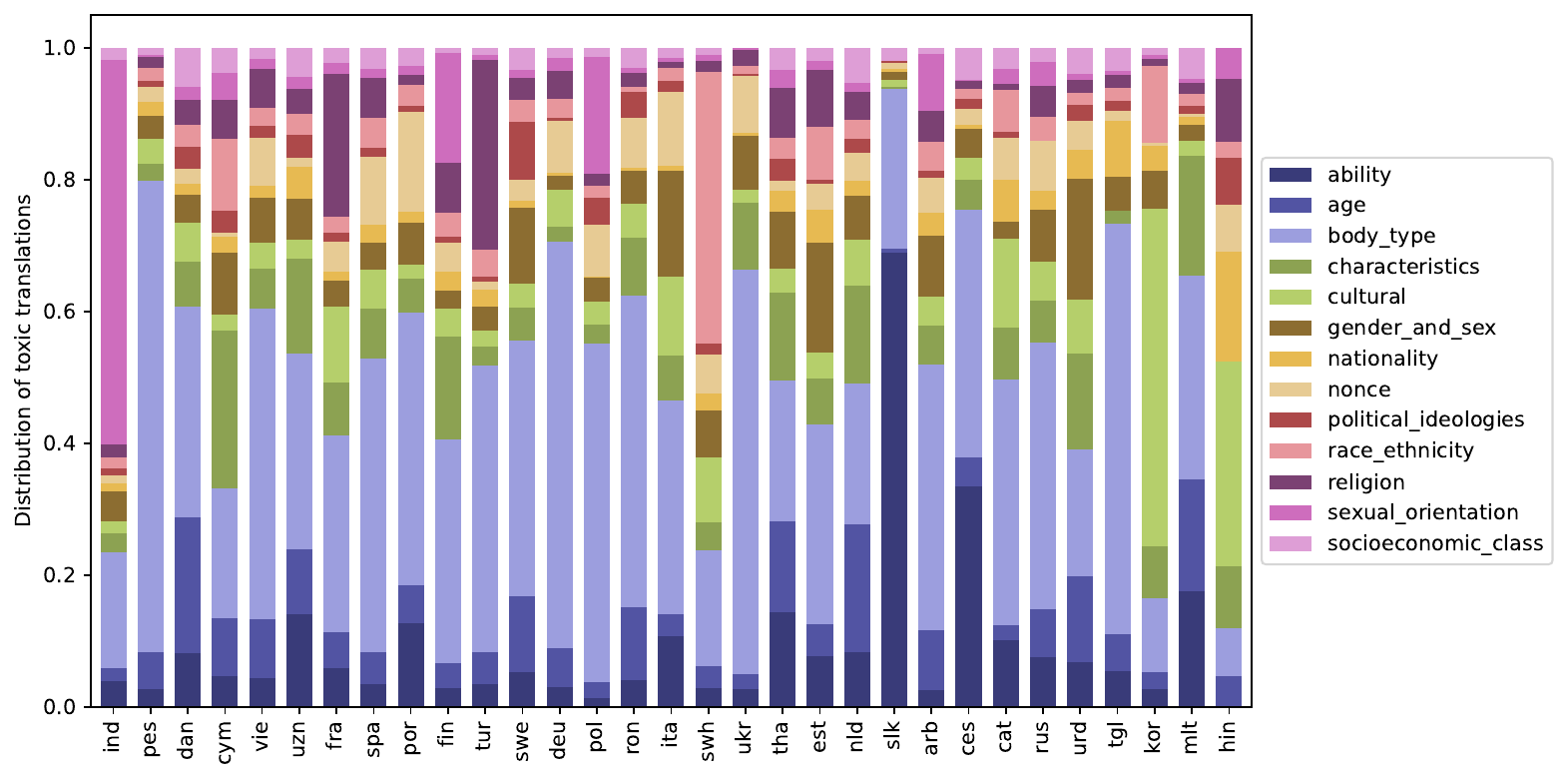}
    \caption{\label{entoxx-toxicity-hb-s2s} (left) Added toxicity for \engx, \sst in \holisticbias. Showing all target languages. (right) Similarly to \st, different languages differ in distributions of toxic terms as a function of the demographic axis, with some languages’ toxicity being dominated by only one or two axes. 
}
\end{figure}

\paragraph{Automatic Toxicity Detection on \holisticbias Dataset}  %
\Cref{entoxx-toxicity-hb-s2t} (left) shows results for \st languages with the highest added toxicity when translating \holisticbias from \engx (note that \holisticbias is only available in English).%
Here, we observe a slightly higher amount of added toxicity compared to \fleurs for \st and a slightly lower amount for \sst. Overall, \holisticbias shows a prevalence of added toxicity of 0.19\% for \st and 0.13\% for \sst, averaged across languages. 
For \st, there are 84 languages that are affected by added toxicity. When looking at added toxicity per sentence, less than 0.003 \% of the outputs contain more than one added toxicity token. 
Figure \ref{entoxx-toxicity-hb-s2s} (left) shows results for \sst languages when translating the \holisticbias dataset. %
In total, there are 34 languages with added toxicity. %

For \st, the \whisperlarge + \nllbmedium cascaded combination adds toxicity by 29\% averaged in languages and added toxicity is prevalent in 81 languages. \mfourtlg reduces the amount of added toxicity by 34\%.

Through manual inspection, when comparing toxic words detected in \st translation but not in \sst, we observed that the word occurrences are similar with minor differences. We hypothesize that using ASR before toxicity detection tends to cause false negatives, which would explain the high decrease in added toxicity from \st to \sst (from 0.19\% to 0.13\%), which also happened in \fleurs\ (from 0.21\% to 0.16\%). For example, in the case of English to Catalan, the word "merda" in the \sst output is usually written as "mereda", and therefore not identified by ETOX. This type of example brings light to the limitations presented by detection based on tokens in a blacklist. %

Following previous work \citep{costajussa2023toxicity}, we perform an analysis of toxicity per \holisticbias' axes and report them in Figures \ref{entoxx-toxicity-hb-s2t} and \ref{entoxx-toxicity-hb-s2s} (right). Figures show the distribution of toxic translations per category and how they vary per language. We see that different languages differ in their distributions of toxic terms as a function of demographic axes. For most languages, the toxicity distribution across an axis is proportional to the axis' overall share. For instance, the main category in terms of volume is 'body type', representing 25\% of the dataset. This same category tends to accumulate a larger amount of toxicity as well. However, for some languages the toxic sentences appear to be highly concentrated in a particular axis—such is the case for Bengali (80\% socio-economic status), Nyanja (66\% characteristics), and Kyrgyz (94\% cultural) to name a few.

The categories that have a higher concentration of toxicity for \st and \sst are nonce (0.79\% and 0.46\%) and sexual orientation (0.62\% and 0.35\%). Nonce category (nonsense) is a bit of an outlier as far as terms are concerned because they do not specifically refer to any demographic groups. In terms of categories for least added toxicity, those would be age for \st (0.37\%), and political ideologies for \sst. 

\paragraph{Human evaluation of added toxicity detection} 

As explained in Section \ref{sec:appendix:rai:rai_limitations}, word-list-based toxicity detection techniques are known to produce a substantial number of false positives.
These are typically due to one of two main factors: tokenization issues and toxicity list items that are only toxic in certain contexts.

We perform a human error analysis of all detected items, and determine whether these are true positives (TP) or false positives (FP) on the following translation outputs:
\begin{itemize}
    \item \fleurs outputs in the \st modality and both sets of translation directions (\engx and \xeng), and for both \mfourt and baseline systems (\mfourtlg, \whisperlarge, \whisperlarge + \nllbmedium, \whisperlarge + \yourtts), %
    \item \holisticbias outputs in the \st and \sst modalities and only the \engx set of directions for the \mfourtlg system.
\end{itemize}

\textit{Different results on \holisticbias and \fleurs} The main difference between \holisticbias and \fleurs, comparing only the comparable set of directions (i.e. \engx) is that the TP and FP rates follow opposite trends:
97–98\% of detected items are TP in \holisticbias, whereas 81–84\% of detected items are FP in \fleurs.

\textit{Comparison between the \sst and \st modalities on \holisticbias}. The S2TT modality produced more (1.2 x) detected added toxicity than S2ST.
The percentages of TP and FP are roughly the same in both modalities, and follow the same clear trend (between 97 and 98\% of detected items are TP).

\textit{Comparison between M4T and baseline on FLEURS in the S2TT modality}.
In the eng-x directions both systems show similar results: 
The baseline system produces more detected added toxicity (~1.2x) than M4T;
Between 80–84\% of detected items are FP for both systems.
In the x-eng directions more differences can be observed:
The baseline system produces 2x the number of detected added toxicity produced by M4T;
While in the baseline system the reversal in direction polarities is mirrored by a reversal in the TP-to-FP ratio (eng-x: 18.6\%TP, 81.4\% FP; x-eng: 80.3\% TP, 19.7\% FP), this does not occur in the M4T system (eng-x: 16.3\%TP, 83.7\% FP; x-eng: 42.3\% TP, 57.7\% FP).

These results re-confirm the real toxicity mitigation found with automatic metrics.

\paragraph{Ethics statement} All annotators who worked on the human evaluation of detected toxicity are team members who are aware of the toxic nature of the samples prior to analyzing them.

\subsubsection{Toxicity key findings and contributions}

To summarize, our key findings and contributions include: (1) proposing a metric for speech toxicity detection for languages at scale (\asretox), (2) showing that while levels and types of added toxicity vary significantly as a function of language and dataset, added toxicity in our systems has a relatively low prevalence (varying from 0.11\% to 0.21\% across modalities, language directions, and datasets), and %
(3) our evaluation against the state-of-the-art shows that \mfourtlg reduces toxicity by 51\% across modalities and language directions in \fleurs and by 34\% in \holisticbias for \engx in \st{}. {\color{red}A manual analysis of toxicity detection outputs confirm this toxicity mitigation rate holds on real toxicity.} %

\subsection{Bias}
\label{apx:holisticbias}

\subsubsection{Motivation}

Unequal training datasets can lead to demographic and representational biases that affect our models and their generated outputs. These biases can adversely impact users by perpetuating allocation biases when used in situated contexts. In recent years, the MT field has made significant progress in uncovering \citep{prates:2020}, evaluating \citep{stanovsky-etal-2019-evaluating,renduchintala-etal-2021-gender, costa-jussa-etal-2022-evaluating,bentivogli-etal-2020-gender}, or even mitigating many of these forms of biases \citep{renduchintala-williams-2022-investigating}. However, much work lies ahead of us when it comes to this domain of research.

\textbf{Related work}  \multilingualholisticbias dataset \citep{costa2023multilingual} consists of an extension to \holisticbias. It contains translations for three different patterns and 118 descriptors, available in 50 different languages. Depending on whether gender inflection exists in a language, each language has one or two references. Each translated sentence includes the masculine, neutral and, when applicable, a feminine iteration.
The dataset enables quantification of gender biases across demographic aspects for \mt and has the highest language coverage at the time of writing. Previous work on this matter is mostly in text \citep{stanovsky-etal-2019-evaluating,renduchintala-etal-2021-gender,levy-etal-2021-collecting-large,costajussaetal:2022,renduchintala-williams-2022-investigating} and tend to be English-centric, with few demographic axes and multilingual references. Similar efforts for the speech modality remain sparse \citep{costa-jussa-etal-2022-evaluating, bentivogli-etal-2020-gender}.

\textbf{Contributions.} In this work, we used \multilingualholisticbias and its speech extension (described in the following section) to compare the performance of \st\ and \sst. The \engx direction allows comparing performance in the presence of masculine or feminine references, and the \xeng direction enables robustness comparisons in translations when we alter gender inflection. A typical example of the language pair of English-Spanish would be "I'm a homemaker" and the corresponding translations "Soy amo de casa" and "Soy ama de casa" in Spanish. When translating from English to Spanish, we can measure if the system overgeneralizes to one gender, while in the other direction, we can evaluate the robustness of the translation to gender inflection.

\subsubsection{Bias Experimental Framework}
\label{subsec:biasexperimental}

\paragraph{Dataset: Speech Extension of \multilingualholisticbias} 
In order to compare the performances across modalities (\sst and \st), we begin by extending the \multilingualholisticbias dataset from text to speech by using the TTS model\footnote{https://github.com/facebookresearch/fairseq/tree/main/examples/mms\#tts-1} provided by \cite{mms}. Due to the limitations of this TTS model in correctly generating speech for numbers, we manually converted all numerical numbers to words for each language. For instance, the sentence “I have friends who are 50 years old.” is transformed into “I have friends who are fifty years old.” After processing through TTS, we obtained the synthesized speech for 325 sentences across 19 languages. These languages are supported both by MMS-TTS and the \multilingualholisticbias\footnote{Arabic, Belarusian, Bulgarian, Catalan, Czech, Danish, German, Greek, French, Italian, Lithuanian, Latvian, Marathi, Dutch, Portuguese, Romanian, Russian, Slovak, Slovenian, Spanish, Swedish, Tamil, Thai, Ukrainian, Urdu.} dataset  \citep{costa2023multilingual}. For each of these languages (except English), we generated two speeches, one for each set of gendered texts.

\paragraph{Language directions and modalities}
We use this generated TTS data as input for \st\ and \sst and as a reference for \sst. We conducted the translations in two directions—\engx and \xeng. Concretely, in \xeng, we translated both masculine and feminine versions of the speech. 
It's worth noting that some target languages are not available in the \mfourt\ \sst model, so we performed translations on only 17 languages for the \sst~ task in the \engx direction. For \st in \engx, we have all languages included in the \multilingualholisticbias dataset (n=25). For reference, the complete language list used in our experiments can be found in Table \ref{tbl:rai-bias-language-summary}.
\begin{table}[ht]
    \begin{center}
    \small
   \begin{tabular}{lm{6cm}m{6cm}}
   \toprule
    & \hfil$\textbf{\xeng}$ & \hfil$\textbf{\engx}$ \\
    \midrule
    \textbf{\sst} & \multirow{2}{*}{\makecell[l]{\\arb,bul,cat,deu,ell,fra,lvs,mar,nld,\\por,ron,rus,spa,swe,tam,tha,ukr,urd}} & arb,cat,ces,dan,deu,fra,ita,nld,por,ron, rus,slk,spa,swe,tha,ukr,urd \\ \cline{1-1}\cline{3-3}
    \textbf{\st} & 
    & arb,bel,bul,cat,ces,dan,deu,ell,fra,ita, lit,lvs,mar,nld,por,ron,rus,slk,slv,spa, swe,tam,tha,ukr,urd \\
    \bottomrule
    \end{tabular}
    \end{center}
    \caption{List of language codes in the bias evaluation experiments, organized by task and language direction.}
    \label{tbl:rai-bias-language-summary}
\end{table}

\paragraph{Evaluation} %
In terms of evaluation metrics for \st,
we used chrF as reported in Table \ref{tbl:eval:metrics}, except that nw:2 was changed to nw:0. 
Instead of using \bleu as the quality metric, we used chrF because it is more equipped to handle shorter utterances, which better suits the evaluation of the
\multilingualholisticbias dataset. This dataset is relatively small (325 utterances) and with short sentences (on average, 6 words per utterance) \citep{costa2023multilingual}. 
In this context, we find chrF more adequate for comparison \citep{ma-etal-2019-results}, since \bleu quickly drops when not enough lengthy n-grams are matched. For \sst, we used \asrchrf.\footnote{The transcription is done by \whisperlarge and \whispermedium \citep{whisper} for \engx and \xeng respectively. chrF has been calculated the same way as \st except that in \sst the text from both prediction and reference are normalized.} 
and \blaser proposed in this work. It is worth noting that when evaluating \blaser, we included only 14 languages (including English)\footnote{The list of 
language codes for these 14 languages: arb,cat,deu,eng,fra,nld,por,ron,rus,spa,swe,tha,ukr,urd.} for the \engx direction (overlaps between the languages from the generated TTS data and the languages available in our \sst model). %
Additionally, since MMS-TTS generations are not deterministic, we repeated the measurements three times for both \sst and \st. The final metric values are then averaged to ensure robustness and accuracy in our evaluations.

\paragraph{Models} We used the \mfourtlg model and several different baselines. For \xeng\ \st, we employed \whisperlarge~\citep{whisper}. As for \xeng\ \sst, we used  \yourtts ~\citep{casanova2022yourtts} to generate synthesized speech from the output of \whisperlarge\ \st. For \engx\ \st, we utilized a cascaded system: \asr from Whisper Large-v2~\citep{whisper}, followed by \mt via \nllbmedium \citep{nllb2022}.
For \mfourtlg\ \st, we used a beam size of ten. For \mfourtlg\  \sst, we set the beam size to five for both the first pass decoder and the second pass decoder. As for the baseline, we set the beam size to five for \nllbmedium and used the default values for \whisperlarge and \yourtts.

\subsubsection{Bias evaluation results}

This section focuses on analyzing gendered translations when using neutral inputs (\engx) and the gap in translation performance between inputs that only differ in gender (\xeng). 

\begin{figure}[h!]
\centering
\includegraphics[width=\linewidth]{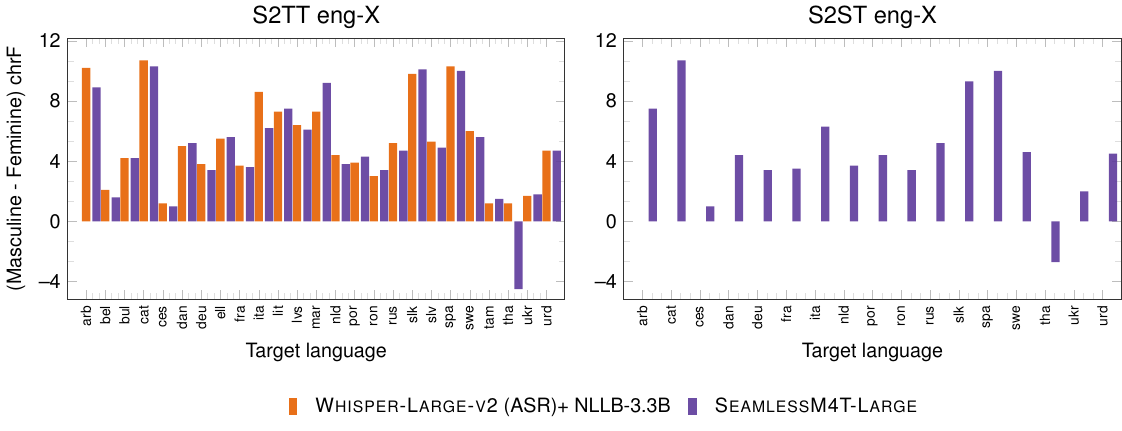}
\caption{Left: The chrF points difference between masculine and feminine forms for \engx\ \st using English speech as source and X text translation (masculine or feminine) as reference. Right: The \asrchrf points difference between masculine and feminine forms for \engx\ \sst using English speech as source and X text translation (masculine or feminine) as reference.}
\label{fig:mhb_en-xx_s2t_s2s}
\end{figure}

\begin{figure}[h!]
\centering
\includegraphics[width=\linewidth]{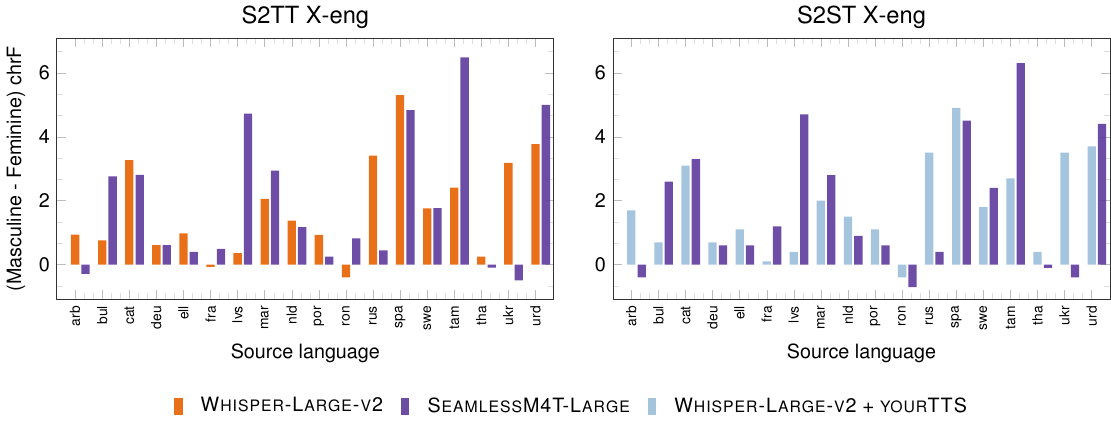}
\caption{(left) The chrF points difference between masculine and feminine for \xeng\ \st using X speech synthesized by the masculine or feminine version of the text and English text as a reference. (right) The \asrchrf points difference between masculine and feminine forms for \xeng\ \sst using X speech synthesized by the masculine or feminine version of text and English text as reference.}
\label{fig:mhb_m4t_vs_baseline}
\end{figure}

\begin{figure}[h!]
\centering
\includegraphics[width=\linewidth]{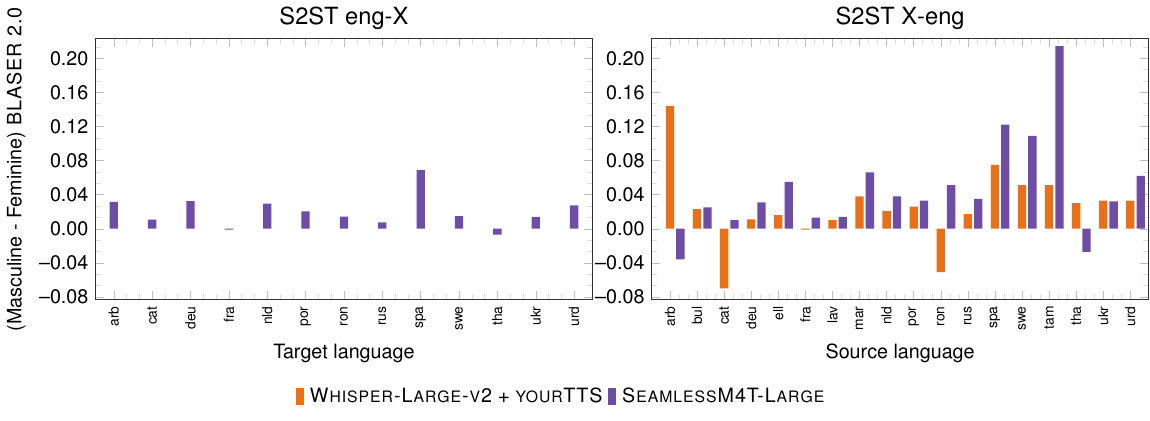}
\caption{(left) The supervised \blaser points difference between masculine and feminine forms for \engx\ \sst using English speech as the source and X text translation (masculine and feminine) as reference. The results are averaged from three experiments. (right) The supervised \blaser points difference for \xeng\ \sst using X speech synthesized by the masculine or feminine version of text and English text as reference.}
\label{fig:mhb_m4t_vs_baseline_blaserv2}
\end{figure}

\begin{table}[ht]
\centering
\small
\begin{tabular}{@{}lccccc@{}}
\toprule
&\multicolumn{5}{c}{\bf \st}\\\cmidrule{2-6}
{\bf Axis} & Masculine & Feminine & Average & Count & Diff\\
\midrule
Cultural & 11.4 & 9.5 & 10.4 & 350 & 1.9 \\
Body type & 14.2 & 12.9 & 13.6 & 3750 & 1.2 \\
Socioeconomic class & 14.6 & 13.3 & 13.9 & 400 & 1.3 \\
Religion & 15.5 & 13.7 & 14.6 & 1800 & 1.8 \\
Gender and sex & 16.0 & 15.1 & 15.5 & 1800 & 1.0 \\
Ability & 16.6 & 15.2 & 15.9 & 3300 & 1.3 \\
Race ethnicity & 17.4 & 15.7 & 16.5 & 900 & 1.7 \\
Characteristics & 18.2 & 16.2 & 17.2 & 1900 & 2.0 \\
Nationality & 18.1 & 16.7 & 17.4 & 300 & 1.4 \\
Sexual orientation & 18.5 & 16.7 & 17.6 & 700 & 1.8 \\
Age & 18.6 & 16.6 & 17.6 & 900 & 1.9 \\
\midrule
&\multicolumn{5}{c}{\bf \sst}\\\cmidrule{2-6}
{\bf Axis} & Masculine & Feminine & Average & Count & Diff\\

\midrule
Cultural & 12.2 & 10.3 & 11.3 & 238 & 1.9 \\
Body type & 14.2 & 13.0 & 13.6 & 2550 & 1.2 \\
Socioeconomic class & 14.4 & 13.1 & 13.7 & 272 & 1.3 \\
Religion & 16.3 & 14.5 & 15.4 & 1224 & 1.9 \\
Gender and sex & 16.7 & 15.7 & 16.2 & 1224 & 1.0 \\
Ability & 16.9 & 15.5 & 16.2 & 2244 & 1.4 \\
Age & 17.7 & 15.8 & 16.7 & 612 & 1.9 \\
Characteristics & 17.7 & 15.9 & 16.8 & 1292 & 1.8 \\
Race ethnicity & 18.0 & 16.4 & 17.2 & 612 & 1.7 \\
Sexual orientation & 18.4 & 16.9 & 17.7 & 476 & 1.5 \\
Nationality & 18.7 & 17.3 & 18.0 & 204 & 1.3 \\
\bottomrule
\end{tabular}
\caption{ Results on mean per axis (across descriptor, template, and language): chrF on \st (top) and \asrchrf on \sst (bottom) results. Columns (from left to right): masculine references, feminine references, average between the two, the total number of measurements (Count), and the difference between masculine and feminine (Diff). The rows are sorted in ascending order by the average chrF for \sst and \st, respectively. The axes are defined in \holisticbias—for more details, refer to Table 5 in the original paper \citep{smith-etal-2022-im}.}\label{tbl:m4t-en-xx-demographic}
\end{table}

\paragraph{\engx.} In our analysis, we utilize the masculine or the feminine human translations of the non-English language as references. The source for this analysis is the English (eng) \multilingualholisticbias dataset, comprising a collection of unique sentences with ambiguous gender. %
Figure \ref{fig:mhb_en-xx_s2t_s2s} shows the results per target language, evincing the following patterns:

\begin{itemize}
    \item In \mfourtlg \st, the translation quality deteriorates for all the languages except Thai when using the feminine reference, and is especially noticeable in languages like Catalan (with a significant 10.3 chrF points difference), Slovak (10.1), and Spanish (10.0). For the \whisperlarge + \nllbmedium combination, a decline in translation quality is observed across all languages. The highest differences are found in Catalan (10.7), Spanish (10.3), and Arabic (10.2). It's worth mentioning that the biases' distribution over languages is similar between \mfourtlg and the \whisperlarge + \nllbmedium combination, with Thai being the only exception. 
\item In \sst, we noticed similar trends in relation to \st, where translation quality is lowered in all languages (except Thai) when assessing with the feminine reference. The highest differences are with Catalan (10.7 \asrchrf points difference), Spanish (10.0), and Slovak (9.3). %

\end{itemize}

The left panel of Figure \ref{fig:mhb_m4t_vs_baseline_blaserv2} shows the results for automatic speech evaluation by way of \blaser. We observe similar trends in the \asrchrf metric. The translation quality deteriorates by an average of 0.02 supervised \blaser points across languages when evaluating with the feminine reference for all languages except Thai. Interestingly, the evaluation for French reveals a negligible difference. The highest differences are found in Spanish (0.07), followed by German (0.03).

These differences show that when no gender information is available in the source sentence, the model will prefer to translate to the masculine form in the target language. Note that for some languages (like Spanish or French), the plural masculine form is indistinguishable from the plural generic form.

\paragraph{\xeng.} Our main objective is to assess the translation quality when starting from a gendered sentence and translating it into English. %
As such, we aim to measure the model's robustness with regard to gender bias and its ability to handle translations between languages that mark grammatical gender towards English. Figure \ref{fig:mhb_m4t_vs_baseline} shows the results per source language for \mfourtlg and \whisperlarge or \whisperlarge + \yourtts. We observe that:

\begin{itemize}
    \item In \st, the performance is better when translating from the masculine reference for most languages (15 out of 18 for \mfourtlg and 16 out of 18 for the \whisperlarge). However, they have different biases towards different languages. The highest differences between the masculine and feminine forms in \mfourtlg are with Tamil (6.4 chrF points difference) and Urdu (5.0).\footnote{We find that in our experiment, Arabic shows the bias toward the cases when translated from feminine version, which contrasts with the findings in the \multilingualholisticbias \citep{costa2023multilingual} where Arabic exhibited significantly higher performance when translating from the masculine version. We hypothesize that this difference is attributed to our use of a different language code "ara" instead of "arb" when applying the MMS-TTS.} On the other hand, the highest differences in \whisperlarge are with Spanish (5.3), Urdu (3.8), and Russian (3.4).

\item In \sst, we observe similar outcomes to those in \st. The model quality is mostly better when translating from masculine cases, as evident in 14 out of 18 languages for \mfourtlg and 17 out of 18 for the \whisperlarge + \yourtts combination. The most significant differences between masculine and feminine sources in \mfourtlg are found in Tamil (with an \asrchrf point difference of 6.3) and Spanish (4.5). The highest differences in \whisperlarge are in Spanish (4.9), Urdu (3.7), and Ukrainian (3.5). %

\end{itemize}

The right panel of Figure \ref{fig:mhb_m4t_vs_baseline_blaserv2} demonstrates the performance comparison using \blaser. Like the findings in \asrchrf, the translation quality generally improves when translating from masculine cases, which is observed in 16 out of 18 languages and 15 out of 18 languages for \mfourtlg and \whisperlarge + \yourtts\ respectively. The highest differences for \mfourtlg are with Tamil (0.21 supervised \blaser points), Spanish (0.12), and Swedish (0.11). For \whisperlarge + \yourtts, the highest differences are found in Arabic (0.14), Spanish (0.075), and Tamil (0.05).

\paragraph{Average comparison across directions and modalities} Table \ref{tbl:m4t_baseline_xx_en_average} presents the average scores per gender and the comparison with the corresponding baseline.\footnote{For \engx\ \sst, we report only the performance for the \mfourtlg in absence of baseline.} $\Delta$ corresponds to the relative variation between genders computed as follows:
\begin{equation*}
\Delta = \omega(M-F)/\omega(min(M,F)), \omega \in \{ \textsc{chrF}, \asrchrf, \blaser \}
\end{equation*}
As mentioned, in \engx, we evaluated translations from neutral to gendered forms and observed the overgeneralization towards one gender, whereas in \xeng, we evaluated the robustness of translating content that only differs in their gender inflection. Focusing sorely on the results of \mfourtlg, we noticed that, except for the evaluation outcomes in \blaser, the difference in performance between the masculine and feminine forms is more pronounced for overgeneralization than for robustness. Turning our attention to the performance comparison, we find that when it comes to overgeneralization, \mfourtlg slightly outperforms \whisperlarge + \nllbmedium. As for the outcome related to the robustness, \mfourtlg falls short against \whisperlarge in \st but outperforms \whisperlarge + \yourtts in \sst. We further noticed a higher percentage gap in \asrchrf than for \blaser. This may imply that ASR (from \asrchrf) adds some extra biases.

\begin{table}[ht]
\centering
\small
\begin{tabular}{ccccc}
\toprule
&&\multicolumn{3}{c}{\textbf{\engx \mfourt/\whisperlarge + \nllbmedium}}\\\cmidrule{3-5}
& & Feminine Reference & Masculine Reference & $\Delta$ \%\\
\midrule
{\st}& chrF & 45.0/\textbf{47.4} & 49.9/\textbf{52.7} & \textbf{10.9}/11.2\\
\midrule
\multirow{2}{*}{\sst} & \asrchrf & 44.9 & 49.7 & 10.6\\
& \blaser & 3.6 & 3.7 & 0.6\\
\midrule
&& \multicolumn{3}{c}{\textbf{\xeng\ \mfourt/\whisperlarge (+ \yourtts)}}\\\cmidrule{3-5}
& & Feminine Source & Masculine Source & $\Delta$ \%\\
\midrule
{\st} & chrF & \textbf{52.4}/50.4 & \textbf{54.3}/52.1 & 3.7/\textbf{3.4}\\
\midrule
\multirow{2}{*}{\sst} & \asrchrf & \textbf{53.1}/52.2 & \textbf{55.0}/54.0 & \textbf{3.5}/\textbf{3.5}\\
& \blaser & \textbf{3.5}/2.7 & \textbf{3.6}/2.8 & 
\textbf{2.9}/3.7\\
\bottomrule
\end{tabular}
\caption{The averaged points across modalities and genders for assessing the overgeneralization (\engx) and the robustness (\xeng). ${\Delta}$ represents the relative difference between masculine and feminine (${\Delta = \omega(M-F)/\omega(min(M,F)), \omega \in \{ chrF, \asrchrf, \blaser \}}$).}
\label{tbl:m4t_baseline_xx_en_average}
\end{table}

\paragraph{Demographic analysis}
We conducted a similar analysis to that in \cite{costa2023multilingual}. Table \ref{tbl:m4t-en-xx-demographic} shows the mean chrF or \asrchrf at the sentence level on the \multilingualholisticbias axes translations, averaged across descriptors, templates, languages, and masculine vs. feminine references. Among all the axes, we find that cultural, body type, socioeconomic class, and religion are most sensitive to quality disruptions. Furthermore, when considering the difference between masculine and feminine references and the number of effective samples, we observe that both \sst and \st show the highest bias in the ability, body type, religion, and characteristics axes. These observations align with the findings reported in \cite{costa2023multilingual} pertaining to \mt.

\subsubsection{Gender data representation}

Based on our concurrent work \citep{muller-etal-2023-pipeline}, we discuss the representation bias of several datasets by focusing on how different genders are represented using lexical matching. %
The closest work that studies gender representation in data is \cite{choubey-etal-2021-gfst}, where the authors took on this research question using a synthetic dataset. The authors, however, did not share the details of the lexical nouns used to extract this representation. 

\begin{figure}[ht!]
\centering
\includegraphics[width=0.8\linewidth]{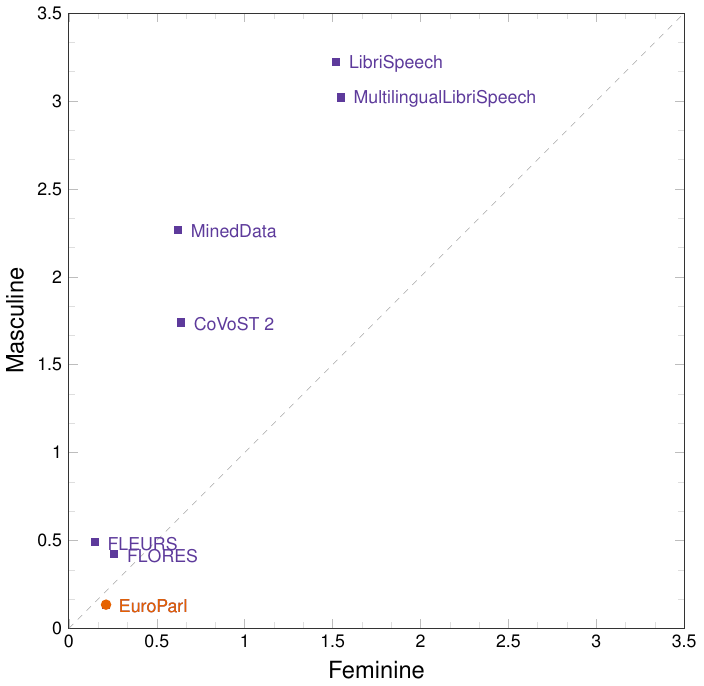}
\caption{Gender representation of English evaluation datasets (EuroParl, \flores, \fleurs, \covost, LibriSpeech and MultilingualLibriSpeech), and training mined data (\mineddata). Vertical axis show the percentage of masculine representation and horizontal axis show the percentage of feminine representation.}\label{fig:datarepr}
\end{figure}

\holisticbias \citep{smith-etal-2022-im} provides a list of gendered nouns and pronouns. We rely on this list to track how many sentences in our data sets contain gendered markers.
Since our analysis is only in English, we tokenize for word boundaries using python word-boundary regular expression ($\backslash$b). As lexical terms, we limited the vocabulary to make our approach scalable to multiple languages \citep{muller-etal-2023-pipeline}. This vocabulary includes 11 masculine nouns\footnote{man, men, bro, bros, guy, guys, boy, boys, father, fathers, dad, dads, son, sons, husband, husbands, grandfather, grandfathers, grandpa, grandpas, brother, brothers.}; 4 masculine pronouns,\footnote{he, him, his, himself.}; 10 feminine nouns\footnote{woman, women, lady, ladies, girl, girls, mother, mothers, mom, moms, daughter, daughters, wife, wives, grandmother, grandmothers, grandma, grandmas, sister, sisters}, and 4 feminine pronouns\footnote{she, her, hers, herself.}. We matched single words, hence we report the number of words out of the total number of words in the dataset. %
 Figure \ref{fig:datarepr} summarizes the results of the gender representations for several English evaluation and training datasets. Results show that masculine representation is predominant in most of the datasets. Extremely low representations of gender (i.e., low matching of gendered words based on our selected vocabulary) are found in EuroParl, \fleurs, and \flores datasets. However, this low representation is the trade-off to make our approach scalable to multiple languages. This scalable effort on data characterization could potentially be used for the purpose of balancing datasets to mitigate gender biases.

\subsubsection{Bias Key Findings}

In this section, we conducted a set of comprehensive evaluations on translation biases for \st\ and \sst. We demonstrate the following: (1) in the absence of gender information, \mfourtlg exhibits an average preference of $\sim$10\% towards translating to the masculine form (for both modalities); (2) utilizing feminine form as the source input leads to lower quality English translations compared to its masculine counterpart, showing a lack of robustness against gender inflection by $\sim$3\% ; (3) \mfourtlg gets comparable bias results to the state-of-the-art, %
 and (4) our gender representation analysis reveals an overrepresentation of masculine lexica compared to feminine in the analyzed datasets.
More importantly, these findings pave the way towards standardizing the bias evaluation of speech translation at a massive scale.

\subsection{Limitations}\label{sec:appendix:rai:rai_limitations}
Due to the lack of available model-based techniques that could be applied to added toxicity or gender imbalance detection in this multimodal and massively multilingual setting, we used string-matching techniques that present known limitations.

First, the use of toxicity lists with ETOX for added toxicity detection shares the same limitations with other word list-based detection techniques, which were previously discussed at length in \cite{nllb2022} and \cite{costajussa2023toxicity}. Briefly, the two main limitations of word list-based detectors are (1) their tendency to over-detect terms that are only toxic in specific contexts, and (2) their reliance on precise tokenization, which is more difficult to achieve in non-segmenting or highly agglutinative languages. When dealing with speech outputs, the process of using ASR before lexical matching adds one more source of error, which tends to lead to false negatives. This particularly affects the directions of \engx, since ASR tends to be of lower quality for non-English languages.

Second, the use of noun lists for the detection of linguistic gender imbalance in large datasets shares all of the limitations of word list-based techniques previously stated, along with the added difficulty of relying on linguistic gender clues as a proxy for overall gender representation. Indeed, linguistic gender assignment does not follow the same pattern across all languages that mark gender, especially when it comes to inclusive plural forms (i.e., plural forms referring to groups that include more than one gender). In addition to general limitations, the use of a specific and limited set of 30 nouns (selected to mirror those used to build the \holisticbias dataset) does not guarantee that results can be generalized to all other sets of nouns that could be used to investigate gender representation (e.g., occupation nouns).

\section{Social Impact \& Conclusion}
\label{sec:conclusion}

Human communication is multisensorial—we take in sensory input from several modalities to process information in a dynamic way \citep{holler2019multimodal}. In multilingual contexts, advancements in text-based machine translation have given rise to tools that help individuals communicate and learn in languages where proficiency is low \citep{lee2023effectiveness}. That said, while foundational models such as NLLB \citep{nllb2022} push \mt{} beyond 200 languages, direct speech translation has yet to achieve similar strides. To bridge this gap, we created a massively multilingual and multimodal machine translation system that paves the way for the next generation of speech translation technologies.

Using novel data and modeling approaches to combine \sst{}, \ttst{}, \st{}, \mt{}, and \asr{} in a single model, our main contributions are as follows. First, we built a new LID model aligned with our language coverage and conducted speech mining with the help of the newly conceived SONAR—a multilingual and multimodal sentence embedding space—to create a corpus of automatically aligned speech translations of more than \totalminedhours hours. By fusing four building blocks, (1) \nllbv, a massively multilingual T2TT model, (2) w2v-BERT 2.0, a speech
representation learning model pre-trained on unlabeled speech audio data, (3) T2U, a text-to-unit sequence-to-sequence model, and (4) HiFi-GAN—a multilingual vocoder for synthesizing
speech from units, we built a unified model that covers \sst{} from \nsslangs{} languages to English (\nsslangs{}-eng), English to \ntslangs{} languages (eng-\ntslangs{}), and \st{} for \nsslangs{}-eng and eng-\ntextlangs{} languages. Notably, compared to previous work on \sst{}, which primarily serves translations into English and not vice versa, \mfourt is capable of performing translation from English towards \ntslangs{} directions. When it comes to \st{}, \mfourt achieves an improvement of 20\% BLEU over the previous state-of-the-art in \st{} translation. Preliminary human evaluations of \st{} outputs evinced similarly impressive results; for translations from English, XSTS scores for 24 evaluated languages are consistently above 4 (out of 5). For into English directions, we see significant improvement over \whisperlarge's baseline for 7 out of 24 languages.
We then evaluated our model for robustness, revealing that \mfourt is more robust than \citep{whisper} when it comes to background noises and
speaker variations. By also including results of the level of added toxicity and gender bias, we hope to motivate future work targeting mitigation efforts.

Made with the goal of promoting accessibility, we open-source all contributions of our work, including two sizes of our model to ensure that even researchers with limited computing resources can use our work. In the section below, we discuss the potential social impact of \mfourt by focusing on its downstream possibilities.

\subsection{Augmenting world-readiness}

The world we live in has never been more interconnected—the global proliferation of the internet, mobile devices, communicative platforms, and social media exposes individuals to more multilingual content than ever before \citep{zuckerman_2008}. The current social order places a demand on a person’s “world-readiness” \citep{ACTFL}, a measure of how competent a person is to take on the polyglot world. Initially developed in the context of language learning, world-readiness underscores the importance of being able to communicate in languages beyond one’s mother tongue for both instrumental (i.e., employment or schooling) and cultural reasons (i.e., to become a global citizen). That said, while we believe that language acquisition should remain a key mechanism for boosting one’s world-readiness, we acknowledge that doing so requires mental and material resources many people may not possess.

The downstream applications that \mfourt supports could allow on-demand access to world-readiness by streamlining multilingual exchange across various contexts. Akin to what \mt{} has accomplished for bridging the comprehension of multilingual texts, \mfourt could have the same effects for speech. Research shows that contrary to one’s native language, where speech is more organically acquired than reading or writing \citep{liberman1992relation}, this tendency is flipped when it comes to foreign languages \citep{cheng1999language}. In other words, speech is often deemed more challenging than reading or writing in a foreign language context. \mfourt-supported applications could act as a co-piloting mechanism that supports users in multilingual conversations and boost their confidence in speech-heavy interactions. As speech-based interfaces (i.e., audio assistants, voice memos, live transcriptions, etc.) and auditory content (i.e., podcasts, audiobooks, short-form videos, etc.) become ever more present in people’s lives, downstream applications enabled by \mfourt could allow a greater variety of multilingual experiences and in ways that feel more natural and dynamic than its text-based counterparts.

From an inclusion standpoint, \mfourt’s focus on multimodality could make a meaningful difference in augmenting the world-readiness of those with accessibility needs and those whose languages contain multiple writing systems (as aforementioned in \ref{sec:appendix:problem}. For many who lack reading or writing skills, or are unable to rely on sight (i.e., people who are blind or with visual impairment), voice-assisted technologies are essential to how they communicate and stay connected \citep{belekar2020voice}. The ability to translate speech not only gives these groups more comprehensive access to information beyond their native languages, but also in a manner that is better suited for their communicative needs. Additionally, recognizing that some languages may have script variance, \mfourt’s offers up affordances that help circumvent the multiscript conundrum. For languages that do not have standardized writing systems, investments in speech recognition and translation may be instrumental in preventing endangerment. We hope that our effort can help contribute to this important movement.

\subsection{Future work}

As is the case with most technologies, the distribution of benefits varies based on user demographics and social situation \citep{wang2023human}. While we make the case that \mfourt could augment world-readiness by lowering the barriers in cross-lingual communication, some users may experience more difficulties using our work than others. For instance, like many other speech technologies, \mfourt’s \asr{} performance may vary based on gender, race, accent, or language \citep{koenecke2020racial,ngueajio2022hey}. Moreover, our system's performance when it comes to translating slang or proper nouns may also be inconsistent across high and low-resource languages. 

Another challenge for \sst{} is that speech hinges on immediate reception and feedback compared to written language. In other words, a speaker is limited in their ability to ascertain the quality of an output or make “edits” in a live conversation. Without the ability to plan and revise with the help of back-translation or a native speaker, \sst{} may carry higher degrees of interactional risks when it comes to mistranslations or toxicity. We urge researchers and developers who fine-tune or build products using \mfourt to think critically about design features that could help users circumvent these potential obstacles. On a related note, we believe that \mfourt-fueled applications should best be viewed as an augmentation device that assists in translation rather than a tool that replaces the need for language learning or reliable human interpreters. This reminder is especially pertinent in high-stakes situations involving legal or medical decision-making.

Finally, speech is not spoken text—it encompasses a suite of prosodic (i.e., rhythm, stress, and intonation) and emotional components that deserve further research \citep{elbow1985shifting}. To create \sst{} systems that feel  organic and natural, more research should be directed at output generation that preserves expressivity \citep{trilla2012sentence}. In addition, the consummate realization of the Babel Fish requires deeper investments into research on low-latency speech translation. Developing systems that enable streaming (i.e., incrementally translating an input sentence as it is being presented) may increase the adoption of such systems in industry or educational contexts \citep{iranzo2022simultaneous,rybakov2022streaming}. We hope that \mfourt opens up new possibilities for both of these research areas.

\section*{Acknowledgments}
We want to extend our gratitude to those who made this work possible. Thanks to Sergey Edunov and Angela Fan for helping shape the earlier stages of the project; Shruti Bhosale, Vedanuj Goswami, Fernando Hernandez and Yun Tang for their help to build better models; Mingda Chen for his contributions to \textsc{Blaser 1.0}; Kiryl Klyushkin for his help to build better experiences; Artyom Kozhevnikov for his contributions to \fairseq and \sonar inference; Zhaoheng Ni and Xiaohui Zhang for benchmarking audio denoising models; Neil Seejoor and Mark Duppenthaler for their help in setting up the demo; 
Vedanuj Goswami, Samuel Hsia, Bilge Acun-Uyan and Carole-Jean Wu who helped with efficiency optimizations;
Belen Alastruey, Mohamed Anwar, Heng-Jui Chang, HyoJung Han, Chao-Wei Huang, Hui Lu, Siqi Ouyang, Yifan Peng, Phillip Rust, Jiatong Shi, Neha Verma, Sung-Lin Yeh, and all of our interns and residents for the energy and candid discussions they brought to the team; 
Mike Clark, Lauren Cohen, Jennifer Pak, Harrison Rudolph for their guidance; 
Emily Astbury, Lydia Baillergeau, Dana Beaty, Jeffrey Bennett, Jon Carvill,  Anne Davidson, Aiman Farooq, Ashley Gabriel, Gopika Jhala, Christopher Johnson, Steph Miles, Ana Paula Kirschner Mofarrej, Raghu Nayani, Alyssa Newcomb, Tamara Piksa, Michelle Restrepo, Noha Rizk, Adébissy Tharinger, who helped our research reach new audiences; 
Geeta Chauhan, Ankit Gunapal, Caleb Ho, Dinesh Kannappan, Apostolos Kokolis, Teng Li, Matthias Reso, Shubho Sengupta, Hamid Shojanazeri, Xinyuan Zhang for assisting us with compute resources and infrastructure;
Emmanuel Dupoux and Eric Michael Smith for their feedback on the paper;
Chris Moghbel,  Manohar Paluri, Joelle Pineau, Laurens van der Maaten, and Mary Williamson for their continued support of the project.

\bibliography{bibliography}%

\newpage
\begin{appendices}
\section{FAIRSEQ2}

FAIRSEQ2 is an open-source library of sequence modeling components that provides researchers and developers with building blocks for machine translation, language modeling, and other sequence generation tasks, specifically around text and audio data format. FAIRSEQ2 is distributed with an MIT license and is available on GitHub at \href{https://github.com/pytorch/fairseq2}{https://github.com/pytorch/fairseq2}.

FAIRSEQ2 features: 
(i) state-of-the-art implementations of transformers and their components (transformer layers, embedding layers, layernorms, attention blocks, etc.);%
(ii) \texttt{fairseq2.data} --  a scalable pipeline API that enables text and audio data pre-processing, transformation, shuffling, and batching in a streaming manner, allowing training over multi-terabyte datasets without explicit data preparation steps or data loading timeouts; (iii) core building components for efficient model training (optimizers, LR schedulers, loss implementations); (iv) sequence generators for optimized inference with incremental beam search. %

Following the spirit of its predecessor FAIRSEQ~\citep{ott2019fairseq}, FARSEQ2 was built with extensibility in mind. The library-like structure of the code enables effortless component drop-ins, including those initially written in FAIRSEQ. 
We expect continuous populating of the library with new components by us and by the open-source community in the following years.

Another guiding principle for FAIRSEQ2 is a clear separation of core and experimental code. 
The original FAIRSEQ has become a hub for numerous research ideas. 
Often they were added in the form of if-else statements mixed with the core functionality. 
Over time, the number of such if-else statements and associated command line options has grown, with each option poorly supported and often subtly incompatible with other options. 
To prevent this scenario, in FAIRSEQ2, all basic components are designed with the “dependency inversion” principle, making it possible to compose them easily. 
Existing model architectures can be modified with just a few lines of code without requiring copy/pasting large amounts of code. All plug-ins and modifications exist as separate components, not interfering with the parent blocks and not hindering access to them for other users. Larger efforts (like \unity or \sonar described in this paper) are moved into separate repositories and use FAIRSEQ2 as a dependency.

We acknowledge the wide range of training and execution environments for Deep Learning models that exist today (from a single-container training via on-demand Cloud Computing Services to huge LLMs training jobs running on exaFLOPS supercomputers with tens of thousands GPUs; from very limited inference capabilities of edge devices to the power of accelerated inference on ASICs). To meet the diverse expectations of these environments, FAIRSEQ2 has shifted from the idea of a self-contained single-stop for all training, evaluation, and inference pipelines towards a set of independent components that can be used and extended outside of FAIRSEQ2. We put an emphasis on compatibility with the existing alternatives in PyTorch and other Deep Learning frameworks, following common API conventions and inheriting from the same base classes. That guarantees effortless drop-in replacement of components from different origins. The user is offered a wide range of usage scenarios: from implementing a complete pipeline using FAIRSEQ2 to fusing multiple Deep Learning frameworks in their project, or even picking a single block like the efficient implementation of an optimizer.

\section{Data Statistics}

We provide in \Cref{tbl:modeling:s2tdata} statistics of \asr and \st data (in hours of speech audio) used to train the \xt models of \mfourt. Similarly, we provide in \Cref{tbl:modeling:s2stdata} statistics of \sst training data.

\begin{table}[!htb]
\scriptsize
    \centering
    \vspace{-53pt}
    \hspace{-30pt}\begin{minipage}[t]{0.45\textwidth}\vspace{0pt}
    \begin{tabular}{@{}ccccccc@{}}
        \toprule
        \multirow{2}{*}{\makecell[c]{\bf language\\\bf code}} & \bf ASR & \multicolumn{2}{c}{\bf \st~\xeng} & \multirow{2}{*}{\bf Resource} & \multicolumn{2}{c}{\bf \st~\engx} \\\cmidrule{3-4}\cmidrule{6-7}
          &  & P & M &  &  P & M \\
        \midrule
        {\bf Total} & 40,012 & 50,596 & 12,682 &  & 17,6827 & 5,701 \\
        \midrule
        afr & 106 & 101 &  & low & 2069 &  \\
        amh & 54 & 49 &  & low & 1921 &  \\
        arb & 934 & 942 & 400 & high & 1959 & 200 \\
        ary & 97 & 95 &  & low & 1776 &  \\
        arz & 93 & 92 &  & low & 2014 &  \\
        asm & 77 & 68 &  & low & 1698 &  \\
        ast & 0 & 0 &  & zero-shot & 0 &  \\
        azj & 95 & 94 &  & low & 1901 &  \\
        bel & 1160 & 1157 &  & high & 1641 &  \\
        ben & 338 & 320 & 400 & high & 1987 & 200 \\
        bos & 99 & 99 &  & low & 2113 &  \\
        bul & 103 & 102 &  & low & 1881 &  \\
        cat & 1767 & 1758 & 400 & high & 1781 & 200 \\
        ceb & 0 & 0 &  & zero-shot & 2020 &  \\
        ces & 189 & 442 & 400 & high & 2066 & 200 \\
        ckb & 93 & 92 &  & low & 2001 &  \\
        cmn & 9784 & 9027 & 400 & high & 1947 &  \\
        cym & 100 & 96 & 400 & medium & 1676 &  \\
        dan & 161 & 371 & 400 & medium & 1954 & 200 \\
        deu & 3354 & 3490 &  & high & 2043 & 200 \\
        ell & 345 & 339 &  & medium & 1725 &  \\
        eng & 3845 & 0 &  & high & 0 &  \\
        est & 133 & 130 & 400 & medium & 1803 & 200 \\
        eus & 276 & 265 &  & medium & 1998 &  \\
        fin & 182 & 449 & 400 & high & 1933 & 200 \\
        fra & 2123 & 2247 &  & high & 2304 & 200 \\
        fuv & 0 & 0 &  & zero-shot & 0 &  \\
        gaz & 0 & 0 &  & zero-shot & 1766 &  \\
        gle & 56 & 55 &  & low & 1973 &  \\
        glg & 123 & 121 &  & low & 2116 &  \\
        guj & 143 & 138 &  & low & 1990 &  \\
        hau & 0 & 0 &  & zero-shot & 0 &  \\
        heb & 96 & 96 &  & low & 2092 &  \\
        hin & 148 & 143 & 400 & medium & 2066 & 200 \\
        hrv & 308 & 219 &  & medium & 2119 &  \\
        hun & 260 & 474 &  & medium & 1900 &  \\
        hye & 148 & 146 &  & low & 1696 &  \\
        ibo & 35 & 28 &  & low & 1738 &  \\
        ind & 250 & 254 & 400 & medium & 1818 & 200 \\
        isl & 132 & 130 &  & low & 2059 &  \\
        ita & 591 & 910 & 400 & high & 2278 & 200 \\
        jav & 302 & 301 &  & medium & 2122 &  \\
        jpn & 381 & 15141 & 400 & high & 1798 & 200 \\
        kam & 0 & 0 &  & zero-shot & 0 &  \\
        kan & 124 & 121 & 208 & low & 1954 &  \\
        kat & 195 & 185 &  & low & 1639 &  \\
        kaz & 330 & 327 &  & medium & 1895 &  \\
        kea & 0 & 0 &  & zero-shot & 0 &  \\
        khk & 152 & 148 &  & low & 1657 &  \\
        khm & 191 & 187 &  & low & 1661 &  \\
        kir & 129 & 123 &  & low & 1839 &  \\
        kor & 387 & 201 & 400 & medium & 2125 &  \\
        lao & 200 & 190 &  & low & 1959 &  \\
        lin & 0 & 0 &  & zero-shot & 0 &  \\
        \bottomrule
\end{tabular}
\end{minipage}\hfill
\begin{minipage}[t]{0.45\textwidth}\vspace{0pt}
    \begin{tabular}{@{}ccccccc@{}}
        \toprule
        \multirow{2}{*}{\makecell[c]{\bf language\\\bf code}} & \bf ASR & \multicolumn{2}{c}{\bf \st~\xeng} & \multirow{2}{*}{\bf Resource} & \multicolumn{2}{c}{\bf \st~\engx} \\\cmidrule{3-4}\cmidrule{6-7} 
          &  & P & M &  &  P & M \\
        \midrule\\
        \midrule
        lit & 40 & 283 &  & low & 1920 &  \\
        ltz & 0 & 0 &  & zero-shot & 0 &  \\
        lug & 369 & 368 &  & medium & 1890 &  \\
        luo & 0 & 0 &  & zero-shot & 1975 &  \\
        lvs & 100 & 98 &  & low & 1779 &  \\
        mai & 0 & 0 &  & zero-shot & 2004 &  \\
        mal & 110 & 57 &  & low & 1754 &  \\
        mar & 112 & 108 &  & low & 1848 &  \\
        mkd & 145 & 143 &  & low & 1918 &  \\
        mlt & 157 & 151 & 74 & low & 1699 & 200 \\
        mni & 0 & 0 &  & zero-shot & 1257 &  \\
        mri & 0 & 0 &  & zero-shot & 0 &  \\
        mya & 137 & 125 &  & low & 1860 &  \\
        nld & 1734 & 1780 & 400 & high & 2249 & 200 \\
        nor & 214 & 193 & & low & 2134 & \\
        npi & 153 & 129 &  & low & 1714 &  \\
        nso & 0 & 0 &  & zero-shot & 0 &  \\
        nya & 103 & 99 &  & low & 2058 &  \\
        oci & 0 & 0 &  & zero-shot & 0 &  \\
        ory & 89 & 86 &  & low & 1721 &  \\
        pan & 196 & 193 &  & low & 1641 &  \\
        pbt & 131 & 121 & 400 & medium & 1847 &  \\
        pes & 386 & 68 &  & low & 1980 &  \\
        pol & 341 & 446 & 400 & high & 1914 & 200 \\
        por & 269 & 246 & 400 & medium & 2250 & 200 \\
        ron & 182 & 443 & 400 & high & 2131 & 200 \\
        rus & 264 & 144 & 400 & medium & 2161 & 200 \\
        sat & 0 & 0 &  & zero-shot & 0 &  \\
        slk & 142 & 390 & 400 & medium & 1931 & 200 \\
        slv & 107 & 370 &  & low & 1800 &  \\
        sna & 0 & 0 &  & zero-shot & 2067 &  \\
        snd & 0 & 0 &  & zero-shot & 1958 &  \\
        som & 143 & 140 &  & low & 1851 &  \\
        spa & 1514 & 1285 &  & high & 2505 & 200 \\
        srp & 101 & 98 &  & low & 1910 &  \\
        swe & 129 & 91 &  & low & 1810 & 200 \\
        swh & 361 & 50 & 400 & medium & 1930 & 200 \\
        tam & 256 & 64 & 400 & medium & 1569 &  \\
        tel & 89 & 80 & 400 & medium & 1934 &  \\
        tgk & 99 & 98 &  & low & 1820 &  \\
        tgl & 99 & 93 & 400 & medium & 2015 &  \\
        tha & 189 & 59 & 400 & medium & 1941 & 101 \\
        tur & 169 & 100 & 400 & medium & 2135 & 200 \\
        umb & 0 & 0 &  & zero-shot & 0 &  \\
        ukr & 132 & 75 & 400 & medium & 2052 & 200 \\
        urd & 185 & 145 & 400 & medium & 1844 & 200 \\
        uzn & 166 & 96 & 400 & medium & 1801 & 200 \\
        vie & 194 & 151 & 400 & medium & 2396 & 200 \\
        wol &  &  &  & zero-shot &  &  \\
        xho & 0 & 0 &  & zero-shot & 0 &  \\
        yor & 132 & 130 &  & low & 1384 &  \\
        yue & 167 & 124 &  & low & 1931 &  \\
        zlm & 155 & 161 &  & low & 0 &  \\
        zul & 62 & 55 &  & low & 2063 &  \\
        \bottomrule
\end{tabular}
\end{minipage}
\caption{\small Statistics of \asr{} and \st{} data used to train our \mfourt{} model. We list the data size in hours of speech between primary (P) i.e.,
open-source \st and pseudo-labeled ASR data,
and mined (M). For each language we distinguish between \engx{} for translating from English into that language, and \xeng{} for translating into English.
We qualify as high-resource, languages with more than 1000 hours of supervision. Languages with between 500 and 1000 hours are dubbed medium-resource, and languages with less than 500 hours are low-resource. If a language is not supervised during the 1+2 stages of finetuning then it is evaluated as zero-shot.
}
\label{tbl:modeling:s2tdata}
\end{table}
\begin{table}[!htb]
\scriptsize
    \centering
    \vspace{-53pt}
    \hspace{-30pt}\begin{minipage}[t]{0.45\textwidth}\vspace{0pt}
    \begin{tabular}{@{}ccccc@{}}
        \toprule
        & \multicolumn{4}{c}{\bf \sst}\\\cmidrule{2-5}
         & \multicolumn{2}{c}{\xeng{}} & \multicolumn{2}{c}{\engx{}} \\
        \cmidrule(r){2-3}
        \cmidrule(l){4-5}
        {\bf Language} & Primary & Mined & Primary & Mined \\
        \midrule
        {\bf Total} & 26,254 & 23,171 & 49,425 & 21,983 \\\midrule
        afr & 100 & 0 & 0 & 0 \\
        amh & 46 & 0 & 0 & 0 \\
        arb & 898 & 736 & 895 & 681 \\
        ary & 94 & 0 & 0 & 0 \\
        arz & 91 & 0 & 0 & 0 \\
        asm & 62 & 0 & 0 & 0 \\
        ast & 0 & 0 & 0 & 0 \\
        azj & 92 & 0 & 0 & 0 \\
        bel & 285 & 0 & 0 & 0 \\
        ben & 292 & 246 & 652 & 221 \\
        bos & 99 & 0 & 0 & 0 \\
        bul & 101 & 0 & 0 & 0 \\
        cat & 276 & 278 & 692 & 293 \\
        ces & 437 & 522 & 832 & 528 \\
        ckb & 89 & 0 & 0 & 0 \\
        cmn & 350 & 1,318 & 857 & 1,388 \\
        cym & 93 & 197 & 700 & 185 \\
        dan & 368 & 420 & 684 & 450 \\
        deu & 2,570 & 1,661 & 962 & 1,618 \\
        ell & 330 & 0 & 0 & 0 \\
        est & 128 & 502 & 691 & 477 \\
        eus & 263 & 0 & 0 & 0 \\
        fin & 446 & 442 & 684 & 414 \\
        fra & 2,255 & 2,438 & 937 & 2,303 \\
        gle & 55 & 0 & 0 & 0 \\
        glg & 120 & 0 & 0 & 0 \\
        guj & 135 & 0 & 0 & 0 \\
        hau & 78 & 0 & 0 & 0 \\
        heb & 96 & 0 & 0 & 0 \\
        hin & 138 & 466 & 656 & 430 \\
        hrv & 218 & 0 & 0 & 0 \\
        hun & 468 & 0 & 0 & 0 \\
        hye & 141 & 0 & 0 & 0 \\
        ibo & 24 & 0 & 0 & 0 \\
        ind & 248 & 443 & 684 & 375 \\
        isl & 127 & 0 & 0 & 0 \\
        ita & 930 & 716 & 1,020 & 636 \\
        jav & 291 & 0 & 0 & 0 \\
        jpn & 624 & 993 & 681 & 779 \\
        kan & 119 & 170 & 703 & 135 \\
        kat & 180 & 0 & 0 & 0 \\
        kaz & 319 & 0 & 0 & 0 \\
        khk & 143 & 0 & 0 & 0 \\
        khm & 184 & 0 & 0 & 0 \\
        kir & 120 & 0 & 0 & 0 \\
        kor & 350 & 541 & 666 & 541 \\
        lao & 183 & 0 & 0 & 0 \\
 
        \bottomrule
    \end{tabular}
    \end{minipage}\hspace{5pt}
    \begin{minipage}[t]{0.45\textwidth}\vspace{0pt}
    \begin{tabular}{@{}ccccc@{}}
        \toprule
        & \multicolumn{4}{c}{\bf \sst}\\\cmidrule{2-5}
         & \multicolumn{2}{c}{\xeng{}} & \multicolumn{2}{c}{\engx{}} \\
        \cmidrule(r){2-3}
        \cmidrule(l){4-5}
        {\bf Language} & Primary & Mined & Primary & Mined \\
        \midrule\\
        \midrule
        lin & 52 & 0 & 0 & 0 \\
        lit & 279 & 0 & 0 & 0 \\
        ltz & 2 & 0 & 0 & 0 \\
        lug & 362 & 0 & 0 & 0 \\
        lvs & 95 & 0 & 0 & 0 \\
        mal & 103 & 0 & 0 & 0 \\
        mar & 106 & 0 & 0 & 0 \\
        mkd & 141 & 0 & 0 & 0 \\
        mlt & 149 & 46 & 688 & 39 \\
        mya & 123 & 0 & 0 & 0 \\
        nld & 1,777 & 1,061 & 1,003 & 962 \\
        nor & 189 & 0 & 0 & 0 \\
        npi & 114 & 0 & 0 & 0 \\
        nya & 99 & 0 & 0 & 0 \\
        oci & 0 & 0 & 0 & 0 \\
        ory & 84 & 0 & 0 & 0 \\
        pan & 188 & 0 & 0 & 0 \\
        pbt & 114 & 0 & 0 & 0 \\
        pes & 366 & 0 & 881 & 0 \\
        pol & 591 & 667 & 726 & 657 \\
        por & 355 & 606 & 983 & 508 \\
        ron & 469 & 588 & 951 & 521 \\
        rus & 290 & 1,093 & 959 & 1,075 \\
        slk & 402 & 427 & 686 & 426 \\
        slv & 377 & 0 & 0 & 0 \\
        som & 138 & 0 & 0 & 0 \\
        spa & 1,694 & 2,335 & 1,035 & 2,209 \\
        srp & 99 & 0 & 0 & 0 \\
        swe & 124 & 0 & 688 & 0 \\
        swh & 342 & 411 & 682 & 392 \\
        tam & 241 & 664 & 654 & 685 \\
        tel & 76 & 426 & 655 & 403 \\
        tgk & 98 & 0 & 0 & 0 \\
        tgl & 82 & 213 & 661 & 169 \\
        tha & 183 & 462 & 641 & 408 \\
        tur & 156 & 375 & 998 & 411 \\
        ukr & 129 & 349 & 662 & 329 \\
        urd & 179 & 555 & 682 & 502 \\
        uzn & 162 & 139 & 695 & 147 \\
        vie & 176 & 666 & 954 & 684 \\
        wol & 13 &  & 0 & 0 \\
        xho & 6 & 0 & 0 & 0 \\
        yor & 128 & 0 & 0 & 0 \\
        yue & 136 & 0 & 0 & 0 \\
        zlm & 157 & 0 & 0 & 0 \\
        zul & 48 & 0 & 0 & 0 \\
        \\
        \bottomrule
    \end{tabular}
    \end{minipage}
    \caption{\small Statistics of \sst{} data used to train our \mfourt{} model. We list the data size in hours of speech. For each language we distinguish between \textsc{Eng-X} for translating from English into that language, and \textsc{X-Eng} for translating into English.
}
\label{tbl:modeling:s2stdata}
\end{table}

\FloatBarrier

\newenvironment{mcsection}[1]
    {%
        \textbf{#1}
        \begin{itemize}[leftmargin=*,topsep=0pt,itemsep=-1ex,partopsep=1ex,parsep=1ex,after=\vspace{\medskipamount}]
    }
    {%
        \end{itemize}
    }

\begin{adjustwidth}{-5pt}{-2pt}
\begin{singlespace}

\tcbset{colback=white!10!white}
\begin{tcolorbox}[title=\textbf{\section{Model Card - \mfourt}},
    breakable, sharp corners, boxrule=0.5pt] 
\begin{mcsection}{Model Details\footnote{For this card, we use the template from \cite{10.1145/3287560.3287596}.}} 
\item Person or organization developing model: \textit{Developed by Meta AI Research}
\item Model date: \textit{August 22nd, 2023}
\item Model version: \mfourtlg and \mfourtmd
\item Model type: \textit{Multitasking \unity with (a) Conformer speech encoder, (b) Transformer text encoder-decoder and (c) Transformer encoder-decoder for \tu}.
\begin{itemize}
    \item \it The exact training algorithm and data used to train \mfourtlg and \mfourtmd are described in the paper: \mfourtcitation 
\item License: \textit{CC-BY-NC 4.0
\footnote{\url{https://creativecommons.org/licenses/by-nc/4.0/legalcode}}}
\item Where to send questions or comments about the model: \\
\url{https://github.com/facebookresearch/seamless_communication/issues}
\end{itemize} 
\end{mcsection}

\begin{mcsection}{Intended Use}
    \item Primary intended uses: \textit{\mfourtlg and \mfourtmd are multilingual and multimodal translation models primarily intended for research in speech and text translation. It allows for:
    \begin{itemize}
        \item \asr: Automatic speech recognition for \nasrlangs languages.
        \item \sst: Speech-to-Speech translation from \nsslangs source speech languages into \ntslangs target speech languages. 
        \item \st: Speech-to-text translation from \nsslangs source speech languages into \ntextlangs target text languages.
        \item \ttst: Text-to-Speech translation from \ntextlangs source text languages into \ntslangs target speech languages.
        \item \mt: Text-to-text translation (MT) from \ntextlangs source text languages into \ntextlangs target text languages.
        \item TTS: Text-to-speech synthesis for 36 languages.
    \end{itemize}
Information on how to use the model can be found in seamless\_communication repository along with recipes for finetuning.}
\item Primary intended users: \textit{Primary users are researchers and machine translation (speech and text) research community. }
\item Out-of-scope use cases: \textit{\mfourt is a research model and is not released for production deployment. \mfourt is trained on general domain data and is not intended to be used with domain specific inputs, such as medical domain or legal domain. The model is not intended to be used for long-form translation. The model was trained on short text and speech inputs, therefore translating longer sequences might result in quality degradation. \mfourt{} translations can not be used as certified translations.} 
\end{mcsection}

\begin{mcsection}{Metrics}
    \item Model performance measures: \textit{For the \st task, \mfourt{} models were evaluated using the \bleu metric adopted by SOTA models in speech-to-text translation. The models were additionally evaluated with \spbleu and \blaser on \st. For \sst, the models are evaluated with \asrbleu and \blaser. For the \mt taks, we report quality in terms of \chrf. For \asr, we report the widely adopted metric of \wer with the text normalized following the normalization in \cite{whisper}.
    Additionally, we performed human evaluation with the XSTS protocol and measured added toxicity, robustness and bias of \mfourtlg. Please refer to \Cref{tbl:eval:metrics} of the \mfourt paper for an exhaustive list of metrics.}
\end{mcsection}

\begin{mcsection}{Evaluation Data}
    \item Datasets: \textit{\fleurs, \flores, \covost and \cvss, \holisticbias and \multilingualholisticbias described in \Cref{sec:statement:evaluation,sec:appendix:ra} of the \mfourt paper.
    }
\item Motivation: \textit{We used \fleurs as it provides
an n-way parallel speech and text dataset in 102 languages, on which we can evaluate \mfourt models on multiple tasks.}

\end{mcsection}

\begin{mcsection}{Training Data}
\item \textit{We used parallel multilingual data from a variety of sources to train the model. %
}
\end{mcsection}

\begin{mcsection}{Ethical Considerations}
 \item \textit{In this work, we took a comprehensive approach to prioritize human users and minimize risks that could be transferred to them. While we have documented various evaluation and responsible AI techniques deployed in our work, here are some additional points to highlight. For one, many languages chosen for this study are low-resource languages.
 While quality translation could improve world readiness and information access for many in these communities, such access could also make groups with lower levels of digital literacy more vulnerable to misinformation or online scams. The latter scenarios could arise if bad actors misappropriate our work for nefarious activities, which we conceive as an example of unintended use.
 Regarding data acquisition, the training data used for model development were mined from various publicly available sources on the web. Although we invested heavily in data cleaning, personally identifiable information may not be entirely eliminated. 
 Finally, although we did our best to optimize for translation quality, toxic, biased, or false outputs produced by the model could remain. These could have an adverse impact on those who rely on these translations to make important decisions (particularly when related to health and safety).} 
\end{mcsection}

\begin{mcsection}{Caveats and Recommendations}
  \item Limitations: \textit{Researchers should consider implementing additional integrity mitigations for ``added toxicity'' when using the model in a research application.}
\end{mcsection}

\end{tcolorbox}
\end{singlespace}
\end{adjustwidth}
\newpage

\end{appendices}

\end{document}